%% file: two_layers_arXiv_v2.tex
\documentclass[11pt]{article}

\usepackage{fullpage}
\usepackage{breqn}
\usepackage{lscape}
\usepackage{setspace}
\usepackage[T1]{fontenc}
\usepackage{lmodern}

\input{arxiv_style}

\input{macros}

\input{nn_symmetry_tikz}

\newcommand{\ploss}{{\cL}}

\newcommand{\ibr}[1]{{[#1]}}
\newcommand{\GLG}[1]{{{\text{GL}(#1,\real)}}}
\newcommand{\Od}[1]{{{\text{O}(#1)}}}

\PassOptionsToPackage{numbers}{natbib}
\usepackage[final]{pdfpages}
\usepackage{parskip}

\definecolor{darkblue}{rgb}{0, 0, 0.6}

\usepackage[utf8]{inputenc} 
\usepackage[T1]{fontenc}    
\usepackage{hyperref}       
\usepackage{url}            
\usepackage{booktabs}       
\usepackage{amsfonts}       
\usepackage{nicefrac}       
\usepackage{microtype}      
\usepackage{xcolor}         
\usepackage[utf8]{inputenc} 
\usepackage[T1]{fontenc}    

\def\ddefloop#1{\ifx\ddefloop#1\else\ddef{#1}\expandafter\ddefloop\fi}
\def\ddef#1{\expandafter\def\csname 
	bb#1\endcsname{\ensuremath{\mathbb{#1}}}}
\ddefloop ABCDEFGHIJKLMNOPQRSTUVWXYZ\ddefloop
\def\ddef#1{\expandafter\def\csname 
	#1\endcsname{\ensuremath{{\bm{#1}}}}}
\ddefloop ABCDEFGHIJKLMNOPQRSTUVWXYZ\ddefloop
\def\ddef#1{\expandafter\def\csname 
	#1\endcsname{\ensuremath{{\bm{#1}}}}}
\ddefloop abcdefghijklmnopqrstuvwxyz\ddefloop
\def\ddefloop#1{\ifx\ddefloop#1\else\ddef{#1}\expandafter\ddefloop\fi}
\def\ddef#1{\expandafter\def\csname 
	b#1\endcsname{\ensuremath{\mathbf{#1}}}}
\ddefloop ABCDEFGHIJKLMNOPQRSTUVWXYZ\ddefloop
\def\ddef#1{\expandafter\def\csname 
	c#1\endcsname{\ensuremath{\mathcal{#1}}}}
\ddefloop ABCDEFGHIJKLMNOPQRSTUVWXYZ\ddefloop
\def\ddef#1{\expandafter\def\csname 
	f#1\endcsname{\ensuremath{\mathfrak{#1}}}}
\ddefloop ABCDEFGHIJKLMNOPQRSTUVWXYZ\ddefloop
\def\ddef#1{\expandafter\def\csname 
	h#1\endcsname{\ensuremath{\widehat{#1}}}}
\ddefloop ABCDEFGHIJKLMNOPQRSTUVWXYZ\ddefloop
\def\ddef#1{\expandafter\def\csname 
	hc#1\endcsname{\ensuremath{\widehat{\mathcal{#1}}}}}
\ddefloop ABCDEFGHIJKLMNOPQRSTUVWXYZ\ddefloop
\def\ddef#1{\expandafter\def\csname 
	t#1\endcsname{\ensuremath{\widetilde{#1}}}}
\ddefloop ABCDEFGHIJKLMNOPQRSTUVWXYZ\ddefloop
\def\ddef#1{\expandafter\def\csname 
	tc#1\endcsname{\ensuremath{\widetilde{\mathcal{#1}}}}}
\ddefloop ABCDEFGHIJKLMNOPQRSTUVWXYZ\ddefloop
\usepackage[normalem]{ulem}
\newcommand{\fx}{\mathfrak{x}}
\newcommand{\fy}{\mathfrak{y}}
\newcommand{\fs}{\mathfrak{s}}
\newcommand{\ft}{\mathfrak{t}}

\usepackage{hyperref}       
\usepackage{url}            
\usepackage{booktabs}       %
\usepackage{amsfonts}       
\usepackage{nicefrac}       
\usepackage{microtype}      

\usepackage{array,multirow}
\title{Analytic Study of Families of Spurious Minima in Two-Layer  \\  
ReLU Neural Networks: A Tale of Symmetry II}

	\author{
	Yossi Arjevani \\
	The Hebrew University\\
	\texttt{yossi.arjevani@gmail.com} \\
	\and
	Michael Field\\                                 %
	UC Santa Barbara\\
	\texttt{mikefield@gmail.com}\\
}
\date{}

\allowdisplaybreaks

\begin{document}
\maketitle

\begin{abstract}
	We study the optimization problem associated with fitting two-layer ReLU 
	neural networks with respect to the squared loss, where labels are 
	generated 
	by a target network. We make use of the rich symmetry structure to 
	develop a novel set of tools for studying families of spurious 
	minima. In contrast to existing approaches which operate 
	in limiting regimes, our technique directly addresses the nonconvex loss 
	landscape for a finite number of inputs $d$ and neurons $k$, and 
	provides analytic, rather than heuristic, information. In particular, we 
	derive 
	analytic estimates for the loss at different minima, and prove 
	that modulo $O(d^{-1/2})$-terms the Hessian spectrum 
	concentrates near small positive constants, with the exception of 
	$\Theta(d)$ 
	eigenvalues which grow linearly with~$d$. We further show that the Hessian 
	spectrum at global and spurious minima coincide to $O(d^{-1/2})$-order, 
	thus challenging our ability to argue about statistical 
	generalization through local curvature.
	Lastly, our technique provides the exact 
	\emph{fractional} dimensionality at which families of critical points turn 
	from saddles into spurious minima. This makes possible the study of 
	the creation and the annihilation of spurious minima using powerful tools 
	from 
	equivariant bifurcation theory. 
\end{abstract}

One of the outstanding conundrums of deep learning concerns the ability 
of simple gradient-based methods to successfully train neural 
networks despite the nonconvexity of the associated optimization 
problems. Indeed, generic nonconvex optimization landscapes can exhibit 
wide and flat basins of attraction around poor local minima which may lead to 
a complete failure of such methods. The nature by which nonconvex problems 
associated with neural networks deviate from generic ones is currently not 
well-understood. In particular, much of the dynamics of gradient-based methods 
follows from the curvature of the loss landscape around local minima. It is 
therefore vital to study the local geometry of spurious (i.e., 
non-global local) and global minima in order to understand the 
mysterious mechanism which drives gradient-based methods towards minima of 
high quality. However, already establishing the very existence of 
spurious minima seems to be beyond reach of existing analytic tools; let alone 
rigorously 
arguing about their height, curvature and structure---the aim of this work. 

\renewcommand{\theequation}{\arabic{equation}}

In this paper, we focus on two-layer ReLU neural networks of the 
form 
\begin{align} \label{net}
	\sum_{i=1}^k \alpha_i \varphi(\inner{\w_i,\x}), 
	~W\in 
	M(k,d),~\balpha \in \RR^k,
\end{align}
where $\varphi(z)=\max\crl{0,z}$ is the ReLU activation function acting 
entry-wise, $M(k,d)$ denotes the space of $k\times d$ matrices 
and $\w_i$ denotes the $i$th row of $W$. We are primarily interested in
characterizing various optimization-related obstructions for local 
search method, independently of the expressive power of two-layer ReLU 
networks. Thus, data is understood to be fully realizable. Concretely, we 
assume that there are $d$ inputs which are drawn from the standard multivariate 
Gaussian distribution and are labeled by a \emph{planted} target network. We 
consider directly optimizing the expected squared loss, which results in 
the following highly nonconvex optimization problem:
\begin{align}\label{opt:problem}
	\cL(W, \balpha) \defeq \frac{1}{2}\EE_{\x\sim 
		\cN(\bzeros,I_d)}\brk*{\Big(\sum_{i=1}^k \alpha_i 
		\varphi(\inner{\w_i,\x}) 
		- \sum_{i=1}^d \beta_i 	\varphi(\inner{\v_i,\x})\Big)^2},
\end{align}
where $k$ denotes the number of hidden neurons, $W\in M(k,d)$ and $\alpha_i 
\in \RR^k$ are the optimization variables, and $\v_i$ and $\beta_i$ are fixed 
parameters. This setting, in which the data distribution is regulated rather 
than being allowed to admit worst-case behavior, has drawn a considerable 
amount of interest in recent years 
\cite{zhang2017electron,du2017gradient,feizi2017porcupine,
	li2017convergence,tian2017analytical,brutzkus2017globally,
	ge2017learning,safran2017spurious,arjevanifield2020hessian,akiyama2021learnability},
in part due to 
the growing number of evidences which indicate that any explanation for the 
empirical success of deep learning (DL) must take into account the intricate 
interplay between the network architecture, the input distribution and the 
label distribution (cf. 
\cite{brutzkus2017globally,blum1992training,shamir2018distribution} and 
references therein for hardness results of optimization and learnability 
under partial sets of assumptions). Moreover, as demonstrated later in the 
paper, in spite of its apparent simplicity, nonconvex problem 
(\ref{opt:problem}) shares a few important characteristics with full-scale 
neural networks, such as low-dimensional minima and extremely skewed Hessian 
spectrum.

Learning problem (\ref{opt:problem}) has also been studied in the statistical 
physics community starting from the 80' 
\cite{gardner1989three, seung1992statistical,kinzel1990improving, 
	watkin1993statistical,engel2001statistical,biehl1995learning,saad1995exact,
	saad1995line,riegler1995line}  under the student-teacher (ST) framework, in 
which one aims to adjust a \emph{student} network so as to fit the 
output of a \emph{teacher} network. The ST framework offers a  clean 
venue for analyzing optimization-related aspects of neural network models in 
the spirit of physical reductionism. The success of DL models in the past 
decade has reinitiated a surge of interest in this framework, e.g., 
\cite{aubin2019committee,	
	goldt2019dynamics,mannelli2020optimization,oostwal2021hidden}. However, 
	despite 
the 
long tradition in the statistical physics community and the wide effort 
put nowadays by the machine learning community, the perplexing geometry of 
problem (\ref{opt:problem}) still seems to be out of reach of existing 
analytic tools in regimes encountered in practice. In this paper, we present a 
novel set of symmetry-based tools which allows us, for the first time, to 
analytically characterize various important aspects of the associated highly 
nonconvex landscape for a finite number of inputs and neurons.

Our contributions, in order of appearance, can be stated as follows:
\begin{itemize}[leftmargin=*]
	\item We demonstrate that, empirically, and in a well-defined sense, 
	minima in two-layer ReLU neural networks \emph{break} the 
	symmetry of the target weight matrix. Although ReLU networks 
	have been studied for many years, this phenomenon of symmetry breaking 
	seems to have gone largely~unnoticed.
	
	\item We show that symmetry breaking makes it possible to derive 
	\emph{analytic} expressions for families of spurious minima in the form of 
	\emph{fractional} power series in terms of $d$ and $k$. Crucially, in 
	contrast to existing approaches which employ various limiting processes, 
	e.g., 
	\cite{aubin2019committee,goldt2019dynamics,mannelli2020optimization,oostwal2021hidden,
		goldt2019generalisation,mei2018mean,chizat2018global,		
		jacot2018neural,daniely2016toward}, our method operates in the natural 
	regime where $d$ and $k$ are finite. 
	
	\item We develop a novel technique which yields an analytic 
	characterization of the Hessian spectrum of minima to 	
	$O(d^{-1/2})$-order for $d\ge k$, and determine the exact \emph{fractional} 
	value of $d$ at which critical points turn from saddles into spurious 
	minima---a~key ingredient in understanding how 
	over-parameterization annihilates spurious minima.
	
	\item Based on the unique access to high-dimensional spectral 
	information, we closely examine a number of hypotheses in the machine 
	learning literature pertaining to curvature, optimization and 
	generalization. In particular, we prove that the Hessian spectrum at minima 
	concentrates near small positive constants, with the exception of 
	$\Theta(d)$ eigenvalues which grow 
	linearly with $d$. Although this phenomenon of extremely skewed spectrum 
	has been observed many times  		
	\cite{bottou1991stochastic,lecun2012efficient,sagun2016eigenvalues,sagun2017empirical},
	to our knowledge, this is the first time it has been established  
	\emph{rigorously} for two-layer ReLU networks. In addition, our analysis 
	shows that the Hessian spectra of spurious and global 
	minima are identical to $O(d^{-1/2})$-terms, and further implies 
	that the inductive bias of stochastic gradient descent (SGD), provably, 
	can not be exclusively explained in terms of local curvature 	 
	\cite{hochreiter1997flat,	
		keskar2016large,jastrzkebski2017three,wu2017towards,yao2018hessian,
		chaudhari2019entropy,dinh2017sharp}.

\end{itemize}

The results presented in the paper are threefold: identifying a 
symmetry breaking principle for two-layer ReLU networks, analytically 
characterizating spurious minima, and computing the Hessian spectrum. The next 
three sections are organized accordingly, 
along with a literature survey of related work. The last section is devoted for 
a high-level description of the novel symmetry-based technique used in this 
work. Proofs, formalities and lengthy technical details are deferred to the 
appendix.

\begin{figure}[!htb]
	\begin{minipage}{.49\textwidth}
		\centering	
		\includegraphics[scale=0.5]{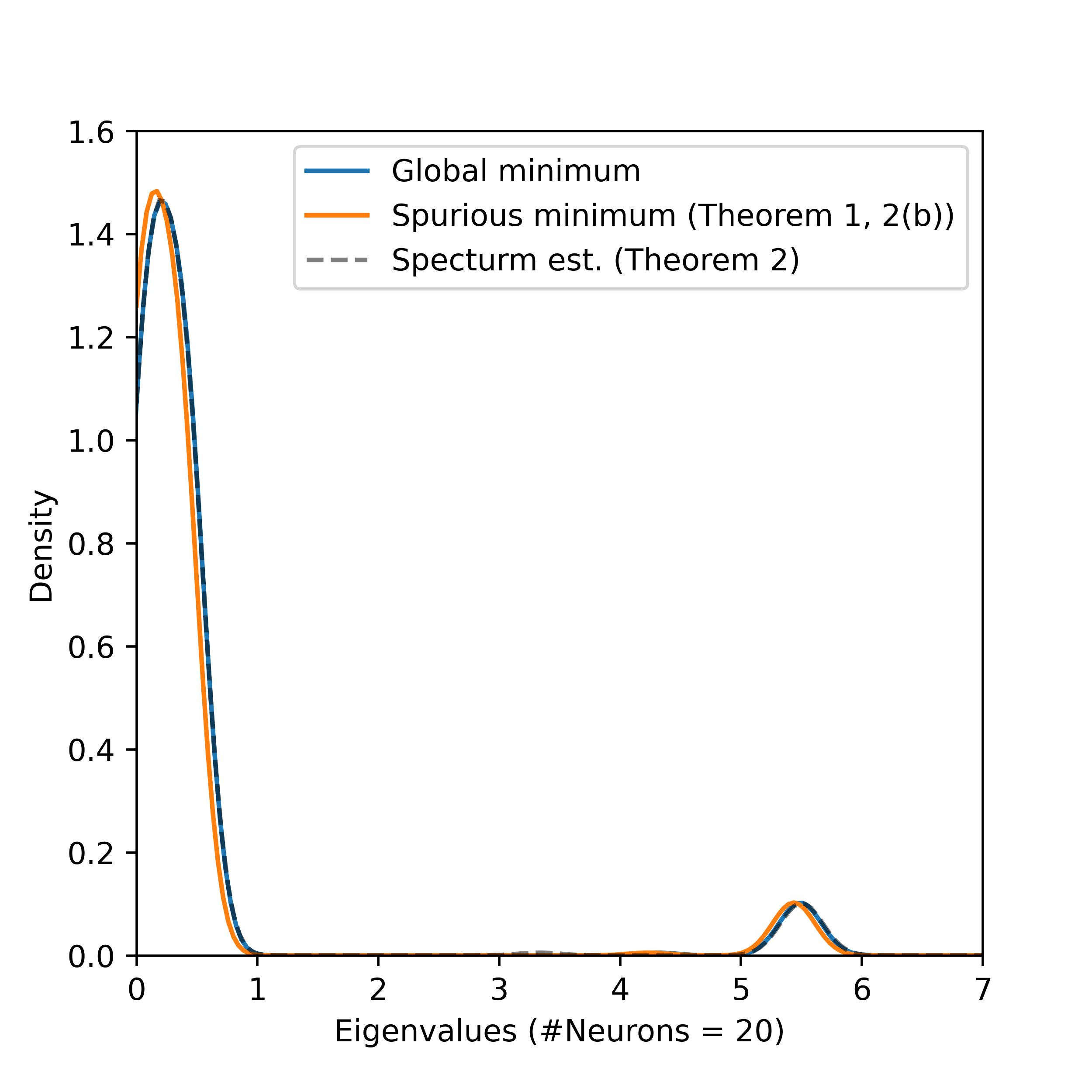}
	\end{minipage}	
	\begin{minipage}{.49\textwidth}
		\centering	
		\includegraphics[scale=0.5]{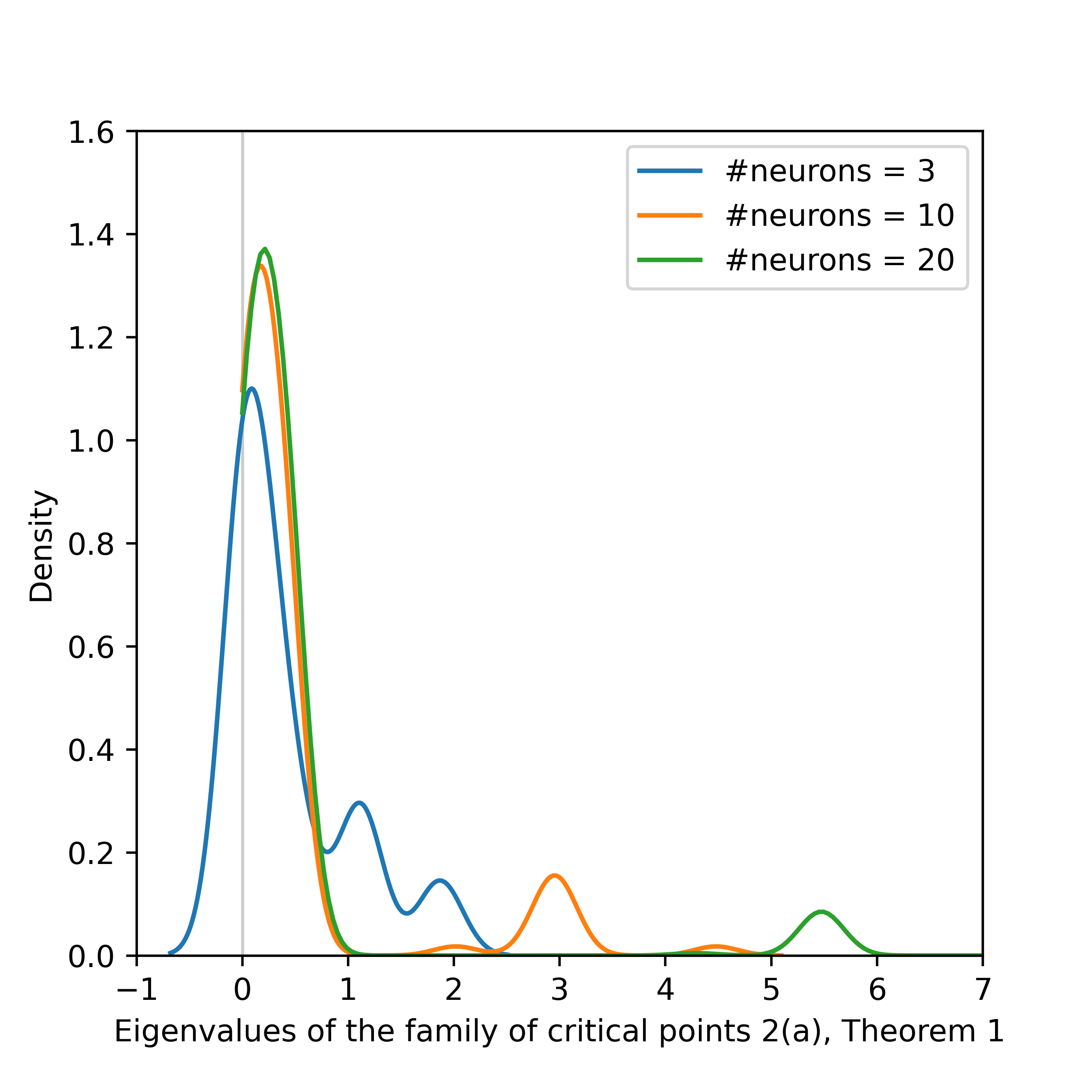}
	\end{minipage}	
	\caption{(Left) Our symmetry-based technique yields 
		an \emph{analytic} 
		characterization of the Hessian spectrum (see \pref{thm:spec}) to 
		$O(d^{-1/2})$-order (in a dashed line) which provides a 
		good approximation already for small values of inputs and 
		neurons (in solid lines). The analysis further  
		implies that the Hessian spectrum of global and 
		various types of spurious minima agree to within 
		$O(d^{-1/2})$-accuracy. 
		(Right) the Hessian spectrum is extremely skewed and tends
		to concentrate near small positive constants with the exception of 
		$\Theta(d)$ eigenvalues which grow linearly with $d$. Observe that 
		the family of critical points considered here (see 
		\pref{thm:power_series}, case \ref{typeII}) turns from saddles 
		at $d=3$ into spurious minima when 	$d\ge 10$. Our analysis shows that 
		the change of stability occurs at 
		$d\approx 5.71$, and more importantly, indicates that the process can 
		in fact be reversed, namely, spurious minima can be turned into saddles 
		by over-parameterizing (i.e., increasing the number of hidden 
		neurons).} 
	\label{fig:evs}
	
\end{figure}
\renewcommand{\bbeta}{\boldsymbol{\beta}}
\section{Symmetry breaking in two-layer ReLU neural networks} 
Optimization problem (\ref{opt:problem}) exhibits a very rich symmetry 
structure. Indeed, the loss function $\ploss$ is invariant to left- and 
right-multiplication of $W$ by permutation matrices  (see 
\pref{sec:inv_properties} for 
a formal proof), i.e., $\ploss(W,\balpha) = \ploss(P_\pi  WP^\top_\rho, 
\balpha)$ for all 
$(\pi,\rho) \in S_k\times S_d$, where $S_m$ generally denotes the symmetric 
group of degree~$m$, and 
\begin{align*}
	(P_\pi)_{ij} = 
	\begin{cases}
		1 & i = \pi(j),\\
		0 & \text{o.w.}
	\end{cases}.
\end{align*}
It is therefore natural to ask how the critical points of $\ploss$ 
reflect this symmetry. For example, the identity matrix $I_d$, 
one of the global minimizers of $\ploss$ for the case in which $V=I_d$, 
$\balpha = \bbeta=\bones_d$ and $k=d$, is invariant under \emph{simultaneous} 
left- and right-multiplication by any permutation matrix. Indeed, $P_\pi I_d
P_\pi^\top = P_\pi P_\pi^\top = I_d$ for any $\pi\in S_d$. (Modulo group 
conjugation, this holds for any global minimizer of $\ploss$. See 
\cite[Proposition 4.14.]{ArjevaniField2020}.) This simple observation is 
conveniently stated using the concept of the \emph{isotropy} group. Given a 
weight matrix $W\in M(k,d)$, we let
\newcommand{\iso}[1]{\mathrm{Iso}(#1)}
\begin{align}\label{def:iso}
	\iso{W} \defeq \{(\pi,\rho)~|~ (\pi,\rho)\in S_k\times S_d,~ P_\pi W 
	P_\rho^\top = W\},
\end{align}
Thus, we have $\iso{I_d} = \Delta S_d$, where 
$\Delta$ maps any subgroup $H\subseteq S_d$ to its diagonal counterpart 
$\Delta H \defeq \{(h,h)~|~h \in H\}\subseteq S_d\times S_d$. Empirically, and 
somewhat miraculously, spurious minima and saddles of $\ploss$ tend to be 
highly symmetric in the (formal) sense that their isotropy groups are 
(conjugated to) large subgroups of $\Delta S_d$, the isotropy of the global 
minima (see 
\pref{fig:max_sym}). Thus, the principle of symmetry breaking can be 
concisely phrased as follows: 
\begin{center}
	\Large Spurious minima \emph{break the symmetry} of global minima.
\end{center}
The principle extends to more general target networks; when 
the isotropy of the target weight matrix $V$ changes, the symmetry of spurious 
minima belonging to the respective optimization problem. In 
\pref{sec:sym_prin}, we provide a series of experiments which empirically 
corroborates symmetry breaking for optimization problem (\ref{opt:problem}).

\newcommand{\bwidth}{1.2in}
\newcommand{\bheight}{1.2in}
\newcommand{\dwidth}{0.11in}
\begin{figure}[ht]
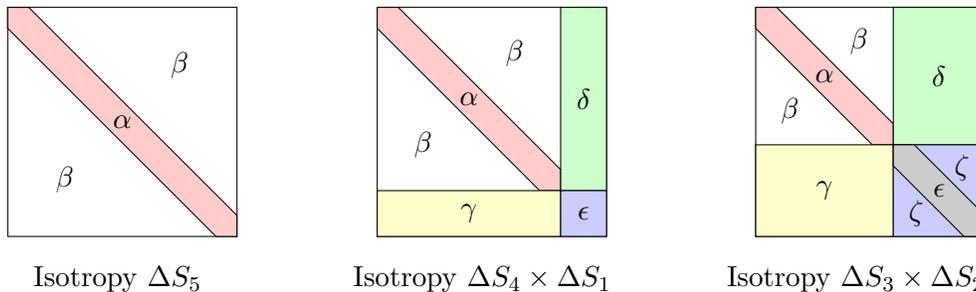
 
	\begin{center}\vskip-0.4cm
		{\setstretch{0.8}
			\begin{blockmatrixtabular}\label{fig:dia_sym}
				
				\valignbox{\dblockmatrixSD[1.0,0.8,0.8]{\bwidth}{\bheight}{$\alpha$}{\dwidth}{$\beta$}}&
				\quad\quad\quad\quad
				
				\valignbox{\dblockmatrixSDMO[1.0,0.8,0.8]{\bwidth}{\bheight}{$\alpha$}{\dwidth}{$\beta$}{0.24in}}&
				\quad\quad\quad\quad
				
				\valignbox{\dblockmatrixSDMOOO[1.0,0.8,0.8]{\bwidth}{\bheight}{$\alpha$}{\dwidth}{$\beta$}{0.48in}}&\\
				\\
				Isotropy $\Delta S_5 $& 		 
				\quad\quad\quad\quad 
				Isotropy  $\Delta S_4 \times 
				\Delta 
				S_1$  & 		 \quad\quad\quad\quad Isotropy 
				$\Delta 
				S_3\times \Delta
				S_2$
		\end{blockmatrixtabular}}
	\end{center}
	\vskip-0.4cm		
	\caption{A schematic description of $5\times 5$ matrices 
		with isotropy 	$\Delta S_5, \Delta S_4 \times \Delta S_1$ 
		and $\Delta S_3\times S_2$, from left to right (borrowed 
		from \cite{arjevanifield2019spurious}). $\alpha, \beta, 
		\gamma,\delta, \epsilon$ and $\zeta$ are assumed to be `sufficiently' 
		different.} 
	\label{fig:max_sym}
\end{figure}
The principle of symmetry breaking for ReLU networks was first studied 
in \cite{arjevanifield2019spurious,ArjevaniField2020}, and was later extended 
to various tensor decomposition problems in \cite{arjevanifield2021tensor}. 
Here, we analyze two-layer ReLU networks where \emph{both} layers are 
trainable, an architecture which has received a considerable amount of 
attention in recent years, e.g.,  
\cite{zhang2017electron,du2017gradient,feizi2017porcupine,li2017convergence,tian2017analytical,brutzkus2017globally,ge2017learning}.
We note in passing that, in its broader sense, the principle of symmetry 
breaking has been observed many times in various scientific fields, e.g., 
Higgs-Landau theory, equivariant bifurcation theory and replica 
symmetry-breaking (see, e.g., 
\cite{Michel1980,Golubitsky1983,field1987symmetry,DingSS15}). Two-layer ReLU 
networks seems to form a rather unexpected instance of this principle.

One intriguing quality of nonconvex landscapes in which the symmetry breaking 
principle applies is that local minima lie in fixed low-dimensional spaces 
(see \pref{sec:method}). A similar phenomenon of  hidden low-dimensional 
structure has been observed in various learning problems in DL with real 
datasets \cite{li2018measuring,gur2018gradient}, and is believed by some to 
be an important factor of learnability in nonconvex settings. In the context of 
this work, the hidden low-dimensionality of spurious minima turns out to be a 
key ingredient to our analytic study, as we now present.

\section{Power series representation of families of spurious minima}
Although the problem of fitting neural networks under 
the ST framework have been studied for more than 30 years, the symmetry 
breaking principle exhibited by optimization problem (\ref{opt:problem}) seems 
to have gone largely unnoticed. Early studies adopted tools (at times, 
heuristic, e.g., \cite{mezard1987spin,mezard2009information}) from 
statistical physics to analyze phase transitions and generalization 
errors \cite{gardner1989three, seung1992statistical,kinzel1990improving, 
	watkin1993statistical}. Later work focused on the dynamics of SGD and 
	studied 
the evolution of the generalization error along the optimization process 
through 
a set of carefully-derived ODEs 
\cite{engel2001statistical,biehl1995learning,saad1995exact,saad1995line,riegler1995line}.
Following the empirical success of DL in the past decade, this line of work 
has recently drawn a renewed interest, e.g, 
\cite{aubin2019committee,goldt2019dynamics,oostwal2021hidden,goldt2019generalisation},
which puts past analyses on rigorous grounds and addresses a broader 
class of architectures and activation functions. The methods used in 
this long line of works operate in the \emph{thermodynamic limit}---where the 
number of inputs is taken to infinity. Thus, the formal validity of the 
results to finite width networks is currently limited. 

Other common approaches for analyzing problem (\ref{opt:problem}) are 
based on: mean-field \cite{mei2018mean}, optimal control 
\cite{chizat2018global}, NTK \cite{jacot2018neural} and compositional kernels 
\cite{daniely2016toward}. These approaches offer, in essence, convex surrogates 
which apply in strict parameter regimes (including algorithmic parameters, such 
as learning rates). Similarly to the thermodynamic limit approach, the 
convex surrogates are obtained by limiting processes---this time by taking the 
number of hidden neurons to infinite. A growing number of works has severely 
limited, if not invalidated, the explanatory power of these approaches for 
network widths encountered in practice 
\cite{yehudai2019power,ghorbani2020neural}. 

In sharp contrast to the approaches discussed above, our technique directly 
addresses the associated highly nonconvex optimization landscape in the natural 
regime where the number of inputs and neurons is finite, and provides 
analytic, rather than heuristic, information. This is obtained by exploiting 
the presence of symmetry breaking phenomena whereby fractional
power series representation for families of spurious minima are 
derived. 

Below, we provide analytic expressions for families of minima of different 
isotropy (see \pref{def:iso} and \pref{fig:max_sym}), along with their 
respective objective value. For brevity, expressions are given to 
$O(d^{-3/2})$-order. In the appendix, we list additional $O(d^{-5/2})$-order 
terms which are required for computing the Hessian spectrum. The 
invariance properties of optimization problem (\ref{opt:problem}) imply 
that additional families of minima can be obtained by permuting the rows and 
the columns of a given family. The number of new distinct minima generated 
using 
such transformations (i.e., the minima multiplicity) depends on the very 
structure of 
the family of minima under consideration, and is stated in 
\pref{thm:power_series}. 
Lastly, for any $\lambda>0$ and $i\in[d]$, the value of the ReLU network 
(\ref{net}) remains fixed under $(\w_i,\alpha_i) \mapsto (\lambda \w_i, 
\alpha_i/\lambda)$. This degree-of-freedom is expressed by using the slack 
variables $\lambda_i>0,~i\in[d]$ below.

\begin{theorem} \label{thm:power_series}
	Optimization problem (\ref{opt:problem})
	with $k=d$, $V=I_d$ and $\bbeta=\bones_d$ possesses the following 
	families of minima for $d\ge9$:
	\begin{enumerate}[leftmargin=*]
		\item Two families of $\Delta S_d$-minima of multiplicity $d!$ with 
		$W(d) = 
		\Diag{\lambda_1,\dots,\lambda_d} A_{d}(a_1,a_2)$ and $\alpha(d) 
		= (\lambda_1^{-1},\dots,\lambda_d^{-1})$, where $\lambda_i >0$, 
		\label{case_one}
		\begin{align*}
			A_{d}(a_1,a_2) \defeq
			\begin{pmatrix}
				a_1 & a_2 & \dots && a_2 \\	
				a_2 & a_1 & a_2 & \dots & a_2 \\
				&\vdots &&\vdots \\
				a_2 & \dots && a_2 & a_1 \\
			\end{pmatrix}\in M(d,d),
		\end{align*} 
		$\mathrm{Diag}(\cdot)$ maps a given vector to the diagonal entries of a 
		diagonal matrix, and
		\begin{enumerate}[leftmargin=*]
			\item $a_1 = 1$ and $a_2 = 0$ (a global minimizer), in which case 
			$$\cL(W(d),\alpha(d)) = 0.$$ \label{global_case}
			\item $a_1 = -1 + \frac{2}{d} +  O\prn*{d^{\frac{-3}{2}}},~
			a_2 = \frac{2}{d} +  O\prn*{d^{\frac{-3}{2}}}$, in which case 
			\label{typeA}
			\begin{align}
				\cL(W(d),\alpha(d)) = - \frac{1}{\pi} + \frac{1}{2}  - 
				\frac{4}{3 \pi \sqrt{d}} 
				+ \frac{- \frac{1}{2} - \frac{2}{\pi^{2}} + \frac{3}{\pi}}{d} + 
				O\prn*{d^{\frac{-3}{2}}}.
			\end{align}
		\end{enumerate}
		
		\item Two families of $\Delta (S_{d-1}\times S_1)$-minima of 
		multiplicity 
		$d\cdot d!$ with 
		\begin{align*}
			W(d) = \Diag{\lambda_1,\dots,\lambda_d} 
			\prn*{\begin{array}{c|c}
					A_{d-1}(a_1,a_2) & a_3\cI_{d-1,1}\\\hline
					a_4\cI_{1,d-1} & a_5
			\end{array}},
		\end{align*}
		and $\alpha(d) = (\lambda_1^{-1},\dots,\lambda_d^{-1})$, where 
		$\lambda_i >0$ 
		and
		\begin{enumerate}[leftmargin=*]
			\item \label{typeII}
			$a_1 = 1 +  O\prn*{d^{\frac{-3}{2}}},~
			a_2 =  O\prn*{d^{\frac{-3}{2}}},~
			a_3 = \frac{2}{d} +  O\prn*{d^{\frac{-3}{2}}},~
			a_4 = \frac{4}{\pi d} +  O\prn*{d^{\frac{-3}{2}}}$ and 
			$a_5 = -1 + \frac{\frac{8}{\pi^{2}} + 2 + \frac{8}{\pi}}{d} +  
			O\prn*{d^{\frac{-3}{2}}}$, in 
			which case 
			\begin{align}
				\cL(W(d),\alpha(d)) = \frac{\frac{1}{2} - 
				\frac{2}{\pi^{2}}}{d}	 +  
				O\prn*{d^{\frac{-3}{2}}}.
			\end{align}
			\item \label{typeI}
			$a_1 = -1 + \frac{2}{d} +  O\prn*{d^{\frac{-3}{2}}},~
			a_2 = \frac{2}{d} +  O\prn*{d^{\frac{-3}{2}}},~
			a_3 =  O\prn*{d^{\frac{-3}{2}}},~
			a_4 = \frac{2 - \frac{4}{\pi}}{d} +  O\prn*{d^{\frac{-3}{2}}}$ and
			$a_5 = 1 + \frac{8 \left(-1 + \pi\right)}{\pi^{2} d} +  	
			O\prn*{d^{\frac{-3}{2}}}$, in which case 
			\begin{align}
				\cL(W(d),\alpha(d)) =- \frac{1}{\pi} + \frac{1}{2} - \frac{4}{3 
				\pi 
					\sqrt{d}} + \frac{-1 - 	\frac{4}{\pi^{2}} + 
					\frac{5}{\pi}}{d}  + 
				O\prn*{d^{\frac{-3}{2}}}.
			\end{align}
		\end{enumerate}
		
		\item Two families of $\Delta (S_{d-2}\times S_2)$-minima of 
		multiplicity 
		$d!\binom{d}{2}$ with 
		\begin{align*}
			W(d) = \Diag{\lambda_1,\dots,\lambda_d} 
			\prn*{\begin{array}{c|c}
					A_{d-2}(a_1,a_2) & a_3\cI_{d-2,2}\\\hline
					a_4\cI_{2,d-2} & A_{2}(a_5,a_6)
			\end{array}},
		\end{align*}
		and $\alpha(d) = (\lambda_1^{-1},\dots,\lambda_d^{-1})$, where 
		$\lambda_i >0$ 
		and
		\begin{enumerate}[leftmargin=*]
			\item \label{typeM_II}
			$a_1 = 1 +  O\prn*{d^{\frac{-3}{2}}},~
			a_2 = 0 +  O\prn*{d^{\frac{-3}{2}}},~
			a_3 = \frac{2}{d} +  O\prn*{d^{\frac{-3}{2}}},~
			a_4 = \frac{4}{\pi d} +  O\prn*{d^{\frac{-3}{2}}},~
			a_5 = -1 + \frac{\frac{8}{\pi^{2}} + 2 + \frac{8}{\pi}}{d} +  
			O\prn*{d^{\frac{-3}{2}}}$ and 
			$a_6 = \frac{2 \left(- 12 \pi + 16 + \pi^{3} + 4 
			\pi^{2}\right)}{\pi^{2} d 
				\left(2 + \pi\right)} +  O\prn*{d^{\frac{-3}{2}}}$, in which 
				case 
			\begin{align}
				\cL(W(d),\alpha(d)) = \frac{-4 + \pi^{2}}{\pi^{2} d} + 
				O\prn*{d^{\frac{-3}{2}}}.
			\end{align}
			\item \label{typeM_I}
			$a_1 = -1 + \frac{2}{d} +  O\prn*{d^{\frac{-3}{2}}},~
			a_2 = \frac{2}{d} +  O\prn*{d^{\frac{-3}{2}}},~
			a_3 = 0 +  O\prn*{d^{\frac{-3}{2}}},~
			a_4 = \frac{2 - \frac{4}{\pi}}{d} +  O\prn*{d^{\frac{-3}{2}}},~
			a_5 = 1 + \frac{8 \left(-1 + \pi\right)}{\pi^{2} d} +  
			O\prn*{d^{\frac{-3}{2}}}$ and $a_6 = \frac{4 \left(- \pi^{2} - 8 + 
			6 
				\pi\right)}{\pi^{2} d \left(2 + \pi\right)} +  
				O\prn*{d^{\frac{-3}{2}}}$, 
			in which case 
			\begin{align}
				\cL(W(d),\alpha(d)) = - \frac{1}{\pi} + \frac{1}{2} - 
				\frac{4}{3 \pi 
					\sqrt{d}} + \frac{- \frac{3}{2} - 
					\frac{6}{\pi^{2}} + \frac{7}{\pi}}{d} + 
				O\prn*{d^{\frac{-3}{2}}}.
			\end{align}
			
		\end{enumerate}

		\item One family of $\Delta (S_{d-3}\times S_3)$-minima of 
		multiplicity $d!\binom{d}{3}$ with  
		\label{typeN_II}
		\begin{align*}
			W(d) = \Diag{\lambda_1,\dots,\lambda_d} 
			\prn*{\begin{array}{c|c}
					A_{d-3}(a_1,a_2) & a_3\cI_{d-3, 3}\\\hline
					a_4\cI_{3,d-3} & A_{2}(a_5,a_6)
			\end{array}},
		\end{align*}
		and $\alpha(d) = (\lambda_1^{-1},\dots,\lambda_d^{-1})$, where 
		$\lambda_i >0$ 
		and $a_1 = 1 +  O\prn*{d^{\frac{-3}{2}}},~
		a_2 =  O\prn*{d^{\frac{-3}{2}}},~
		a_3 = \frac{2}{d} +  O\prn*{d^{\frac{-3}{2}}},~
		a_4 = \frac{4}{\pi d} +  O\prn*{d^{\frac{-3}{2}}},~
		a_5 = -1 + \frac{\frac{8}{\pi^{2}} + 2 + \frac{8}{\pi}}{d} +  
		O\prn*{d^{\frac{-3}{2}}}$ and $a_6 = \frac{2 \left(- 12 \pi + 16 + 
		\pi^{3} + 4 
			\pi^{2}\right)}{\pi^{2} d \left(2 + \pi\right)} +  
			O\prn*{d^{\frac{-3}{2}}}$, 
		in which case 
		\begin{align}
			\cL(W(d),\alpha(d)) = \frac{\frac{3}{2} - \frac{6}{\pi^{2}}}{d} + 
			O\prn*{d^{\frac{-3}{2}}}.
		\end{align}
		
	\end{enumerate}
\end{theorem}
The idea of the proof of \pref{thm:power_series} is given in 
\pref{sec:pf_power_series}, and the multiplicity computation, a simple 
application of the orbit-stabilizer theorem, in \pref{sec:mul}. We note that  
the same technique can be used for other choices of target networks, mutatis 
mutandis. The power series stated in \pref{thm:power_series} also represent 
spurious minima for the under-parameterized $d>k$-case. This is obtained by 
appropriately padding the entries of the weight matrices by zeros (see 
\cite[section 4.3]{ArjevaniField2020} for details). The derivation of the 
respective Hessian spectrum is then an immediate application of \cite[section 
E.1]{arjevanifield2020hessian} to the spectral analysis given in 
\pref{thm:spec} below (for the case where $d=k$). By contrast, studying 
the over-parameterized regime for which $d<k$ requires qualitatively different 
tools and is outside the scope of this work.

The analysis above reveals that not all local minima are alike: while the 
objective value of some minima decays like $\Theta(1/d)$, in other cases the 
objective value converges to the positive constant $\frac{1}{2}- 
\frac{1}{\pi}$. In particular, except for the global minima case 
\ref{global_case}, all minima described above are spurious. The 
difference between the two types of behavior seems to lie in the limiting 
values of the entries of $W(d)$; in the former, the diagonal mainly consists 
of ones, whereas in the latter mainly minus ones. This is consistent with the 
fact that all the teacher's diagonal entries are ones. 

For reasons which are yet to be understood, empirically, the bias 
induced by Xavier initialization \cite{glorot2010understanding} seems to favor 
the class of minima for which the objective value decays as $\Theta(1/d)$. 
This is not to say that under Xavier initialization the expected value upon 
convergence must decrease with $d$. The objective value decays to zero at 
different rates (see cases \ref{typeII}, \ref{typeM_II}, \ref{typeN_II} 
above) and depends on the probability to converge to a given type of minima. (A 
possible proxy to the latter is the minima multiplicity stated in 
\pref{thm:power_series}.)

\section{Analytic study of the Hessian spectrum}
Once a power series representation has been obtained, it is possible to 
analytically characterize various important properties of families of 
minima. Here, we use this representation to compute yet another fractional
power series---this time, of the Hessian spectrum.

\begin{theorem}\label{thm:spec}
	Assuming all the weights of the second layer of (\ref{net}) are set to 
	one, the (nonnegative) Hessian spectrum of the families of minima 
	considered 
	in 
	\pref{thm:power_series} is 
	\begin{center}
		\begin{tabular}[H]{l|l}
			Eigenvalue & Multiplicity\\\hline
			\rule{0pt}{4ex}
			$O(d^{-1/2})$ & $d$ \\
			\rule{0pt}{3ex}    
			$
			\frac{1}{4} - \frac{1}{2\pi}
			+ O(d^{-1/2})$ & $\frac{(d-1)(d-2)}{2}$ \\
			\rule{0pt}{3ex}    
			$
			\frac{1}{2} - \frac{1}{\pi}
			+ O(d^{-1/2})$ & $d -1$ \\
			\rule{0pt}{3ex}    
			$ \frac{1}{4} + O(d^{-1/2})$ & $d-1$ \\
			\rule{0pt}{3ex}    
			$
			\frac{1}{4} + \frac{1}{2\pi}
			+ O(d^{-1/2})$ & $\frac{d(d-3)}{2}$ \\
			\rule{0pt}{3ex}    
			$ \frac{d}{4} + \frac{1}{2} + O(d^{-1/2})$ & $d-1$ \\
			\rule{0pt}{3ex}    
			$ \frac{d}{4} + \frac{-4 + \pi + \pi^{2}}{2 \pi \left(-4 + 
			\pi\right)} + 
			O(d^{-1/2})$ & 
			$1$ \\
			\rule{0pt}{3ex}    
			$ \frac{d}{\pi} + \frac{- 10 \pi + 8 + \pi^{2}}{2 \pi \left(-4 + 
				\pi\right)} + O(d^{-1/2})$ & $1$. \\
		\end{tabular}	
	\end{center}
\end{theorem}
The method we developed for computing the Hessian spectrum builds on 
\cite{arjevanifield2020hessian}, but differs in three crucial aspects. 
First, in our setting the second layer is trainable. The associated so-called 
\emph{isotypic decomposition} (a set of `simple' subspaces compatible with the 
action of row- and column-permutations formally introduced in 
\pref{sec:hes_spec}) must therefore cover the space 
of $d\times d$ weight matrices, as well as the $d$ weights of the second layer. 
Secondly, instead of expressing the Hessian entries 
in terms of $d, W(d)$ and $\balpha(d)$ and then extracting fractional power 
series for eigenvalues, we directly express the eigenvalues 
in these terms. This considerably simplifies computations and 
facilitates the computation of eigenvalues to potentially any order. Thirdly, 
and perhaps most importantly, 
computing the Hessian to high accuracy 
reveals that for small values of $d$ the families of minima presented in 
\pref{thm:power_series} are in fact saddles. 
Regarding $d$ as a real number, we pinpoint the exact \emph{fractional} 
dimension at which a given family 
of critical points turns 
from saddles into spurious minima (see also \pref{fig:evs}). One is 
now led into the dual question: is there a mechanism by which spurious minima 
may be annihilated 
(that is, turn into saddles)? We have found that the process of the creation 
of minima can be reversed in the over-parameterized regime. This provides a 
strong evidence for the benefits of increasing the number of student neurons 
from an optimization point of view---adding more neurons turns spurious 
minima into saddles, thus encouraging gradient-based 
methods to converge to minima of better quality. A rigorous study of this 
process requires tools from equivariant bifurcation theory, and is deferred 
to future work (cf. \cite{arjevani2021equivariant}).

Quite remarkably, although families of minima presented in 
\pref{thm:power_series} differ significantly, their Hessian spectra coincide 
to $O(d^{-1/2})$-order. Another noticeable implication of \pref{thm:spec} is 
that 
the Hessian spectrum tends to be extremely skewed. Both phenomena, as 
well as their consequences for optimization-related aspects, are discussed 
in detail below .

\paragraph{Positively-skewed Hessian spectral density.} 
Although first reported nearly 30 years ago \cite{bottou1991stochastic}, 
to our knowledge, this is the first time that this 
phenomenon of extremely skewed spectral density has been 
established \emph{rigorously} for two trainable ReLU layers of arbitrarily 
large dimensionality. Early empirical studies of the Hessian spectrum 
\cite{bottou1991stochastic} revealed that local minima tend to be extremely 
ill-conditioned. This intriguing observation was corroborated and further 
refined in a series of works 
\cite{lecun2012efficient,sagun2016eigenvalues,sagun2017empirical}
which studied how the spectrum evolves along the training 
process. It was noticed that, upon convergence, the spectral 
density decomposes into two parts: a bulk of 
eigenvalues concentrated near small positive constants, 
and a set of positive outliers located away from zero.

Due to the high computational cost of an exact computation of 
the Hessian spectrum (cubic in the problem parameters $k$ and $d$), 
this phenomenon of extremely skewed spectral densities has only been confirmed 
for small-scale networks. Other methods for extracting second-order 
information in 
large-scale problems roughly fall into two general categories. The 
first class of 
methods approximate the Hessian spectral density by 
employing various 
numerical estimation techniques, most notably stochastic Lanczos 
method (e.g., \cite{ghorbani2019investigation,papyan2018full}). 
These methods have provided various numerical evidences which 
indicate that a similar skewed spectrum phenomenon also occurs in 
full-scale modern neural networks. The second class of 
techniques builds on tools from random matrix theory. The latter approach 
yields an 
exact computation of the limiting spectral 
distribution (where the number of neurons is taken to infinity),
assuming the inputs, as well as the model weights are drawn at random 
\cite{pennington2017nonlinear,pennington2018spectrum,louart2018random,karakida2019universal}.
In contrast, our method gives an analytic description of the 
spectral density for any number of neurons (granted $d\ge9$), and 
at critical points rather than randomly drawn weight 
matrices.		

\paragraph{The flat minima conjecture and implicit bias.} 
It has long been debated whether some notion of local  curvature 
can be 
used to explain the remarkable generalization capabilities of modern 
neural networks 
\cite{hochreiter1997flat, 
	keskar2016large,jastrzkebski2017three,wu2017towards,yao2018hessian,chaudhari2019entropy,dinh2017sharp}.
One intriguing hypothesis
argues that minima with wider basins of attraction tend to generalize 
better. The suggested intuitive explanation is that 
flatness promotes statistical and numerical stability; 
together with low empirical 
loss, these 
ingredients are widely used to achieve good generalization, cf. 
\cite{shalev2010learnability}. However, the analysis presented in 
\pref{thm:spec} shows that the spectra of global minima and spurious minima 
agree to $O(d^{-1/2})$-order. Consequently, in our setting, local second-order 
curvature \emph{cannot} be used to separate global from spurious 
minima. This rules out any notion of flatness which exclusively relies on the 
Hessian spectrum. Of course, other metrics of a `wideness of basins' may well 
apply. 

\section{A symmetry-based analytic framework}\label{sec:method}
In the sequel, we present the main ingredients of the symmetry-based 
technique developed in this work. The technique builds on, and significantly 
extends various components from  
\cite{arjevanifield2020hessian,arjevanifield2019spurious,ArjevaniField2020}. 
To ease exposition, 
we illustrate with reference to the case where $k=d$ and the weights of 
the second layer are set to one. We let the 
resulting nonconvex function be denoted by $f(W) \defeq \ploss(W,\bones)$. 

\subsection{Power series representation of  minima} \label{sec:pf_power_series}
The first step in deriving a power series representation is to 
restrict the nonconvex function under consideration to \emph{fixed point 
	subspaces} (see \cite{field2007} for a more complete account); spaces that 
consists of 
weight matrices which remain fixed under row- 
and column-permutations in a certain subgroup of $S_d\times S_d$.
For concreteness, consider the space of all matrices 
which are invariant to permutations in $\Delta S_d$, i.e., 
\newcommand{\dspace}{\cW_d}
\begin{align}\label{def:fps}
	\dspace \defeq \{W \in M(d,d)~|~ W = P_\pi WP_{\pi}^\top~ \text{for all 
	}(\pi,\pi)\in \Delta S_d\}.
\end{align}
It is easy to verify that $\dspace$ is in fact a 2-dimensional 
subspace of the form $\dspace = \{(a_1,a_2)~|~a_1 I_d + 
a_2(\bones_d\bones_d^\top 
-I)\}$, and that if $W\in \dspace$, then also $\grad f(W)\in \dspace$ 
(see \cite[Proposition 3]{arjevanifield2019spurious}). Thus, one may 
regard $f$ as a function from $\RR^2$ to $\RR^2$. Next, we make the dependence 
of $f$ on $d$ explicit by defining $F:\RR^3\to \RR^2:~(a_1,a_2,d) \mapsto \grad 
f(a_1 I_d + a_2(\bones_d\bones_d^\top -I_d))$. 
Note that although $d$ was initially taken to be a natural number, in the 
following explicit expressions of $F(a_1,a_2,d)$,  we regard $d$ as a real 
variable
\vspace{0.2cm}\\
\scalebox{0.59}{$
	\begin{pmatrix}
		\frac{a_{1} d}{2} - \frac{a_{1} d \sin{\left(\beta^{(1)}_{(1)} 
				\right)}}{2 \pi 
			\nu_{(1)}} + \frac{a_{1} \left(d^{2} - d\right) 
			\sin{\left(\alpha^{(2)}_{(1)} 
				\right)}}{2 \pi} - \frac{a_{1} \left(d^{2} - d\right) 
			\sin{\left(\beta^{(2)}_{(1)} \right)}}{2 \pi \nu_{(1)}} - 
		\frac{a_{2} 
			\alpha^{(2)}_{(1)} \left(d^{2} - d\right)}{2 \pi} + 
		\frac{a_{2} \left(d^{2} - 
			d\right)}{2} + \frac{\beta^{(1)}_{(1)} d}{2 \pi} - 
		\frac{d}{2}	\\
		- \frac{a_{1} \alpha^{(2)}_{(1)} \left(d^{2} - d\right)}{2 
			\pi} + \frac{a_{1} 
			\left(d^{2} - d\right)}{2} - \frac{a_{2} 
			\alpha^{(2)}_{(1)} \left(d^{3} - 3 
			d^{2} + 2 d\right)}{2 \pi} + \frac{a_{2} \left(d^{3} - 2 
			d^{2} + d\right) 
			\sin{\left(\alpha^{(2)}_{(1)} \right)}}{2 \pi} + 
		\frac{a_{2} \left(d^{3} - 2 
			d^{2} + d\right)}{2} - \frac{a_{2} \left(d^{2} - d\right) 
			\sin{\left(\beta^{(1)}_{(1)} \right)}}{2 \pi \nu_{(1)}} - 
		\frac{a_{2} 
			\left(d^{3} - 2 d^{2} + d\right) 
			\sin{\left(\beta^{(2)}_{(1)} \right)}}{2 \pi 
			\nu_{(1)}} + \frac{\beta^{(2)}_{(1)} \left(d^{2} - 
			d\right)}{2 \pi} - 
		\frac{d^{2}}{2} + \frac{d}{2}
	\end{pmatrix}$\vspace{0.2cm}}
where, here and below, $\alpha_{(i)}^{(j)}$ (resp. 
$\beta_{(i)}^{(j)}$) denotes the angles 
between the $i$th row of $W$ and the $j$th of $W$ (resp. $V$). Using 
the implicit function theorem one can establish the existence as well as the 
uniqueness of a path of critical points $(a_1(d),a_2(d))$. This gives a 
formal meaning for examining a given family of minima at a \emph{fractional} 
dimensionality. Formalities are covered in length in 
\cite{ArjevaniField2020}.

We are left with the following two issues. First, the implicit 
function theorem does not yield an explicit form of $a_1(d)$ and $a_2(d)$. This 
issue is addressed by using a \emph{real analytic} version of the implicit 
function theorem, and then computing the coefficients of the power 
series of $a_1(d)$ and $a_2(d)$ to a desired order. Secondly, the implicit 
function theorem requires initial values for $a_1$ and $a_2$ which solve 
$F(a_1,a_2,d) =0$. To handle this issue, we form power series in $1/d$ and, 
loosely speaking, develop them at $d=\infty$. The advantage of this 
nonstandard manipulation is that the limiting entries of $W(d)$ have a very 
simple form, which are then used as the initial values required for invoking 
the implicit function theorem. In fact, due to the dependence of $\grad f$ on 
the angels between rows of $W$ and $V$, the power series of $a_1(d)$ and 
$a_2(d)$ are expressed in terms of $1/\sqrt{d}$. \pref{thm:power_series} 
is established by following the recipe described here for various families of 
critical points of different isotropy. The gradient expressions involved in 
the process are lengthy and are therefore relegated to 
\pref{sec:eig_trans}.

\subsection{Computing the Hessian spectrum} 
\label{sec:compute_hess}
Our next goal is to derive an analytic characterization of the Hessian 
spectrum. The invariance properties of $\ploss$ imply that although the 
dimension of the Hessian, $k(d+1)\times k(d+1)$, depends on $k$ and $d$, the 
number of distinct eigenvalues remains fixed. This is an immediate consequence 
of the respective isotypic decomposition (see \pref{sec:hes_spec}). A formal 
introduction of the isotypic decomposition requires some familiarity 
with group actions, and is therefore deferred to \pref{sec:spec}. Here, we 
shall focus on one particularly simple case.

The isotypic decomposition associated with $\Delta S_d$, the isotropy 
of $I_d$, implies that for any $d$,
\newcommand{\ev}{\overline{\fY_d}} 
\begin{align*}
	\ev \defeq 
	\vect\prn*{
		\begin{bmatrix}
			0&d-3&3-d&\cdots&0&0&0\\
			d-3&0&0&\cdots&-1&-1&-1\\
			3-d&0&0&\cdots&1&1&1\\
			0&-1&1&\cdots&0&0&0\\
			\cdots&\cdots&\cdots&\cdots&\cdots&\cdots\\
			0&-1&1&\cdots&0&0&0\\
			0&-1&1&\cdots&0&0&0\\
	\end{bmatrix}},
\end{align*}
is an eigenvector of $\nabla^2 f(I_d)$, where $\vect$ denotes the linear 
transformation which stacks the rows of a given matrix on top of one another 
as a column vector. Thus, the eigenvalue corresponding to $\ev$ equals 
$(\nabla^2 f(I_d) \ev)_2/(d-3)$, which takes the following explicit form 
\vspace{0.2cm}\\
\scalebox{0.63}{$
	\frac{a_{1}^{2}}{2 \pi \nu_{(1)}^{2} \nu^{(2)}_{(1)}} + \frac{a_{1}^{2} 
		\sin{\left(\alpha^{(2)}_{(1)} \right)}}{2 \pi \nu_{(1)}^{2} 
		\left(\nu^{(2)}_{(1)}\right)^{2}} - 
	\frac{a_{1} a_{2}}{\pi \nu_{(1)}^{2} \nu^{(2)}_{(1)}} - \frac{a_{1} a_{2} 
	\sin{\left(\alpha^{(2)}_{(1)} 
			\right)}}{\pi \nu_{(1)}^{2} \left(\nu^{(2)}_{(1)}\right)^{2}} + 
			\frac{a_{2}^{2}}{2 \pi \nu_{(1)}^{2} 
		\nu^{(2)}_{(1)}} + \frac{a_{2}^{2} \sin{\left(\alpha^{(2)}_{(1)} 
		\right)}}{2 \pi \nu_{(1)}^{2} 
		\left(\nu^{(2)}_{(1)}\right)^{2}}+ \frac{\alpha^{(2)}_{(1)}}{2 \pi} + 
	\frac{\left(d - 1\right) 
		\sin{\left(\alpha^{(2)}_{(1)} \right)}}{2 \pi} - \frac{\left(d - 
		1\right) \sin{\left(\beta^{(2)}_{(1)} 
			\right)}}{2 \pi \nu_{(1)}} - \frac{\sin{\left(\beta^{(1)}_{(1)} 
			\right)}}{2 \pi \nu_{(1)}} - 
	\frac{\sin{\left(\beta^{(2)}_{(1)} \right)}}{2 \pi 
		\left(\mu^{(2)}_{(1)}\right)^{2} \nu_{(1)}},$}\\\vspace{0.3cm}
where $\nu_{(i)}$ denotes the norm of the $i$th row of $W$, and 
$\nu_{(i)}^{(j)}$ (resp. $\mu_{(i)}^{(j)}$) denotes 
$\sin\arccos(\alpha_{(i)}^{(j)})$ (resp. $\sin\arccos(\beta_{(i)}^{(j)})$).
To complete the derivation of the eigenvalue $\ev$ for families of isotropy 
$\Delta S_d$, e.g., cases 1(a) and 1(b) in \pref{thm:power_series}, one 
plugs-in the power series representation of $a_1(d)$ and $a_2(d)$ into the 
expression above, and forms a new power series to get an estimate for $\ev$. 
The same  procedure is used for all families of minima presented in 
\pref{thm:power_series}, and is described in detail in~\pref{sec:compute_hess}.

\section{Conclusion}
When present, the principle of symmetry breaking forms a powerful tool of 
reducing the complexity of nonconvex problems in a nonlinear way. 
This yields unprecedented analytic information for ReLU networks with a finite 
number of inputs and neurons (rather than limiting models of varying 
qualities). 
At this stage of the theory, the extent to which the symmetry breaking 
principle applies is not well-understood. However, it does seem to hold in a 
broader class of fundamental nonconvex problems, e.g., 
\cite{arjevanifield2021tensor}.  

An intriguing finding allowed by the principle of symmetry breaking is that 
various families of spurious and global minima are locally identical to 
$O(d^{-1/2})$-order. 
Since the multiplicity of local minima for $k=d$ is 
essentially exponential in $d$, it is rather unlikely for SGD to be able to 
escape spurious minima when $d$ grows 
(see \cite[Table 1]{safran2017spurious}). Why is it then that SGD is generally 
able to find good minima for optimization problems associated with fitting ReLU 
neural networks? 
Preliminary experiments we conducted indicate that \emph{continuously} 
increasing $k$ and leaving $d$ fixed turns families of spurious minima turn 
into saddles. (The objective value seems to remain constant.) 
We believe that this phenomenon is instrumental for explaining the 
effectiveness of gradient-based methods, SGD in particular. A thorough study 
of this requires tools from equivariant bifurcation theory and is 
postponed to future work.

\section*{Acknowledgements}
Part of this work was completed while YA was a postdoctoral researcher at NYU. 
We thank Joan Bruna, Ohad Shamir, Shimon Shpiz and Daniel Soudry for valuable 
discussions.

\bibliographystyle{plain}
\bibliography{bib}

\newpage
\appendix
\input{A_sym_break}

\input{A_gradient_expressions}

\input{appendix}

\end{document}

%% file: arxiv_style.tex
\usepackage{breakcites}
\usepackage[dvipsnames]{xcolor}
\usepackage[hidelinks]{hyperref}
\hypersetup{
	colorlinks=true,
	linkcolor=blue!70!black,
	citecolor=blue!70!black,
	urlcolor=blue!70!black}

\usepackage{microtype,bm}
\usepackage{xspace}
\usepackage{algorithm}
\usepackage{verbatim}
\usepackage[noend]{algpseudocode}
\usepackage{mathtools}
\usepackage{amsmath}
\usepackage{bbm}
\usepackage{amsfonts}
\usepackage{amssymb}
\usepackage{xpatch}
\usepackage{enumitem}
\usepackage{amsthm}

\usepackage[nointegrals]{wasysym}

\theoremstyle{definition}  

\newtheorem{fact}{Fact}

\theoremstyle{plain}

\newtheorem{theorem}{Theorem}
\newtheorem*{theorem*}{Theorem}

\newtheorem{lemma}{Lemma}

\xpatchcmd{\proof}{\itshape}{\normalfont\proofnameformat}{}{}
\newcommand{\proofnameformat}{\bfseries}

\usepackage{prettyref}
\newcommand{\pref}[1]{\prettyref{#1}}

\newcommand{\savehyperref}[2]{\texorpdfstring{\hyperref[#1]{#2}}{#2}}
\newrefformat{eq}{\savehyperref{#1}{\textup{(\ref*{#1})}}}
\newrefformat{eqn}{\savehyperref{#1}{Equation~\ref*{#1}}}
\newrefformat{lem}{\savehyperref{#1}{Lemma~\ref*{#1}}}
\newrefformat{obs}{\savehyperref{#1}{Observation~\ref*{#1}}}
\newrefformat{def}{\savehyperref{#1}{Definition~\ref*{#1}}}
\newrefformat{line}{\savehyperref{#1}{line~\ref*{#1}}}
\newrefformat{thm}{\savehyperref{#1}{Theorem~\ref*{#1}}}
\newrefformat{corr}{\savehyperref{#1}{Corollary~\ref*{#1}}}
\newrefformat{sec}{\savehyperref{#1}{Section~\ref*{#1}}}
\newrefformat{exam}{\savehyperref{#1}{Example~\ref*{#1}}}
\newrefformat{prop}{\savehyperref{#1}{Proposition~\ref*{#1}}}
\newrefformat{table}{\savehyperref{#1}{Table~\ref*{#1}}}
\newrefformat{app}{\savehyperref{#1}{Appendix~\ref*{#1}}}
\newrefformat{ass}{\savehyperref{#1}{Assumption~\ref*{#1}}}
\newrefformat{tbl}{\savehyperref{#1}{Table~\ref*{#1}}}
\newrefformat{ex}{\savehyperref{#1}{Example~\ref*{#1}}}
\newrefformat{fig}{\savehyperref{#1}{Figure~\ref*{#1}}}
\newrefformat{alg}{\savehyperref{#1}{Algorithm~\ref*{#1}}}
\newrefformat{rem}{\savehyperref{#1}{Remark~\ref*{#1}}}
\newrefformat{conj}{\savehyperref{#1}{Conjecture~\ref*{#1}}}
\newrefformat{prop}{\savehyperref{#1}{Proposition~\ref*{#1}}}
\newrefformat{proto}{\savehyperref{#1}{Protocol~\ref*{#1}}}
\newrefformat{prob}{\savehyperref{#1}{Problem~\ref*{#1}}}
\newrefformat{claim}{\savehyperref{#1}{Claim~\ref*{#1}}}

%% file: macros.tex
\def\ddefloop#1{\ifx\ddefloop#1\else\ddef{#1}\expandafter\ddefloop\fi}
\def\ddef#1{\expandafter\def\csname 
	bb#1\endcsname{\ensuremath{\mathbb{#1}}}}
\ddefloop ABCDEFGHIJKLMNOPQRSTUVWXYZ\ddefloop
\def\ddef#1{\expandafter\def\csname 
	#1\endcsname{\ensuremath{{\bm{#1}}}}}
\ddefloop ABCDEFGHIJKLMNOPQRSTUVWXYZ\ddefloop
\def\ddef#1{\expandafter\def\csname 
	#1\endcsname{\ensuremath{{\bm{#1}}}}}
\ddefloop abcdefghijklmnopqrstuvwxyz\ddefloop
\def\ddefloop#1{\ifx\ddefloop#1\else\ddef{#1}\expandafter\ddefloop\fi}
\def\ddef#1{\expandafter\def\csname 
	b#1\endcsname{\ensuremath{\mathbf{#1}}}}
\ddefloop ABCDEFGHIJKLMNOPQRSTUVWXYZ\ddefloop
\def\ddef#1{\expandafter\def\csname 
	c#1\endcsname{\ensuremath{\mathcal{#1}}}}
\ddefloop ABCDEFGHIJKLMNOPQRSTUVWXYZ\ddefloop
\def\ddef#1{\expandafter\def\csname 
	f#1\endcsname{\ensuremath{\mathfrak{#1}}}}
\ddefloop ABCDEFGHIJKLMNOPQRSTUVWXYZ\ddefloop
\def\ddef#1{\expandafter\def\csname 
	h#1\endcsname{\ensuremath{\widehat{#1}}}}
\ddefloop ABCDEFGHIJKLMNOPQRSTUVWXYZ\ddefloop
\def\ddef#1{\expandafter\def\csname 
	hc#1\endcsname{\ensuremath{\widehat{\mathcal{#1}}}}}
\ddefloop ABCDEFGHIJKLMNOPQRSTUVWXYZ\ddefloop
\def\ddef#1{\expandafter\def\csname 
	t#1\endcsname{\ensuremath{\widetilde{#1}}}}
\ddefloop ABCDEFGHIJKLMNOPQRSTUVWXYZ\ddefloop
\def\ddef#1{\expandafter\def\csname 
	tc#1\endcsname{\ensuremath{\widetilde{\mathcal{#1}}}}}
\ddefloop ABCDEFGHIJKLMNOPQRSTUVWXYZ\ddefloop

\newcommand{\EE}{\mathbb{E}}
\newcommand{\RR}{\mathbb{R}}

\newcommand{\balpha}{\boldsymbol{\alpha}}

\newcommand{\bbeta}{\boldsymbol{\eta}}

\newcommand{\bzeros}{{\bm{0}}}
\newcommand{\bones}{\bm{1}}

\newcommand{\inner}[1]{\langle #1 \rangle}

\renewenvironment{proof}{\par\noindent{\bf Proof\ }}{\hfill\BlackBox\\[2mm]}
\newcommand{\BlackBox}{\rule{1.5ex}{1.5ex}}  
\newcommand{\vect}{\mathrm{vec}}

\newcommand{\overbar}[1]{\mkern 
1.5mu\overline{\mkern-1.5mu#1\mkern-1.5mu}\mkern 1.5mu}

\newcommand{\todoc}[1]{\ignorespaces}

\newcommand{\Diag}[1]{\text{Diag}\prn*{#1}}

\newcommand{\defeq}{\coloneqq}

\DeclarePairedDelimiter{\brk}{[}{]}
\DeclarePairedDelimiter{\crl}{\{}{\}}
\DeclarePairedDelimiter{\prn}{(}{)}

\usepackage{nicefrac}

\newsavebox\CBox

\renewcommand{\theequation}{\thesection.\arabic{equation}}

\newcommand{\grad}[1]{{\nabla {#1}}}

\newcommand{\dd}{\mbox{$\;|\;$}}

\newcommand{\real}{{\mathbb{R}}}

\newcommand{\On}[1]{{\text{\rm O}({#1})}}
\newcommand{\ON}{\On{d}}

\newcommand{\defoo}{~\dot{=}~}

\newcommand{\pint}{\mbox{$\mathbb{N}$}}

\newcommand{\arr}{{\rightarrow}}

\newtheorem{thm}[lemma]{Theorem}

\theoremstyle{definition}

\newtheorem{exam}[lemma]{Example}
\newtheorem{exams}[lemma]{Examples}

\theoremstyle{remark}

%% file: nn_symmetry_tikz.tex
\usepackage{tikz}					
\usepackage{ifthen}

\tikzset{%
	box/.style={
		align=center, minimum size=1cm, inner 
		sep=0pt, font=\tiny\bfseries,
		draw=black, #1
	},
	c1/.style={fill=yellow},
	c2/.style={fill=blue},
	c3/.style={fill=green},
	c3/.style={fill=red},
}

\newcommand{\blockmatrix}[9]{
	\draw[draw=#4,fill=#5] (0,0) rectangle( #1,#2);
	\ifthenelse{\equal{#6}{true}}
	{
		\draw[draw=#7,fill=#8] (0,#2) -- (#9,#2) -- ( #1,#9) -- ( #1,0) -- ( #1 
		- #9,0) -- (0,#2 -#9) -- cycle;
	}
	{}
	\draw ( #1/2, #2/2) node { #3};
}

\newcommand{\mblockmatrix}[4][none]{
	\begin{tikzpicture} 
	\ifthenelse{\equal{#1}{none}}
	{
		\blockmatrix{#2}{#3}{#4}{none}{none}{false}{none}{none}{0.0}
	}
	{
		\definecolor{fillcolor}{rgb}{#1}
		\blockmatrix{#2}{#3}{#4}{none}{fillcolor}{false}{none}{none}{0.0}
	}
	\end{tikzpicture}
}

\newcommand{\fblockmatrix}[4][none]{
	\begin{tikzpicture} 
	\ifthenelse{\equal{#1}{none}}
	{
		\blockmatrix{#2}{#3}{#4}{black}{none}{false}{none}{none}{0.0}
	}
	{
		\definecolor{fillcolor}{rgb}{#1}
		\blockmatrix{#2}{#3}{#4}{black}{fillcolor}{false}{none}{none}{0.0}
	}
	\end{tikzpicture}
}

\newcommand{\dblockmatrixSD}[6][none]{
	\begin{tikzpicture} 
	\draw[draw=black,fill=none] (0,0) rectangle( #2,#3);
	\definecolor{fillcolor}{rgb}{#1};
	\draw[draw=black,fill=fillcolor] (0,#3) -- (#5,#3) -- ( #2,#5) -- ( #2,0) 
	-- ( #2 - #5,0) -- (0,#3 -#5) -- cycle;
	\draw ( #2/2, #3/2) node { #4};
	\draw ( #2*0.25, #3*0.25) node {#6};
	\draw ( #2*0.75, #3*0.75) node { #6};
	\end{tikzpicture}
}

\newcommand{\dblockmatrixSDMO}[7][none]{
	\begin{tikzpicture} 
	\draw[draw=black,fill=none] (0,0) rectangle( #2,#3);
	
	\definecolor{upperleft}{rgb}{#1};
	\draw[draw=black,fill=upperleft] (0,#3) -- (#5,#3) -- ( #2- #7,#5+#7) -- ( 
	#2 -  #7,#7) -- ( #2 - #5 - #7 ,#7) -- (0,#3 -#5) -- cycle;
	
	\definecolor{lowerleft}{rgb}{1,1,0.8};
	\draw[draw=black,fill=lowerleft] (0,0) rectangle (#2-#7,#7);
	
	\definecolor{upperright}{rgb}{0.8,1,0.8};
	\draw[draw=black,fill=upperright] (#2-#7,#7) rectangle (#2,#3);
	
	\definecolor{lowerright}{rgb}{0.8,0.8,1};
	\draw[draw=black,fill=lowerright] (#2-#7,0) rectangle (#2,#7);
	\draw ( #2*0.5-#7*0.5, #3*0.5+#7*0.5) node { #4};
	\draw ( #2*0.25-#7*0.25, #3*0.25 + #7*0.75) node {#6};
	\draw ( #2*0.75 - #7*0.75, #3*0.75 + #7*0.25) node { #6};
	\draw ( #2*0.5 - #7*0.5, #7*0.5) node {$\gamma$};
	\draw ( #2 - #7*0.5, #3*0.5+ #7*0.5) node {$\delta$};
	\draw ( #2 - #7*0.5, #7*0.5) node {$\epsilon$};
	\end{tikzpicture}
}

\newcommand{\dblockmatrixSDMOOO}[7][none]{
	\begin{tikzpicture} 
	\draw[draw=black,fill=none] (0,0) rectangle( #2,#3);
	
	\definecolor{upperleft}{rgb}{#1};
	\draw[draw=black,fill=upperleft] (0,#3) -- (#5,#3) -- ( #2- #7,#5+#7) -- ( 
	#2 -  #7,#7) -- ( #2 - #5 - #7 ,#7) -- (0,#3 -#5) -- cycle;
	
	\definecolor{lowerleft}{rgb}{1,1,0.8};
	\draw[draw=black,fill=lowerleft] (0,0) rectangle (#2-#7,#7);
	
	\definecolor{upperright}{rgb}{0.8,1,0.8};
	\draw[draw=black,fill=upperright] (#2-#7,#7) rectangle (#2,#3);
	
	\definecolor{lowerright}{rgb}{0.8,0.8,1};
	\draw[draw=black,fill=lowerright] (#2-#7,0) rectangle (#2,#7);
	\draw ( #2*0.5-#7*0.5, #3*0.5+#7*0.5) node { #4};
	\draw ( #2*0.25-#7*0.25, #3*0.25 + #7*0.75) node {#6};
	\draw ( #2*0.75 - #7*0.75, #3*0.75 + #7*0.25) node { #6};
	\draw ( #2*0.5 - #7*0.5, #7*0.5) node {$\gamma$};
	\draw ( #2 - #7*0.5, #3*0.5+ #7*0.5) node {$\delta$};
	\definecolor{lowerdia}{rgb}{0.8,0.8,0.8};
	\draw[draw=black,fill=lowerdia] (#2-#7,#7) -- (#2-#7+#5,#7) -- ( #2,#5) 
	-- ( 
	#2 ,0) -- ( #2 - #5  ,0) -- (#2-#7,#7 -#5) -- cycle;
	\draw ( #2 - #7*0.5, #7*0.5) node {$\epsilon$};
	\draw ( #2-#7*0.25 , #7*0.75) node {$\zeta$};
	\draw ( #2 - #7*0.75 , #7*0.25) node { $\zeta$};
	\end{tikzpicture}
	
}

\newcommand{\diagonalblockmatrix}[5][none]{
	\begin{tikzpicture} 
	
	\ifthenelse{\equal{#1}{none}}
	{
		\blockmatrix{#2}{#3}{#4}{black}{none}{true}{black}{none}{#5}
	}
	{
		\definecolor{fillcolor}{rgb}{#1}
		\blockmatrix{#2}{#3}{#4}{black}{none}{true}{black}{fillcolor}{#5}
	}
	
	\end{tikzpicture}
}

\newcommand{\valignbox}[1]{
	\vtop{\null\hbox{#1}}
}

\newenvironment{blockmatrixtabular}
{
	\begin{tabular}{
			@{}c@{}c@{}c@{}c@{}c@{}c@{}c@{}l@{}l@{}l@{}l@{}l@{}l@{}l@{}l@{}l@{}l@{}l@{}l
			@{}c@{}c@{}c@{}l@{}l@{}l@{}l@{}l@{}l@{}l@{}l@{}l@{}l@{}l@{}l@{}l@{}l@{}l@{}l
			@{}l@{}l@{}l@{}l@{}l@{}l@{}l@{}l@{}l@{}l@{}l@{}l@{}l@{}l@{}l@{}l@{}l@{}l@{}l
			@{}
		}
	}
	{
	\end{tabular}
}


%% file: A_sym_break.tex
\section{The principle of symmetry breaking} \label{sec:sym_prin}

\subsection{$S_k\times S_d$-invariance properties of 
(\ref{opt:problem})}\label{sec:inv_properties}
\newcommand{\jloss}{\overline{\ploss}}
\renewcommand{\bbeta}{\boldsymbol{\beta}}
We show that $\ploss$ is $S_k \times S_d$-invariant in its first parameter. The 
proof is a straightforward adaption of \cite[Section 
4.1]{arjevanifield2019spurious}. First, we make the 
dependence of $\ploss$ on the target weight $d\times 
d$-matrix $V$ explicit:
\begin{align}\label{def:jloss}
	\jloss(W, \balpha; V,\bbeta) \defeq \frac{1}{2}\EE_{\x\sim 
		\cN(\bzeros,I_d)}\brk*{\Big(\sum_{i=1}^k \alpha_i 
		\varphi(\inner{\w_i,\x}) 
		- \sum_{i=1}^d \beta_i 	\varphi(\inner{\v_i,\x})\Big)^2},
\end{align}
Next, we observe that for any $\pi\in S_k, \rho \in S_d$ and 
$U\in \ON$, the group of all $d\times d$-orthogonal matrices, we have
\begin{align}
	\jloss(W, \balpha; V,\bbeta) &= 
	\jloss(P_\pi W, \balpha; V,\bbeta)
	=	\jloss(W, \balpha; P_\rho V,\bbeta),\label{property_one}\\
	\jloss(W, \balpha; V,\bbeta) &= 
	\jloss(WU, \balpha; V U,\bbeta)\label{property_two},
\end{align}
where the last equality follows by the $\ON$-invariance of the standard 
multivariate Gaussian distribution. Therefore, for any $\rho \in S_d$ and 
$U\in\ON$ such that $V = P_\rho VU^\top$ and any $\pi \in S_k$, we have
\begin{align*} 
	\jloss(W,\balpha,V,\bbeta) &= \jloss(W,\balpha,P_\rho VU^\top,\bbeta) 
	\stackrel{(\ref{property_one})}{=} 
	\jloss(W,\balpha,VU^\top,\bbeta) \stackrel{(\ref{property_two})}{=}  
	\jloss(WU, \balpha, VU^\top U, \bbeta) \\
	&=  \jloss(WU, \balpha, V, 
	\bbeta)\stackrel{(\ref{property_one})}{=} 
	\jloss(P_\pi WU, \balpha, V, \bbeta).
\end{align*}
In particular, for $V=I_{d}$, we have $V=P_\pi V P_\pi^\top$ for any 
$\pi \in S_d$, thus
$\ploss(W,\balpha)=\jloss(W,\balpha,I_d,\bbeta)$ is  
$S_k\times S_d$-invariant w.r.t. $W$. Note that here we do not exploit the 
rotational invariance of the standard Gaussian distribution, but 
rather its invariance to permutations. Hence, the same 
$S_k\times S_d$-invariance holds for any product distribution if 
$V=I_d$. Indeed, critical points admit maximal isotropy types also when the 
input distribution is $\cD = \cU({[-1,1]^d})$ (but not when $\cD = 
\cU({[0,2]^d})$).

\subsection{Examples of minima for Problem 
(\ref{opt:problem})}\label{sec:num_exp}
We display  several examples for optimization problem (\ref{opt:problem}) with 
$k=d=10$ obtained by running SGD until the gradient norm is driven below 
$1\mathrm{e-}8$.

\begin{figure}[H]
\centering
\includegraphics[scale=0.4]{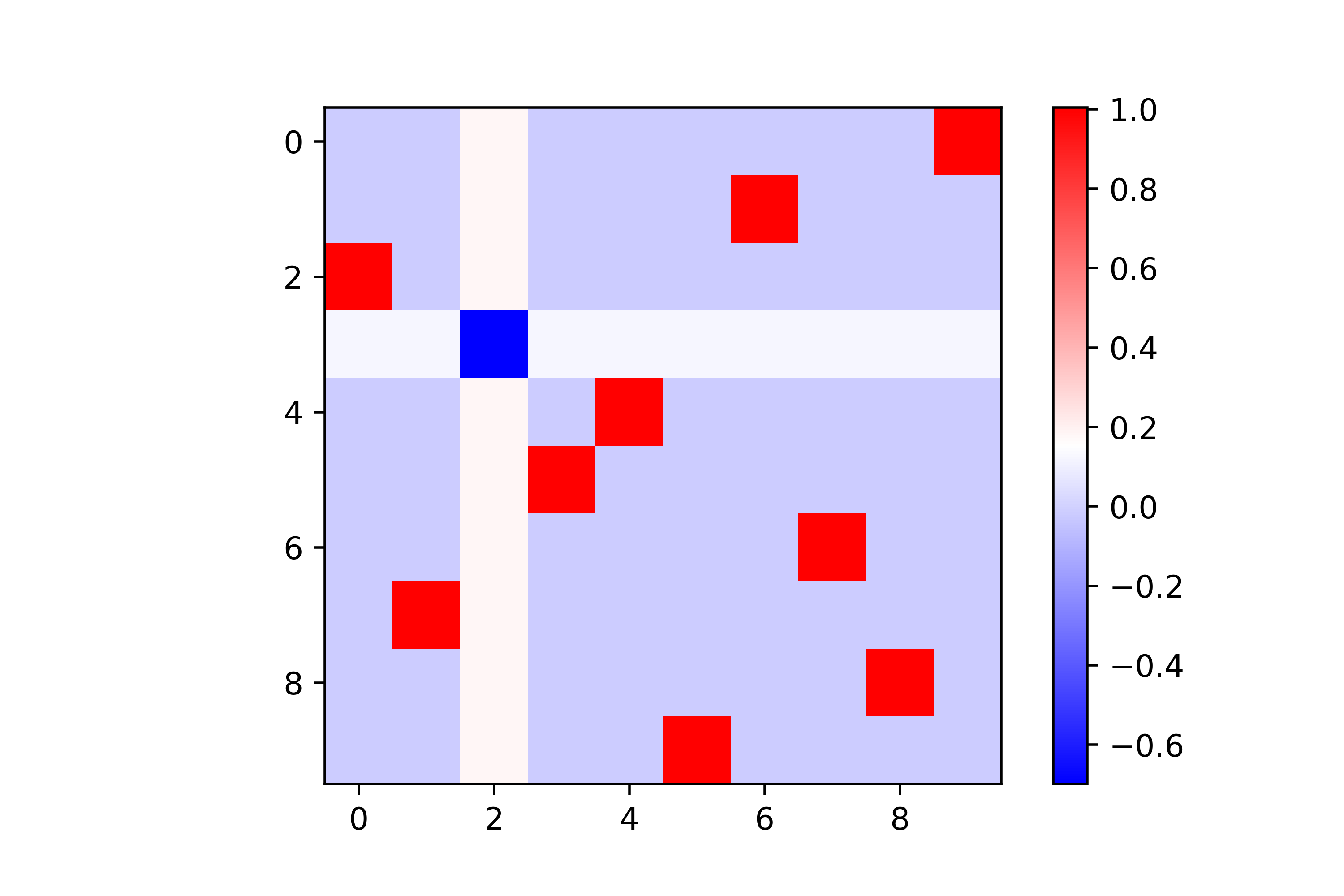}
\caption{A spurious minimum of isotropy (conjugated to) $\Delta 
(S_{d-1}\times S_1)$ of (\ref{opt:problem}) with $k=d=10$. The 
objective value is $\approx 0.018$.}
\end{figure}

\begin{figure}[H]
	\centering
	\includegraphics[scale=0.4]{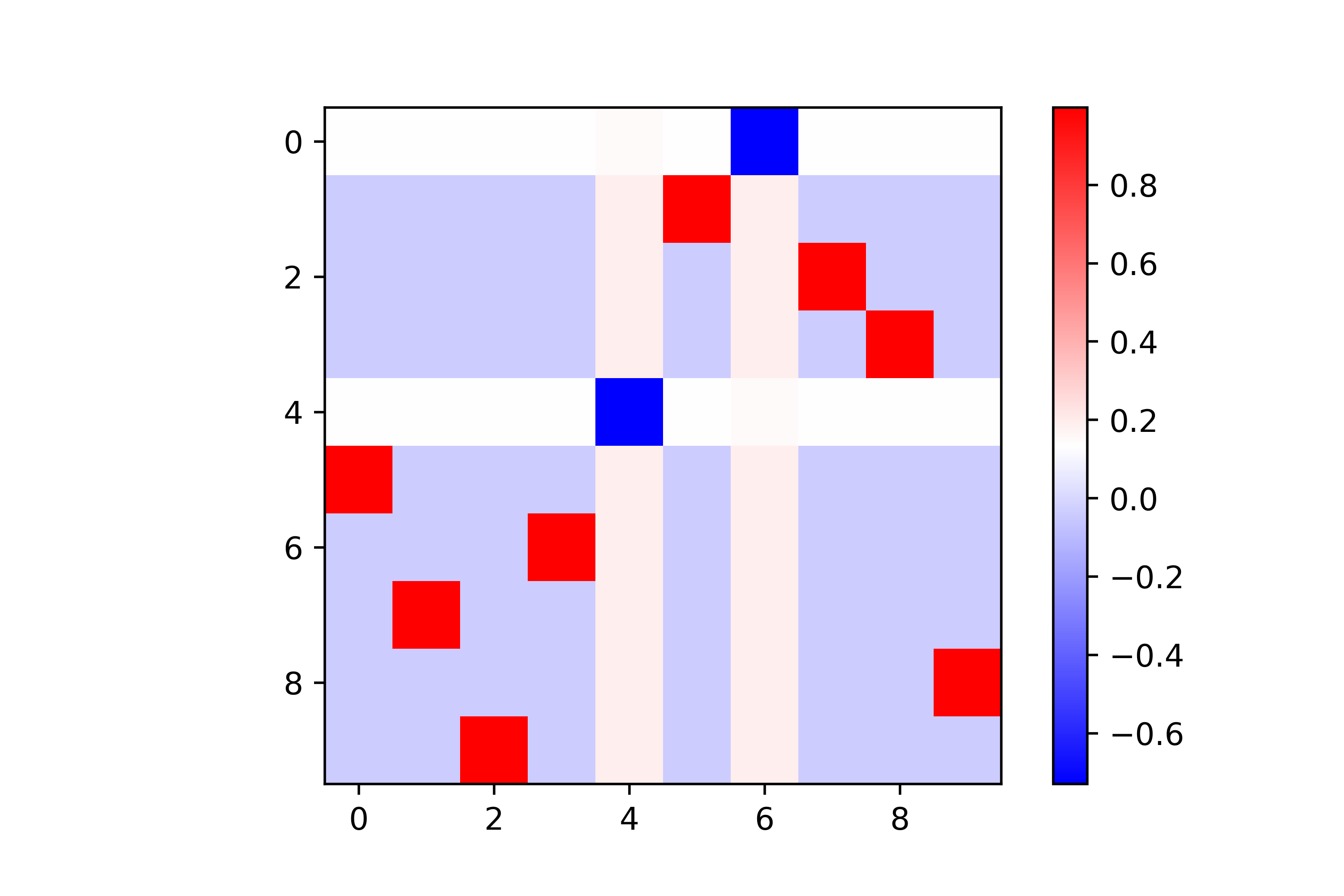}
	\caption{A spurious minimum of isotropy (conjugated to) 
	$\Delta (S_{d-2}\times S_2)$ of (\ref{opt:problem}) with $k=d=10$. The 
	objective value is $\approx 0.035$.}
\end{figure}

\begin{figure}[H]
	\centering
	\includegraphics[scale=0.4]{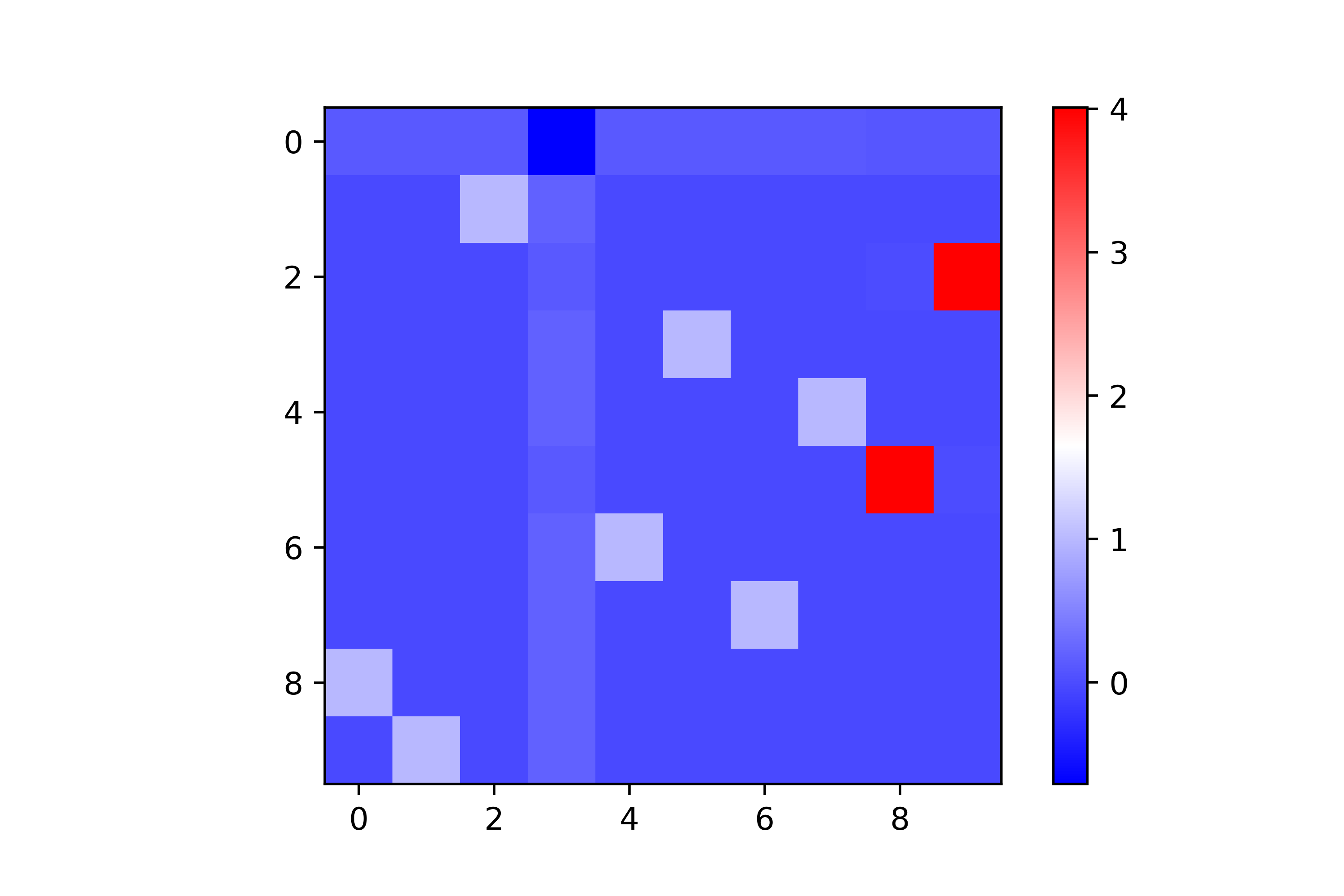}
	\caption{A spurious minimum of (\ref{opt:problem}) with $k=d=10$, where  
	$V=\Diag{1,\dots,1,2,2}$ and $\beta=(1,\dots,1,2,2)$. The symmetry of the 
	minimum	adapt to that of the global minimizer $V$.}
\end{figure}

\begin{figure}[H]
	\centering
	\includegraphics[scale=0.4]{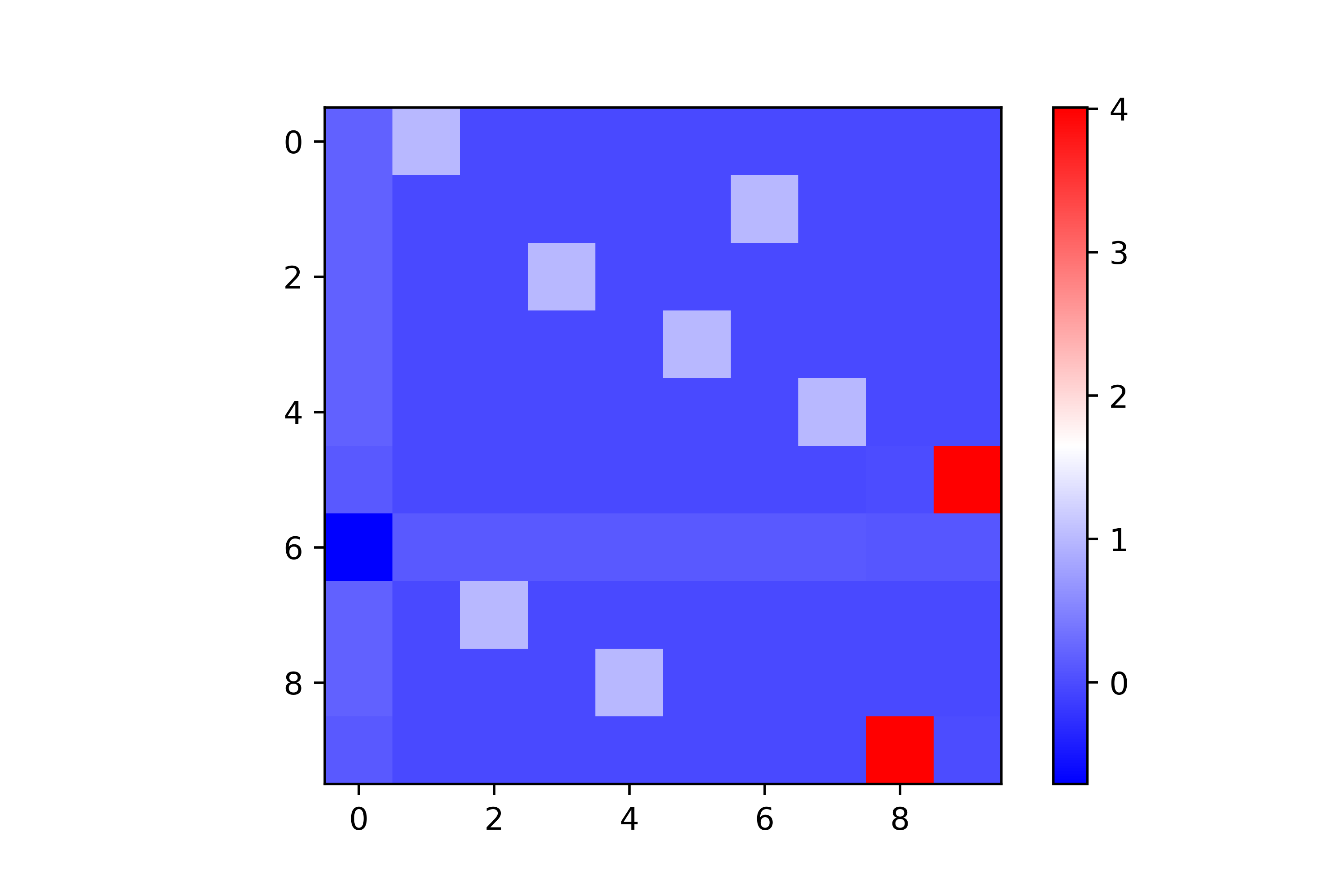}
	\caption{A spurious minimum of (\ref{opt:problem}) with $k=d=10$, where the 
	$V=\Diag{1,\dots,1,2,2}$ and $\beta=(1,\dots,1,2,2)$. The symmetry of the 
	minimum
	adapt to that of the global minimizer $V$.}
\end{figure}

\section{Counting multiplicity of families of minima} \label{sec:mul}
The computation of the multiplicity of minima is based on the orbit-stabilizer 
theorem. Instantiating this theorem to the natural action of $S_k\times S_d$ (i.e., row- and 
column- permutation. See Section \ref{sec:group_action} below for a formal 
introduction of group action) 
yields
\begin{align*}
\mathrm{Multiplicity}(W) = \frac{|S_k\times S_d|}{|\iso{W}|}.
\end{align*}
Observing that $|S_k\times S_d|=d!k!,~~|\Delta S_d| = d!,~~|\Delta 
(S_{d-1}\times S_1)| = 
(d-1)!,~~|\Delta (S_{d-2}\times S_2)| = (d-2)!2!$ and $|\Delta (S_{d-3}\times S_3)| = (d-3)!3!$ 
gives the multiplicities stated in \pref{thm:power_series}.

%% file: A_gradient_expressions.tex

\section{Gradient expressions}
In the sequel, we provide explicit expressions for $\grad \ploss$ 
(defined in (\ref{opt:problem})) restricted the fixed 
point spaces
$\cW_{d-1}, \cW_{d-2}, \cW_{d-3}$, which naturally extend 
\pref{def:fps} as follows
\begin{align*}
\cW_{p} \defeq \{W \in M(d,d)~|~ W = P_\pi WP_{\pi}^\top~ \text{for all 
}(\pi,\pi)\in \Delta (S_{d-p}\times S_p)\}.
\end{align*}
The gradient expressions corresponding to 
$\cW_d$ are given in the body of the paper in \pref{sec:pf_power_series}. 

In all expressions below:
\begin{itemize}
	\item $\alpha_{(i)}^{(j)}$ (resp. 
	$\beta_{(i)}^{(j)}$) denotes the angles 
	between the $i$th row of $W$ and the $j$th of $W$ (resp. $V$, the target 
	weight matrix). 
	\item $\nu_{(i)}$ (resp. $\mu_{(i)}$) denotes the norm of the $i$th row of $W$ 
	(resp. $V$).
	\item $\nu_{(i)}^{(j)}$ (resp. $\mu_{(i)}^{(j)}$) denotes 
	$\sin\arccos(\alpha_{(i)}^{(j)})$ (resp. $\sin\arccos(\beta_{(i)}^{(j)})$).
\end{itemize} 
In \pref{sec:power_series_m5o2}, we list the coefficients 
of the families of minima considered in 
\pref{thm:power_series} to $O(d^{-5/2})$-order.

\subsection{Gradient expressions for $\cW_1$}
The space $\cW_1$ is five-dimensional. A weight 
matrix for $d=8$ can be parameterized as follows 
\begin{align*}
	\left[\begin{matrix}a_{1} & a_{2} & a_{2} & a_{2} & a_{2} & a_{2} & a_{2} & 
		a_{3}\\a_{2} & a_{1} & a_{2} & a_{2} & a_{2} & a_{2} & a_{2} & 
		a_{3}\\a_{2} & 
		a_{2} & a_{1} & a_{2} & a_{2} & a_{2} & a_{2} & a_{3}\\a_{2} & a_{2} & 
		a_{2} & 
		a_{1} & a_{2} & a_{2} & a_{2} & a_{3}\\a_{2} & a_{2} & a_{2} & a_{2} & 
		a_{1} & 
		a_{2} & a_{2} & a_{3}\\a_{2} & a_{2} & a_{2} & a_{2} & a_{2} & a_{1} & 
		a_{2} & 
		a_{3}\\a_{2} & a_{2} & a_{2} & a_{2} & a_{2} & a_{2} & a_{1} & 
		a_{3}\\a_{4} & 
		a_{4} & a_{4} & a_{4} & a_{4} & a_{4} & a_{4} & a_{5}\end{matrix}\right].
\end{align*}
The gradient entries, denoted by $g_1, g_2, g_3, g_4$ and $g_5$, are:
\begin{dmath*}
	{g_1 =}\frac{a_{1} \left(d - 1\right)}{2} + \frac{a_{1} \left(d^{2} - 3 d + 
		2\right) 
		\sin{\left(\alpha^{(2)}_{(1)} \right)}}{2 \pi} + \frac{a_{1} \nu_{(d)} 
		\left(d 
		- 1\right) \sin{\left(\alpha^{(d)}_{(1)} \right)}}{2 \pi \nu_{(1)}} - 
	\frac{a_{1} \left(d - 1\right) \sin{\left(\beta^{(1)}_{(1)} \right)}}{2 \pi 
		\nu_{(1)}} - \frac{a_{1} \left(d - 1\right) 
		\sin{\left(\beta^{(d)}_{(1)} 
			\right)}}{2 \pi \nu_{(1)}} - \frac{a_{1} \left(d^{2} - 3 d + 
			2\right) 
		\sin{\left(\beta^{(2)}_{(1)} \right)}}{2 \pi \nu_{(1)}} - \frac{a_{2} 
		\alpha^{(2)}_{(1)} \left(d^{2} - 3 d + 2\right)}{2 \pi} + \frac{a_{2} 
		\left(d^{2} - 3 d + 2\right)}{2} - \frac{a_{4} \alpha^{(d)}_{(1)} 
		\left(d - 
		1\right)}{2 \pi} + \frac{a_{4} \left(d - 1\right)}{2} + 
	\frac{\beta^{(1)}_{(1)} \left(d - 1\right)}{2 \pi} - \frac{d}{2} + 
	\frac{1}{2},\\
	{g_2 =} - \frac{a_{1} \alpha^{(2)}_{(1)} \left(d^{2} - 3 d + 2\right)}{2 
	\pi} 
	+ 
	\frac{a_{1} \left(d^{2} - 3 d + 2\right)}{2} - \frac{a_{2} 
	\alpha^{(2)}_{(1)} 
		\left(d^{3} - 6 d^{2} + 11 d - 6\right)}{2 \pi} + \frac{a_{2} 
		\left(d^{3} - 5 
		d^{2} + 8 d - 4\right) \sin{\left(\alpha^{(2)}_{(1)} \right)}}{2 \pi} + 
	\frac{a_{2} \left(d^{3} - 5 d^{2} + 8 d - 4\right)}{2} + \frac{a_{2} 
	\nu_{(d)} 
		\left(d^{2} - 3 d + 2\right) \sin{\left(\alpha^{(d)}_{(1)} \right)}}{2 
		\pi 
		\nu_{(1)}} - \frac{a_{2} \left(d^{2} - 3 d + 2\right) 
		\sin{\left(\beta^{(1)}_{(1)} \right)}}{2 \pi \nu_{(1)}} - \frac{a_{2} 
		\left(d^{2} - 3 d + 2\right) \sin{\left(\beta^{(d)}_{(1)} \right)}}{2 
		\pi 
		\nu_{(1)}} - \frac{a_{2} \left(d^{3} - 5 d^{2} + 8 d - 4\right) 
		\sin{\left(\beta^{(2)}_{(1)} \right)}}{2 \pi \nu_{(1)}} - \frac{a_{4} 
		\alpha^{(d)}_{(1)} \left(d^{2} - 3 d + 2\right)}{2 \pi} + \frac{a_{4} 
		\left(d^{2} - 3 d + 2\right)}{2} + \frac{\beta^{(2)}_{(1)} \left(d^{2} 
		- 3 d + 
		2\right)}{2 \pi} - \frac{d^{2}}{2} + \frac{3 d}{2} - 1,
	\end{dmath*}
	\begin{dmath*}
	{g_3 =} - \frac{a_{3} \alpha^{(2)}_{(1)} \left(d^{2} - 3 d + 2\right)}{2 
	\pi} 
	+ 
	\frac{a_{3} \left(d^{2} - 3 d + 2\right) \sin{\left(\alpha^{(2)}_{(1)} 
			\right)}}{2 \pi} + \frac{a_{3} \left(d^{2} - 2 d + 1\right)}{2} + 
			\frac{a_{3} 
		\nu_{(d)} \left(d - 1\right) \sin{\left(\alpha^{(d)}_{(1)} \right)}}{2 
		\pi 
		\nu_{(1)}} - \frac{a_{3} \left(d - 1\right) 
		\sin{\left(\beta^{(1)}_{(1)} 
			\right)}}{2 \pi \nu_{(1)}} - \frac{a_{3} \left(d - 1\right) 
		\sin{\left(\beta^{(d)}_{(1)} \right)}}{2 \pi \nu_{(1)}} - \frac{a_{3} 
		\left(d^{2} - 3 d + 2\right) \sin{\left(\beta^{(2)}_{(1)} \right)}}{2 
		\pi 
		\nu_{(1)}} - \frac{a_{5} \alpha^{(d)}_{(1)} \left(d - 1\right)}{2 \pi} 
		+ 
	\frac{a_{5} \left(d - 1\right)}{2} + \frac{\beta^{(d)}_{(1)} \left(d - 
		1\right)}{2 \pi} - \frac{d}{2} + \frac{1}{2},\\
	{g_4 =} 	- \frac{a_{1} \alpha^{(d)}_{(1)} \left(d - 1\right)}{2 \pi} + 
	\frac{a_{1} \left(d - 1\right)}{2} - \frac{a_{2} \alpha^{(d)}_{(1)} 
		\left(d^{2} - 3 d + 2\right)}{2 \pi} + \frac{a_{2} \left(d^{2} - 3 d + 
		2\right)}{2} + \frac{a_{4} \nu_{(1)} \left(d^{2} - 2 d + 1\right) 
		\sin{\left(\alpha^{(d)}_{(1)} \right)}}{2 \pi \nu_{(d)}} + \frac{a_{4} 
		\left(d 
		- 1\right)}{2} - \frac{a_{4} \left(d - 1\right) 
		\sin{\left(\beta^{(d)}_{(d)} 
			\right)}}{2 \pi \nu_{(d)}} - \frac{a_{4} \left(d^{2} - 2 d + 
			1\right) 
		\sin{\left(\beta^{(1)}_{(d)} \right)}}{2 \pi \nu_{(d)}} + 
	\frac{\beta^{(1)}_{(d)} \left(d - 1\right)}{2 \pi} - \frac{d}{2} + 
	\frac{1}{2},\\
	{g_5=}	- \frac{a_{3} \alpha^{(d)}_{(1)} \left(d - 1\right)}{2 \pi} + 
	\frac{a_{3} \left(d - 1\right)}{2} + \frac{a_{5} \nu_{(1)} \left(d - 
	1\right) 
		\sin{\left(\alpha^{(d)}_{(1)} \right)}}{2 \pi \nu_{(d)}} + 
		\frac{a_{5}}{2} - 
	\frac{a_{5} \left(d - 1\right) \sin{\left(\beta^{(1)}_{(d)} \right)}}{2 \pi 
		\nu_{(d)}} - \frac{a_{5} \sin{\left(\beta^{(d)}_{(d)} \right)}}{2 \pi 
		\nu_{(d)}} + \frac{\beta^{(d)}_{(d)}}{2 \pi} - \frac{1}{2}.
\end{dmath*}

\subsection{Gradient expressions for $\cW_2$}
The space $\cW_2$ is six-dimensional. A weight 
matrix for $d=8$ can be parameterized as follows 
\begin{align*}
\left[\begin{matrix}a_{1} & a_{2} & a_{2} & a_{2} & a_{2} & a_{2} & a_{3} & 
a_{3}\\a_{2} & a_{1} & a_{2} & a_{2} & a_{2} & a_{2} & a_{3} & a_{3}\\a_{2} & 
a_{2} & a_{1} & a_{2} & a_{2} & a_{2} & a_{3} & a_{3}\\a_{2} & a_{2} & a_{2} & 
a_{1} & a_{2} & a_{2} & a_{3} & a_{3}\\a_{2} & a_{2} & a_{2} & a_{2} & a_{1} & 
a_{2} & a_{3} & a_{3}\\a_{2} & a_{2} & a_{2} & a_{2} & a_{2} & a_{1} & a_{3} & 
a_{3}\\a_{4} & a_{4} & a_{4} & a_{4} & a_{4} & a_{4} & a_{5} & a_{6}\\a_{4} & 
a_{4} & a_{4} & a_{4} & a_{4} & a_{4} & a_{6} & a_{5}\end{matrix}\right].
\end{align*}
The gradient entries, denoted by $g_1, g_2, g_3, g_4, g_5$ and $g_6$, are:
\begin{dmath*}
{g_1 =} \frac{a_{1} \left(d - 2\right)}{2} + \frac{a_{1} \left(d^{2} - 5 d + 
6\right) \sin{\left(\alpha^{(2)}_{(1)} \right)}}{2 \pi} + \frac{a_{1} 
\nu_{(d-1)} \left(d - 2\right) \sin{\left(\alpha^{(d-1)}_{(1)} \right)}}{\pi 
\nu_{(1)}} - \frac{a_{1} \left(d - 2\right) \sin{\left(\beta^{(1)}_{(1)} 
\right)}}{2 \pi \nu_{(1)}} - \frac{a_{1} \left(d - 2\right) 
\sin{\left(\beta^{(d-1)}_{(1)} \right)}}{\pi \nu_{(1)}} - \frac{a_{1} 
\left(d^{2} - 5 d + 6\right) \sin{\left(\beta^{(2)}_{(1)} \right)}}{2 \pi 
\nu_{(1)}} - \frac{a_{2} \alpha^{(2)}_{(1)} \left(d^{2} - 5 d + 6\right)}{2 
\pi} + \frac{a_{2} \left(d^{2} - 5 d + 6\right)}{2} - \frac{a_{4} 
\alpha^{(d-1)}_{(1)} \left(d - 2\right)}{\pi} + a_{4} \left(d - 2\right) + 
\frac{\beta^{(1)}_{(1)} \left(d - 2\right)}{2 \pi} - \frac{d}{2} + 1,
\end{dmath*}
\begin{dmath*}
{g_2 =} - \frac{a_{1} \alpha^{(2)}_{(1)} \left(d^{2} - 5 d + 6\right)}{2 \pi} + 
\frac{a_{1} \left(d^{2} - 5 d + 6\right)}{2} - \frac{a_{2} \alpha^{(2)}_{(1)} 
\left(d^{3} - 9 d^{2} + 26 d - 24\right)}{2 \pi} + \frac{a_{2} \left(d^{3} - 8 
d^{2} + 21 d - 18\right) \sin{\left(\alpha^{(2)}_{(1)} \right)}}{2 \pi} + 
\frac{a_{2} \left(d^{3} - 8 d^{2} + 21 d - 18\right)}{2} + \frac{a_{2} 
\nu_{(d-1)} \left(d^{2} - 5 d + 6\right) \sin{\left(\alpha^{(d-1)}_{(1)} 
\right)}}{\pi \nu_{(1)}} - \frac{a_{2} \left(d^{2} - 5 d + 6\right) 
\sin{\left(\beta^{(1)}_{(1)} \right)}}{2 \pi \nu_{(1)}} - \frac{a_{2} 
\left(d^{2} - 5 d + 6\right) \sin{\left(\beta^{(d-1)}_{(1)} \right)}}{\pi 
\nu_{(1)}} - \frac{a_{2} \left(d^{3} - 8 d^{2} + 21 d - 18\right) 
\sin{\left(\beta^{(2)}_{(1)} \right)}}{2 \pi \nu_{(1)}} - \frac{a_{4} 
\alpha^{(d-1)}_{(1)} \left(d^{2} - 5 d + 6\right)}{\pi} + a_{4} \left(d^{2} - 5 
d + 6\right) + \frac{\beta^{(2)}_{(1)} \left(d^{2} - 5 d + 6\right)}{2 \pi} - 
\frac{d^{2}}{2} + \frac{5 d}{2} - 3,\\{g_3 =} - \frac{a_{3} \alpha^{(2)}_{(1)} 
\left(d^{2} - 5 d + 6\right)}{\pi} + \frac{a_{3} \left(d^{2} - 5 d + 6\right) 
\sin{\left(\alpha^{(2)}_{(1)} \right)}}{\pi} + a_{3} \left(d^{2} - 4 d + 
4\right) + \frac{2 a_{3} \nu_{(d-1)} \left(d - 2\right) 
\sin{\left(\alpha^{(d-1)}_{(1)} \right)}}{\pi \nu_{(1)}} - \frac{a_{3} \left(d 
- 2\right) \sin{\left(\beta^{(1)}_{(1)} \right)}}{\pi \nu_{(1)}} - \frac{2 
a_{3} \left(d - 2\right) \sin{\left(\beta^{(d-1)}_{(1)} \right)}}{\pi 
\nu_{(1)}} - \frac{a_{3} \left(d^{2} - 5 d + 6\right) 
\sin{\left(\beta^{(2)}_{(1)} \right)}}{\pi \nu_{(1)}} - \frac{a_{5} 
\alpha^{(d-1)}_{(1)} \left(d - 2\right)}{\pi} + a_{5} \left(d - 2\right) - 
\frac{a_{6} \alpha^{(d-1)}_{(1)} \left(d - 2\right)}{\pi} + a_{6} \left(d - 
2\right) + \frac{\beta^{(d-1)}_{(1)} \left(d - 2\right)}{\pi} - d + 2,
\end{dmath*}

\begin{dmath*}
{g_4 =} - \frac{a_{1} \alpha^{(d-1)}_{(1)} \left(d - 2\right)}{\pi} + a_{1} 
\left(d - 
2\right) - \frac{a_{2} \alpha^{(d-1)}_{(1)} \left(d^{2} - 5 d + 6\right)}{\pi} 
+ a_{2} \left(d^{2} - 5 d + 6\right) - \frac{a_{4} \alpha^{(d)}_{(d-1)} \left(d 
- 2\right)}{\pi} + \frac{a_{4} \nu_{(1)} \left(d^{2} - 4 d + 4\right) 
\sin{\left(\alpha^{(d-1)}_{(1)} \right)}}{\pi \nu_{(d-1)}} + \frac{a_{4} 
\left(d - 2\right) \sin{\left(\alpha^{(d)}_{(d-1)} \right)}}{\pi} + 2 a_{4} 
\left(d - 2\right) - \frac{a_{4} \left(d - 2\right) 
\sin{\left(\beta^{(d)}_{(d-1)} \right)}}{\pi \nu_{(d-1)}} - \frac{a_{4} \left(d 
- 2\right) \sin{\left(\beta^{(d-1)}_{(d-1)} \right)}}{\pi \nu_{(d-1)}} - 
\frac{a_{4} \left(d^{2} - 4 d + 4\right) \sin{\left(\beta^{(1)}_{(d-1)} 
\right)}}{\pi \nu_{(d-1)}} + \frac{\beta^{(1)}_{(d-1)} \left(d - 2\right)}{\pi} 
- d + 2,\\
{g_5 =} - \frac{a_{3} \alpha^{(d-1)}_{(1)} \left(d - 2\right)}{\pi} + 
a_{3} \left(d - 2\right) + \frac{a_{5} \nu_{(1)} \left(d - 2\right) 
\sin{\left(\alpha^{(d-1)}_{(1)} \right)}}{\pi \nu_{(d-1)}} + \frac{a_{5} 
\sin{\left(\alpha^{(d)}_{(d-1)} \right)}}{\pi} + a_{5} - \frac{a_{5} \left(d - 
2\right) \sin{\left(\beta^{(1)}_{(d-1)} \right)}}{\pi \nu_{(d-1)}} - 
\frac{a_{5} \sin{\left(\beta^{(d)}_{(d-1)} \right)}}{\pi \nu_{(d-1)}} - 
\frac{a_{5} \sin{\left(\beta^{(d-1)}_{(d-1)} \right)}}{\pi \nu_{(d-1)}} - 
\frac{a_{6} \alpha^{(d)}_{(d-1)}}{\pi} + a_{6} + 
\frac{\beta^{(d-1)}_{(d-1)}}{\pi} - 1\\{g_6 =} - \frac{a_{3} 
\alpha^{(d-1)}_{(1)} \left(d - 2\right)}{\pi} + a_{3} \left(d - 2\right) - 
\frac{a_{5} \alpha^{(d)}_{(d-1)}}{\pi} + a_{5} + \frac{a_{6} \nu_{(1)} \left(d 
- 2\right) \sin{\left(\alpha^{(d-1)}_{(1)} \right)}}{\pi \nu_{(d-1)}} + 
\frac{a_{6} \sin{\left(\alpha^{(d)}_{(d-1)} \right)}}{\pi} + a_{6} - 
\frac{a_{6} \left(d - 2\right) \sin{\left(\beta^{(1)}_{(d-1)} \right)}}{\pi 
\nu_{(d-1)}} - \frac{a_{6} \sin{\left(\beta^{(d)}_{(d-1)} \right)}}{\pi 
\nu_{(d-1)}} - \frac{a_{6} \sin{\left(\beta^{(d-1)}_{(d-1)} \right)}}{\pi 
\nu_{(d-1)}} + \frac{\beta^{(d)}_{(d-1)}}{\pi} - 1.
\end{dmath*}

\subsection{Gradient expressions for $\cW_3$}
The space $\cW_3$ is six-dimensional. A weight 
matrix for $d=8$ can be parameterized as follows 
\begin{align*}
\left[\begin{matrix}a_{1} & a_{2} & a_{2} & a_{2} & a_{2} & a_{3} & a_{3} & 
a_{3}\\a_{2} & a_{1} & a_{2} & a_{2} & a_{2} & a_{3} & a_{3} & a_{3}\\a_{2} & 
a_{2} & a_{1} & a_{2} & a_{2} & a_{3} & a_{3} & a_{3}\\a_{2} & a_{2} & a_{2} & 
a_{1} & a_{2} & a_{3} & a_{3} & a_{3}\\a_{2} & a_{2} & a_{2} & a_{2} & a_{1} & 
a_{3} & a_{3} & a_{3}\\a_{4} & a_{4} & a_{4} & a_{4} & a_{4} & a_{5} & a_{6} & 
a_{6}\\a_{4} & a_{4} & a_{4} & a_{4} & a_{4} & a_{6} & a_{5} & a_{6}\\a_{4} & 
a_{4} & a_{4} & a_{4} & a_{4} & a_{6} & a_{6} & a_{5}\end{matrix}\right]
\end{align*}
The gradient entries, denoted by $g_1, g_2, g_3, g_4, g_5$ and $g_6$, are:
\begin{dmath*}
{g_1 =} \frac{a_{1} \left(d - 3\right)}{2} + \frac{a_{1} \left(d^{2} - 7 d + 
12\right) \sin{\left(\alpha^{(2)}_{(1)} \right)}}{2 \pi} + \frac{3 a_{1} 
\nu_{(d-2)} \left(d - 3\right) \sin{\left(\alpha^{(d-2)}_{(1)} \right)}}{2 \pi 
\nu_{(1)}} - \frac{a_{1} \left(d - 3\right) \sin{\left(\beta^{(1)}_{(1)} 
\right)}}{2 \pi \nu_{(1)}} - \frac{3 a_{1} \left(d - 3\right) 
\sin{\left(\beta^{(d-2)}_{(1)} \right)}}{2 \pi \nu_{(1)}} - \frac{a_{1} 
\left(d^{2} - 7 d + 12\right) \sin{\left(\beta^{(2)}_{(1)} \right)}}{2 \pi 
\nu_{(1)}} - \frac{a_{2} \alpha^{(2)}_{(1)} \left(d^{2} - 7 d + 12\right)}{2 
\pi} + \frac{a_{2} \left(d^{2} - 7 d + 12\right)}{2} - \frac{3 a_{4} 
\alpha^{(d-2)}_{(1)} \left(d - 3\right)}{2 \pi} + \frac{3 a_{4} \left(d - 
3\right)}{2} + \frac{\beta^{(1)}_{(1)} \left(d - 3\right)}{2 \pi} - \frac{d}{2} 
+ \frac{3}{2},\\{g_2 =} - \frac{a_{1} \alpha^{(2)}_{(1)} \left(d^{2} - 7 d + 
12\right)}{2 \pi} + \frac{a_{1} \left(d^{2} - 7 d + 12\right)}{2} - \frac{a_{2} 
\alpha^{(2)}_{(1)} \left(d^{3} - 12 d^{2} + 47 d - 60\right)}{2 \pi} + 
\frac{a_{2} \left(d^{3} - 11 d^{2} + 40 d - 48\right) 
\sin{\left(\alpha^{(2)}_{(1)} \right)}}{2 \pi} + \frac{a_{2} \left(d^{3} - 11 
d^{2} + 40 d - 48\right)}{2} + \frac{3 a_{2} \nu_{(d-2)} \left(d^{2} - 7 d + 
12\right) \sin{\left(\alpha^{(d-2)}_{(1)} \right)}}{2 \pi \nu_{(1)}} - 
\frac{a_{2} \left(d^{2} - 7 d + 12\right) \sin{\left(\beta^{(1)}_{(1)} 
\right)}}{2 \pi \nu_{(1)}} - \frac{3 a_{2} \left(d^{2} - 7 d + 12\right) 
\sin{\left(\beta^{(d-2)}_{(1)} \right)}}{2 \pi \nu_{(1)}} - \frac{a_{2} 
\left(d^{3} - 11 d^{2} + 40 d - 48\right) \sin{\left(\beta^{(2)}_{(1)} 
\right)}}{2 \pi \nu_{(1)}} - \frac{3 a_{4} \alpha^{(d-2)}_{(1)} \left(d^{2} - 7 
d + 12\right)}{2 \pi} + \frac{3 a_{4} \left(d^{2} - 7 d + 12\right)}{2} + 
\frac{\beta^{(2)}_{(1)} \left(d^{2} - 7 d + 12\right)}{2 \pi} - \frac{d^{2}}{2} 
+ \frac{7 d}{2} - 6,\\
{g_3 =} - \frac{3 a_{3} \alpha^{(2)}_{(1)} \left(d^{2} - 7 
d + 12\right)}{2 \pi} + \frac{3 a_{3} \left(d^{2} - 7 d + 12\right) 
\sin{\left(\alpha^{(2)}_{(1)} \right)}}{2 \pi} + \frac{3 a_{3} \left(d^{2} - 6 
d + 9\right)}{2} + \frac{9 a_{3} \nu_{(d-2)} \left(d - 3\right) 
\sin{\left(\alpha^{(d-2)}_{(1)} \right)}}{2 \pi \nu_{(1)}} - \frac{3 a_{3} 
\left(d - 3\right) \sin{\left(\beta^{(1)}_{(1)} \right)}}{2 \pi \nu_{(1)}} - 
\frac{9 a_{3} \left(d - 3\right) \sin{\left(\beta^{(d-2)}_{(1)} \right)}}{2 \pi 
\nu_{(1)}} - \frac{3 a_{3} \left(d^{2} - 7 d + 12\right) 
\sin{\left(\beta^{(2)}_{(1)} \right)}}{2 \pi \nu_{(1)}} - \frac{3 a_{5} 
\alpha^{(d-2)}_{(1)} \left(d - 3\right)}{2 \pi} + \frac{3 a_{5} \left(d - 
3\right)}{2} - \frac{3 a_{6} \alpha^{(d-2)}_{(1)} \left(d - 3\right)}{\pi} + 3 
a_{6} \left(d - 3\right) + \frac{3 \beta^{(d-2)}_{(1)} \left(d - 3\right)}{2 
\pi} - \frac{3 d}{2} + \frac{9}{2},\\
\end{dmath*}
\begin{dmath*}
{g_4 =} - \frac{3 a_{1} 
\alpha^{(d-2)}_{(1)} \left(d - 3\right)}{2 \pi} + \frac{3 a_{1} \left(d - 
3\right)}{2} - \frac{3 a_{2} \alpha^{(d-2)}_{(1)} \left(d^{2} - 7 d + 
12\right)}{2 \pi} + \frac{3 a_{2} \left(d^{2} - 7 d + 12\right)}{2} - \frac{3 
a_{4} \alpha^{(d-1)}_{(d-2)} \left(d - 3\right)}{\pi} + \frac{3 a_{4} \nu_{(1)} 
\left(d^{2} - 6 d + 9\right) \sin{\left(\alpha^{(d-2)}_{(1)} \right)}}{2 \pi 
\nu_{(d-2)}} + \frac{3 a_{4} \left(d - 3\right) 
\sin{\left(\alpha^{(d-1)}_{(d-2)} \right)}}{\pi} + \frac{9 a_{4} \left(d - 
3\right)}{2} - \frac{3 a_{4} \left(d - 3\right) 
\sin{\left(\beta^{(d-1)}_{(d-2)} \right)}}{\pi \nu_{(d-2)}} - \frac{3 a_{4} 
\left(d - 3\right) \sin{\left(\beta^{(d-2)}_{(d-2)} \right)}}{2 \pi 
\nu_{(d-2)}} - \frac{3 a_{4} \left(d^{2} - 6 d + 9\right) 
\sin{\left(\beta^{(1)}_{(d-2)} \right)}}{2 \pi \nu_{(d-2)}} + \frac{3 
\beta^{(1)}_{(d-2)} \left(d - 3\right)}{2 \pi} - \frac{3 d}{2} + 
\frac{9}{2},\\{g_5 =} - \frac{3 a_{3} \alpha^{(d-2)}_{(1)} \left(d - 
3\right)}{2 
\pi} + \frac{3 a_{3} \left(d - 3\right)}{2} + \frac{3 a_{5} \nu_{(1)} \left(d - 
3\right) \sin{\left(\alpha^{(d-2)}_{(1)} \right)}}{2 \pi \nu_{(d-2)}} + \frac{3 
a_{5} \sin{\left(\alpha^{(d-1)}_{(d-2)} \right)}}{\pi} + \frac{3 a_{5}}{2} - 
\frac{3 a_{5} \left(d - 3\right) \sin{\left(\beta^{(1)}_{(d-2)} \right)}}{2 \pi 
\nu_{(d-2)}} - \frac{3 a_{5} \sin{\left(\beta^{(d-1)}_{(d-2)} \right)}}{\pi 
\nu_{(d-2)}} - \frac{3 a_{5} \sin{\left(\beta^{(d-2)}_{(d-2)} \right)}}{2 \pi 
\nu_{(d-2)}} - \frac{3 a_{6} \alpha^{(d-1)}_{(d-2)}}{\pi} + 3 a_{6} + \frac{3 
\beta^{(d-2)}_{(d-2)}}{2 \pi} - \frac{3}{2},\\{g_6 =} - \frac{3 a_{3} 
\alpha^{(d-2)}_{(1)} \left(d - 3\right)}{\pi} + 3 a_{3} \left(d - 3\right) - 
\frac{3 a_{5} \alpha^{(d-1)}_{(d-2)}}{\pi} + 3 a_{5} - \frac{3 a_{6} 
\alpha^{(d-1)}_{(d-2)}}{\pi} + \frac{3 a_{6} \nu_{(1)} \left(d - 3\right) 
\sin{\left(\alpha^{(d-2)}_{(1)} \right)}}{\pi \nu_{(d-2)}} + \frac{6 a_{6} 
\sin{\left(\alpha^{(d-1)}_{(d-2)} \right)}}{\pi} + 6 a_{6} - \frac{3 a_{6} 
\left(d - 3\right) \sin{\left(\beta^{(1)}_{(d-2)} \right)}}{\pi \nu_{(d-2)}} - 
\frac{6 a_{6} \sin{\left(\beta^{(d-1)}_{(d-2)} \right)}}{\pi \nu_{(d-2)}} - 
\frac{3 a_{6} \sin{\left(\beta^{(d-2)}_{(d-2)} \right)}}{\pi \nu_{(d-2)}} + 
\frac{3 \beta^{(d-1)}_{(d-2)}}{\pi} - 3.\\
\end{dmath*}

%% file: appendix.tex

\section{Hessian spectrum} \label{sec:hes_spec}

Below, we describe the technique we use to derive an analytic description of 
the Hessian spectrum. Some parts follow \cite{arjevanifield2020hessian} 
verbatim. In order to avoid a long preliminaries section, key ideas and 
concepts are introduced and organized so as to illuminate our strategy for 
analyzing the Hessian. We illustrate with reference to the global minimum $W=V$
where $d=k$, the second layer is all ones, and the target weight matrix $V$ is 
the identity $I_d$. In \pref{sec:eig_trans}, we provide the eigenvalue 
expressions for $\Delta (S_{d-1}\times S_1)$, organized by their isotypic 
component.

\subsection{Studying invariance properties via group action} 
\label{sec:group_action}
We first review background material on group actions
and fix notations (see~\cite[Chapters 1, 2]{field2007} for a 
more complete account). Elementary concepts from group theory 
are assumed known. We start with two examples that are used later.
\begin{exams} \label{exams:groups}
	(1) The \emph{symmetric group} $S_d$, $d\in\pint$, is the group of 
	permutations of $\ibr{d}\defoo \{1,\dots,d\}$.  \\
	(2) Let $\GLG{d}$ denote the space of invertible 
	linear maps on $\real^d$. Under composition, $\GLG{d}$ has the 
	structure of a 
	group. The \emph{orthogonal group} $\text{O}(d)$ is the 
	subgroup 
	of $\GLG{d}$ 
	defined by
	$
	\text{O}(d) =   \{A \in \GLG{d} \dd \|Ax\| = 
	\|x\|,\;\text{for all } x \in \real^d\}.
	$
	Both $\GLG{d}$ and $\text{O}(d)$ can be viewed as groups of 
	invertible $d \times d$ matrices.
\end{exams}
Characteristically, these groups consist of 
\emph{transformations}  
of a set and so we are led to the notion of a 
\emph{$G$-space} $X$ where we have 
an \emph{action} of a group $G$
on a set $X$. Formally, this is a  group homomorphism from $G$ 
to the
group of bijections of $X$.
For example, $S_d$ naturally acts on $[d]$ as permutations and 
both $\GLG{d}$ and $\text{O}(d)$ act on
$\real^d$ as linear transformations (or matrix 
multiplication). 

An example, which we use extensively in studying the invariance 
properties of $\ploss$, is given by the action of the
group $S_k \times S_d\subset S_{k\times d},~k,d\in 
\pint$, on
$\ibr{k} \times \ibr{d}$ defined by
\begin{align}\label{eq: Gamma-action}
	(\pi,\rho)(i,j) = (\pi^{-1}(i),\rho^{-1}(j)),\; \pi \in 
	S_k, 
	\rho \in S_d,\; (i,j) \in \ibr{k} \times \ibr{d}.
\end{align}
This action induces an action on the space $M(k,d)$ of $k \times 
d$-matrices $A = [A_{ij}]$ by
$ (\pi,\rho)[A_{ij}] = [A_{\pi^{-1}(i),\rho^{-1}(j)}]$.
The action can be defined in terms of permutation matrices but 
is easier to describe in terms of rows and columns: 
$ (\pi,\rho)A$ permutes rows (resp.~columns) of $A$ according to 
$\pi$ (resp.~$\rho$). As mentioned in the introduction, for our choice of 
$V=I_d$, $\ploss$ is $S_k \times S_d$-invariant. Note that $\Delta 
S_{d}\approx S_d$. When we restrict the $S_d \times S_d$-action on $M(k,k)$ to 
$\Delta S_d$, we refer to the diagonal $S_d$-action, or just the $S_d$-action 
on $M(d,d)$. This action of $S_{d}$ on $M(d,d)$ maps diagonal matrices to 
diagonal matrices and should not be confused with the actions of $S_d$ on 
$M(d,d)$ defined by either permuting rows or columns. 

\begin{exam}
	Take $p,q \in\pint$, $p+q = d$, and consider the 
	diagonal
	action
	of $S_{p} \times S_q\subset S_d$ on	$M(d,d)$.  Write $A \in M(d,d)$ in 
	block matrix form as $A =
	\left[\begin{matrix} A_{p,p} & A_{p,q}\\
		A_{q,p}
		&
		A_{q,q}
	\end{matrix}\right]$.
	If $(g,h) \in S_{p} \times S_q\subset S_d$, then $(g,h)A =
	\left[\begin{matrix} gA_{p,p} & (g,h)A_{p,q}\\
		(h,g)A_{q,p} & hA_{q,q}
	\end{matrix}\right]$
	where $gA_{p,p}$ (resp.~$hA_{q,q}$) are defined via the 
	diagonal
	action of $S_p$ (resp.~$S_q$) on $A_{p,p}$ 
	(resp.~$A_{q,q}$), and
	$(g,h)A_{p,q}$ and $(h,g)A_{q,p}$ are defined
	through the natural action of $S_p \times S_q$ on rows and 
	columns. Thus,
	for $(g,h)A_{p,q}$ (resp.~$(h,g)A_{q,p}$) we permute rows 
	(resp.~columns) according to $g$ and
	columns (resp.~rows) according to $h$.
	In the case when $p=d-1$, $q = 1$,
	$S_{d-1}$ will act diagonally on
	$A_{d-1,d-1}$, fix
	$a_{dd}$, and act by permuting the first $(d-1)$ entries of 
	the	last row and column.
\end{exam}

As mentioned in body of the paper, given $W \in M(d,d)$, the largest 
subgroup of $S_d \times S_d$ fixing $W$ is called the \emph{isotropy} 
subgroup of $W$ and is used as means of measuring the symmetry of $W$. The 
isotropy subgroup of $V\in M(d,d)$ is the diagonal subgroup $\Delta S_d$. 
Our focus 
will be on critical points $W$ whose isotropy groups are 
subgroups of the target matrix $V=I_d$, that is, $\Delta S_d$ and 
$\Delta (S_{d-1}\times S_1)$ (see \pref{fig:max_sym}---we use the notation 
$\Delta 
S_d$ as the
isotropy is a \emph{subgroup} of $S_d \times S_d$). In the 
next section, we show how the symmetry of local minima greatly simplifies 
the analysis of their Hessian.

\subsection{The spectrum of equivariant linear isomorphisms} 
\label{sec:spec}

If $G$ is a subgroup of $\Od{d}$, the action on $\real^d$  
is called an \emph{orthogonal} representation of $G$ (we 
often drop the qualifier orthogonal). Denote by 
$(\real^d,G)$ as necessary. The \emph{degree} of a 
representation $(V,G)$ is the dimension of $V$ ($V$  will 
always be a linear subspace of some $\real^n$ with the 
induced Euclidean inner product). The action of $S_k \times 
S_d\subset S_{k\times d}$ on $M(k,d)$  is {orthogonal} with 
respect to the standard Euclidean inner product on $M(k,d) 
\approx \real^{k\times d}$ since the action permutes the 
coordinates of $\real^{k\times d}$ (equivalently, 
components of $k\times d$ matrices). Given two 
representations $(V,G)$ and $(W,G)$, a map $A: V \arr W$ is 
called $G$-equivariant if $A(gv) = gA(v)$, for all $g \in 
G, v \in V$. If $A$ is linear and equivariant, we say $A$ 
is a \emph{$G$-map}. Invariant functions naturally provide
examples of equivariant maps. Thus the gradient $\nabla 
\ploss$ is a $S_k \times S_d$-equivariant self map of $M(k,d)$ 
and if $W$ is a critical point of $\nabla \ploss$ with 
isotropy $G \subset S_k \times S_d$, then
$\nabla^2 \ploss(W): M(k,d)\arr M(k,d)$ is a $G$-map 
(see~\cite{ArjevaniField2020}). The equivariance 
of the Hessian is the key ingredient that allows us to 
study the spectral density at \emph{symmetric} local 
minima.

A representation $(\real^n,G)$ is \emph{irreducible} 
if the only linear subspaces of $\real^n$ that are preserved 
(invariant)  by the $G$-action are
$\real^n$ and $\{0\}$. Two orthogonal 
representations $(V,G)$, $(W,G)$ are \emph{isomorphic} (and have the same 
\emph{isomorphism class}) if there 
exists a $G$-map $A: V \arr W$ which is a linear isomorphism. If $(V,G)$, 
$(W,G)$ are irreducible but not isomorphic then
every $G$-map $A: V \arr W$ is zero 
(as the kernel and the image of a $G$-map are $G$-invariant). If $(V,G)$ is 
irreducible, then the space $\text{Hom}_G(V,V)$ of $G$-maps (endomorphisms) 
of $V$ is a real associative division algebra and is isomorphic by a theorem 
of Frobenius to either $\real, \mathbb{C}$ or $\mathbb{H}$ (the quaternions). 
The \emph{only} case that will concern us here is when $\text{Hom}_G(V,V) 
\approx \real$ when we say the representation is \emph{real}.
\begin{exam}\label{ex: iso}
	Let $n > 1$. Take the natural (orthogonal) action of $S_n$ on $\real^n$ 
	defined by permuting coordinates.  The representation is not 
	irreducible since the subspace $T = 
	\{(x,x,\cdots,x)\in\real^n \dd x \in \real\}$ 
	is 	invariant by the action of $S_n$, as is the hyperplane 
	$H_{n-1} 
	= T^\perp =\{(x_1,\cdots,x_n)\dd \sum_{i\in\ibr{n}}x_i =0\}$.
	It is easy to check that $(T,S_n)$, also called the 
	\emph{trivial }representation of 
	$S_n$, and $(H_{n-1},S_n)$, the 
	\emph{standard} representation, are irreducible, real, and 
	not isomorphic.	
\end{exam}

Every representation $(\real^n,G)$ can be written uniquely, up to order, as an 
orthogonal
direct sum $\oplus_{i\in\ibr{m}} V_i$, where each $(V_i,G)$ is an orthogonal 
direct sum of isomorphic irreducible representations $(V_{ij},G)$, $j \in 
\ibr{p_i}$, and $(V_{ij},G)$ is isomorphic to $(V_{i'j'},G)$
if and only if $i' = i$. The subspaces $V_{ij}$ are \emph{not} uniquely 
determined if $p_i > 1$. 
If there are $m$ distinct isomorphism classes 
$\mathfrak{v}_1,\cdots,\mathfrak{v}_m$ of irreducible representations, then 
$(\real^n,G)$
may be represented by the sum $p_1 \mathfrak{v}_1 + \cdots + p_m 
\mathfrak{v}_m$, where $p_i\ge 1$ counts the number of representations with 
isomorphism class $\mathfrak{v}_i$. Up to order, this sum (that is, the 
$\mathfrak{v}_i$ and their multiplicities) is uniquely determined by 
$(\real^n,G)$. This is the \emph{isotypic decomposition} of $(\real^n,G)$ (see 
\cite{thomas2004representations}). The isotypic decomposition is a powerful 
tool for 
extracting information about the spectrum of $G$-maps.

If $G = S_d$, then every irreducible representation of $S_d$ is 
real~\cite[Thm.~4.3]{fulton1991representation}. Suppose, as above, that 
$(\real^n,S_d)=\oplus_{i\in\ibr{m}} V_i$ and $A: \real^n\arr\real^n$ is an 
$S_d$-map. 
Since the induced maps $A_{ii'}: V_i \arr V_{i'}$ must be zero if $i \ne i'$, 
$A$ is uniquely determined by the $S_d$-maps $A_{ii}: V_i \arr V_i$, $i 
\in\ibr{m}$. Fix $i$ and choose an $S_d$-representation $(W,S_d)$
in the isomorphism class $\mathfrak{v}_i$. Choose $S_d$-isomorphisms $W \arr 
V_{ij}$, $j \in \ibr{p_i}$.
Then $A_{ii}$ induces $\overline{A}_{ii}: W^{p_i} \arr W^{p_i}$ and so 
determines a (real) matrix $M_i \in M(p_i,p_i)$ since $\text{Hom}_{S_d}(W,W) 
\approx \real$. Different choices of $V_{ij}$, or isomorphism $W \arr V_{ij}$, 
yield a matrix similar to $M_i$. Each eigenvalue of $M_i$ of multiplicity $r$ 
gives an eigenvalue of $A_{ii}$, and so of $A$, of multiplicity 
$r\,\text{degree}(\mathfrak{v}_i)$.

\begin{fact}\label{fact: iso} (Notations and assumptions as 
	above.)
	If $A$ is the Hessian, all eigenvalues are real and
	each eigenvalue of $M_{i}$ of multiplicity $r$ will be an eigenvalue of 
	$A$ with 
	multiplicity $r\,\text{degree}(\mathfrak{v}_i)$. In 
	particular, $A$ has most $\sum_{i \in \ibr{m}}p_i$ distinct 
	real eigenvalues---regardless of the dimension of the underlying 
	space.
\end{fact}

Our strategy can be now summarized as follows. Given a local minima 
$W$, we compute the isotropy 
group $G \subset S_k \times S_d$ of $W$. Since the 
Hessian of $\cF$ at $W$ is a $G$-map, may use the 
isotypic decomposition of the action of $G$ on $M(k,d)$ to extract the 
spectral properties of the Hessian. In our setting, local minima have 
large isotropy groups, typically, as large as $\Delta(S_p 
\times S_{d-p}),~0\le p <d/2$. Studying the Hessian at these 
minima requires the isotopic decomposition corresponding to 
$\Delta(S_p \times S_{d-p}),~0\le p <d/2$, which we detail in 
\pref{thm: isot2_body} below.

\subsection{The isotypic decomposition of $(M(d,d),S_d)$ 
	and the spectrum at $W=V$} \label{sec: iso_global}

Regard $M(d,d)$ as an $S_d$-space (diagonal action). The trivial 
representation, denoted by $\ft_d$, and the standard representation, denoted 
by 
$\fs_d$, introduced in 
\pref{ex: iso} are examples of the many irreducible 
representations of $S_d$. In the general theory, each irreducible 
representation of $S_d$ is associated to a partition of the set $[d]$. 
The description of the isotypic decomposition of $(M(d,d),S_d)$
is relatively simple  
and uses just 4 irreducible representations of $S_d$ for $d \ge 4$.
\begin{itemize}
	\item The trivial representation $\mathfrak{t}_d$ of degree 
	1.
	\item The standard representation $\mathfrak{s}_d$ of $S_d$ of degree 
	$k-1$.
	\item The exterior square representation 
	$\mathfrak{x}_d=\wedge^2 
	\mathfrak{s}_d$ of degree $\frac{(d-1)(d-2)}{2}$.
	\item A representation $\mathfrak{y}_d$ of degree 
	$\frac{d(d-3)}{2}$. We describe $\mathfrak{y}_d$ explicitly later in terms 
	of 	symmetric matrices (formally, it is the representation associated to 
	the 
	partition $(d-2,2)$).

\end{itemize}
We omit the subscript $d$ when clear from the context. Assume that $d\ge4$. We 
begin with a 
well-known result about the 
representation 
$\mathfrak{s}\otimes\mathfrak{s}$ 
(see, e.g., \cite{fulton1991representation}). If 
$\mathfrak{s}\odot \mathfrak{s}$ denotes the symmetric tensor 
product of $\mathfrak{s}$, then
\begin{equation}\label{eq: ten}
	\mathfrak{s}\otimes\mathfrak{s} = \mathfrak{s}\odot \mathfrak{s} 
	+ \mathfrak{x} = \mathfrak{t}+  \mathfrak{s} + \mathfrak{y} + 
	\mathfrak{x}.
\end{equation}
Since all the irreducible $S_d$-representations are real, they are isomorphic 
to their dual representations and so we have the isotypic decomposition
\begin{eqnarray}\label{eq: Miso1}
	M(k,k)&\approx&\real^k \otimes \real^k \approx 
	(\mathfrak{s}+\mathfrak{t})  \otimes (\mathfrak{s}+\mathfrak{t}) 
	\label{eq: Miso2}
	=  2\mathfrak{t} + 3\mathfrak{s} + \mathfrak{x} + 
	\mathfrak{y},
\end{eqnarray}
since $\mathfrak{t}  \otimes \mathfrak{s} = \mathfrak{s}$ and 
$\mathfrak{t}  \otimes\mathfrak{t} =\mathfrak{t} $.

Using Fact \ref{fact: iso}, information can 
immediately be deduced from  Equation \pref{eq: Miso1}. For example,
if $W$ is a critical point of isotropy $\Delta S_d$ (a fixed point of the 
$S_d$-action on $M(d,d)$), then
the spectrum of the Hessian contains at most $2+3+1+1=7$ distinct 
eigenvalues which distribute as follows: $\ft$ contributes 2 
eigenvalues of multiplicity 1, $\fs$ contributes $2$ eigenvalues of 
multiplicity $d-1$, $\fx$ contributes one eigenvalue of multiplicity 
$\frac{(d-1)(d-2)}{2}$, and $\fy$ contributes one eigenvalue  of 
multiplicity $\frac{d(d-3)}{2}$. This applies to the global minimum 
$W=V$ and the spurious minimum of type A.

Next, we would like to compute the actual eigenvalues. We demonstrate the 
method for the single $\fx$-eigenvalue (the example given in the body of the 
paper in 
section \pref{sec:compute_hess} refers to the single $\fy$-eigenvalue). Pick a 
non-zero vector 
from the $\fx$-representation. For example, 
\[
\mathfrak{X}^d = \left[\begin{matrix}
	0 & 1 &  \ldots & 1 & - (d-2)\\
	-1 & 0  & \ldots & 0 & 1 \\
	\cdots  & \cdots& \cdots& \cdots & \cdots\\
	-1 & 0   &  \ldots & 0 & 1 \\
	(d-2) & -1  &  \ldots & -1 & 0 
\end{matrix}\right],
\]
where rows and columns sum to zero and the only non-zero entries are in rows 
and columns 1 and $d$. Let $\overbar{\mathfrak{X}^d}\in\real^{d\times d}$
be defined by concatenating the rows of $\mathfrak{X}^d$. 
Since $\fx$ only occurs once in the isotopic decomposition and
$\nabla^2\ploss(V)$ is $S_d$-equivariant, 
$\overbar{\mathfrak{X}^d}$ must be an eigenvector. In particular,
$(\nabla^2\ploss(V)\overbar{\mathfrak{X}^d})_i=\lambda_{\mathfrak{x}}\overbar{\mathfrak{X}_i^d}$,
 all $i \in
\ibr{d^2}$. Choose $i$ so that $\overbar{\mathfrak{X}_i^d} \ne 0$. For 
example, $\overbar{\mathfrak{X}_2^d} = 1$.
Matrix multiplication, yields $\lambda_{\mathfrak{x}} =1/4-1/2\pi$. A similar 
analysis 
holds for the eigenvalue associated to $\mathfrak{y}$. The multiple factors 
$2\mathfrak{t}$ 
and $3\mathfrak{s}$ are handled by making judicious choices of orthogonal 
invariant 
subspaces and representative vectors in $M(k,k)$. 

Having described the general strategy for analyzing the Hessian spectrum 
for global minima, we now examine the spectrum at various types of 
spurious minima. We need two additional ingredients: a specification of the 
entries of a 
given family of spurious minima and the respective isotypic decomposition; we 
begin with 
the latter. 

As discussed in the body of the paper, the symmetry-based analysis of the 
Hessian 
relies on the fact that isotropy groups of spurious minima tend to be (and 
some 
provably are) maximal subgroups of the target matrix isotropy. 
For $V=I$, the relevant maximal isotropy groups are of the form 
$\Delta(S_p\times S_q),~p+q=d$. Below, we provide the corresponding 
isotypic decomposition. Assume $d = k$ and regard $M(d,d)$ as an 
$S_p\times S_q$-space, where $S_p \times S_q \subset S_d$ and the (diagonal) 
action of 
$S_d$ is restricted to the subgroup $S_p\times S_q$. 
\begin{theorem*}[{\cite[Theorem 4]{arjevanifield2020hessian}}]\label{thm: 
isot2_body}
	The isotypic decomposition of $(M(d,d),S_p\times S_q)$ is given by:
	\begin{enumerate}
		\item If $p= d-1$, $q=1$, and $k \ge 5$,
		\[
		M(d,d) = 5\mathfrak{t} + 5\mathfrak{s}_{d-1} + \mathfrak{x}_{d-1} + 
		\mathfrak{y}_{d-1}.
		\]
		\item If $q \ge 2$, $d-1 > p > p/2$ and $d \ge 4+q$, then
		\[
		M(d,d) = 6\mathfrak{t}+5 \mathfrak{s}_p  + a \mathfrak{s}_{q} 
		+\mathfrak{x}_p +\mathfrak{y}_p + b\mathfrak{x}_{q} +  
		c\mathfrak{y}_{q} + 2\mathfrak{s}_p \boxtimes \mathfrak{s}_{q},
		\]
		where if
		$q=2$, then $a = 4, b = c = 0$; if  $q=3$, then $a = 5,b = 1, c= 0$; 
		and if $q \ge 
		4$, then $a = 5, b = c = 1$.
	\end{enumerate}
\end{theorem*}
(There is a minor error in {\cite[Theorem 4]{arjevanifield2020hessian}}.
The correct value for $a$ in the second case is as stated here.)

In contrast to the setting considered in \cite{arjevanifield2020hessian}, in 
our case the second layer is trainable. To compute  the isotypic 
decomposition corresponding to this case (i.e., $M(d,d)\times \RR^d$), we 
simply treat the weights of the second layer as an additional row of the 
weight matrix of the first layer. The additional row is split into 
$p$ entries and $d-p$ entries. This adds $\ft + \fs_p $ to the isotypic 
decomposition if $q = 0$, $2\ft + \fs_p $ to the isotypic decomposition if $q 
= 1$, and $2\ft + \fs_p + \fs_q$ otherwise. 
\begin{thm}\label{thm: isot2_body_2L}
	The isotypic decomposition of $(M(d,d),S_p\times S_q) \otimes 
	(\RR^d,S_p\times S_q)$ is 
	given by:
	\begin{enumerate}
		\item If $p= d$, $q=0$, and $k \ge 5$,
		\[
		M(d,d) = 3\mathfrak{t} + 4\mathfrak{s}_{d} + \mathfrak{x}_{d} + 
		\mathfrak{y}_{d}.
		\]
		
		\item If $p= d-1$, $q=1$, and $k \ge 5$,
		\[
		M(d,d) = 7\mathfrak{t} + 6\mathfrak{s}_{d-1} + \mathfrak{x}_{d-1} + 
		\mathfrak{y}_{d-1}.
		\]
		\item If $q \ge 2$, $d-1 > p > p/2$ and $d \ge 4+q$, then
		\[
		M(d,d) = 8\mathfrak{t}+ 6 \mathfrak{s}_p  + a \mathfrak{s}_{q} 
		+\mathfrak{x}_p +\mathfrak{y}_p + b\mathfrak{x}_{q} +  
		c\mathfrak{y}_{q} + 2\mathfrak{s}_p \boxtimes \mathfrak{s}_{q},
		\]
		where if
		$q=2$, then $a = 5, b = c = 0$; if  $q=3$, then $a = 6, b = 1, c= 0$; 
		and if $q \ge 
		4$, then $a = 6, b = c = 1$.
	\end{enumerate}
\end{thm}

\pref{thm: isot2_body_2L} implies that the Hessian 
spectrum of local minima (or critical points) with isotropy $\Delta(S_p\times 
S_q)$ has at most 9 distinct eigenvalues if (1) applies, 
at most 15 distinct eigenvalues if (2) applies, and if (3) 
holds, at most 24 distinct eigenvalues if $q=2$, at most 25 distinct 
eigenvalues if $q=3$, and at most 26 distinct eigenvalues if $q \ge 4$. We 
omit some less interesting cases when $k$ is small.

Following the same lines of argument described  in \pref{sec: iso_global}, the 
next step is to pick a set of non-zero representative vectors for each 
irreducible representation that will allow us to compute the spectrum. We 
adopt the same of choice of representative vectors from 
\cite{arjevanifield2020hessian}. We demonstrate the final step with respect to 
the two trivial 
factors of $S_d$ in \pref{eq: Miso1}. Let $\mathfrak{D}_1$ and 
$\mathfrak{D}_2$ be the two representatives and let $\nabla^2 \ploss 
(\mathfrak{D}_i) = \alpha_{i1} \mathfrak{D}_1 + \alpha_{i2} \mathfrak{D}_2 $, 
$i = 1,2$. The eigenvalues of $\nabla^2 \ploss |2\mathfrak{t}$ are then the 
eigenvalues of the $2\times 2$ transition matrix $A = [\alpha_{ij}]$. To 
compute the eigenvalues of $A$ to, say, $O(d^{-1/2})$-order, one solves the 
equation
\begin{align*}
	\det(A - (x_1 d + x_2 d^{1/2}+x^3)) =0,
\end{align*}
for $x_1,x_2,x_3$, where sufficiently many coefficients of the power series 
of the entries of $A$ are assumed known. The same recipe is used for computing 
the rest of the eigenvalues.

\section{Eigenvalues transition matrices} \label{sec:eig_trans}
Below, we provide the explicit form of the eigenvalue transition matrices 
for the natural representation of $S_d$.

\paragraph{$\fx$-rep.} Let the associated $1\times1$ transition matrix be 
denoted by $T^\fx$. Then,
\begin{dmath*}
	T^{\fx}_{1,1} = - \frac{a_{1}^{2}}{2 \pi \nu_{(1)}^{2} \nu^{(2)}_{(1)}} + 
	\frac{a_{1}^{2} \sin{\left(\alpha^{(2)}_{(1)} \right)}}{2 \pi 
	\nu_{(1)}^{2} \left(\nu^{(2)}_{(1)}\right)^{2}} + \frac{a_{1} a_{2}}{\pi 
	\nu_{(1)}^{2} \nu^{(2)}_{(1)}} - \frac{a_{1} a_{2} 
	\sin{\left(\alpha^{(2)}_{(1)} \right)}}{\pi \nu_{(1)}^{2} 
	\left(\nu^{(2)}_{(1)}\right)^{2}} - \frac{a_{2}^{2}}{2 \pi \nu_{(1)}^{2} 
	\nu^{(2)}_{(1)}} + \frac{a_{2}^{2} \sin{\left(\alpha^{(2)}_{(1)} 
	\right)}}{2 \pi \nu_{(1)}^{2} \left(\nu^{(2)}_{(1)}\right)^{2}} + 
	\frac{\alpha^{(2)}_{(1)}}{2 \pi} + \frac{\left(d - 1\right) 
	\sin{\left(\alpha^{(2)}_{(1)} \right)}}{2 \pi} - \frac{\left(d - 1\right) 
	\sin{\left(\beta^{(2)}_{(1)} \right)}}{2 \pi \nu_{(1)}} - 
	\frac{\sin{\left(\beta^{(1)}_{(1)} \right)}}{2 \pi \nu_{(1)}} - 
	\frac{\sin{\left(\beta^{(2)}_{(1)} \right)}}{2 \pi 
	\left(\mu^{(2)}_{(1)}\right)^{2} \nu_{(1)}}.
\end{dmath*}

\paragraph{$\fy$-rep.} Let the associated $1\times1$ transition matrix be 
denoted by $T^\fy$. Then,
\begin{dmath*}
	T^{\fy}_{1,1} = \frac{a_{1}^{2}}{2 \pi \nu_{(1)}^{2} \nu^{(2)}_{(1)}} + 
	\frac{a_{1}^{2} \sin{\left(\alpha^{(2)}_{(1)} \right)}}{2 \pi 
	\nu_{(1)}^{2} \left(\nu^{(2)}_{(1)}\right)^{2}} - \frac{a_{1} a_{2}}{\pi 
	\nu_{(1)}^{2} \nu^{(2)}_{(1)}} - \frac{a_{1} a_{2} 
	\sin{\left(\alpha^{(2)}_{(1)} \right)}}{\pi \nu_{(1)}^{2} 
	\left(\nu^{(2)}_{(1)}\right)^{2}} + \frac{a_{2}^{2}}{2 \pi \nu_{(1)}^{2} 
	\nu^{(2)}_{(1)}} + \frac{a_{2}^{2} \sin{\left(\alpha^{(2)}_{(1)} 
	\right)}}{2 \pi \nu_{(1)}^{2} \left(\nu^{(2)}_{(1)}\right)^{2}} + 
	\frac{\alpha^{(2)}_{(1)}}{2 \pi} + \frac{\left(d - 1\right) 
	\sin{\left(\alpha^{(2)}_{(1)} \right)}}{2 \pi} - \frac{\left(d - 1\right) 
	\sin{\left(\beta^{(2)}_{(1)} \right)}}{2 \pi \nu_{(1)}} - 
	\frac{\sin{\left(\beta^{(1)}_{(1)} \right)}}{2 \pi \nu_{(1)}} - 
	\frac{\sin{\left(\beta^{(2)}_{(1)} \right)}}{2 \pi 
	\left(\mu^{(2)}_{(1)}\right)^{2} \nu_{(1)}}.
\end{dmath*}

\paragraph{$\fs$-rep.} Let the associated $4\times4$ transition matrix be 
denoted by $T^\fs$. Then,
\begin{dmath*}
	T^{\fs}_{1,1} = - \frac{a_{1}^{2} \left(d - 1\right) 
	\sin{\left(\alpha^{(2)}_{(1)} \right)}}{2 \pi \nu_{(1)}^{2}} - 
	\frac{a_{1}^{2}}{2 \pi \nu_{(1)}^{2} \nu^{(2)}_{(1)}} + \frac{a_{1}^{2} 
	\left(d - 1\right) \sin{\left(\alpha^{(2)}_{(1)} \right)} 
	\cos^{2}{\left(\alpha^{(2)}_{(1)} \right)}}{2 \pi \nu_{(1)}^{2} 
	\left(\nu^{(2)}_{(1)}\right)^{2}} + \frac{a_{1}^{2} \left(d - 1\right) 
	\sin{\left(\beta^{(2)}_{(1)} \right)}}{2 \pi \nu_{(1)}^{3}} + 
	\frac{a_{1}^{2} \sin{\left(\beta^{(1)}_{(1)} \right)}}{2 \pi 
	\nu_{(1)}^{3}} - \frac{a_{1}^{2} \left(d - 1\right) 
	\sin{\left(\beta^{(2)}_{(1)} \right)} \cos^{2}{\left(\beta^{(2)}_{(1)} 
	\right)}}{2 \pi \left(\mu^{(2)}_{(1)}\right)^{2} \nu_{(1)}^{3}} - 
	\frac{a_{1}^{2} \sin{\left(\beta^{(1)}_{(1)} \right)} 
	\cos^{2}{\left(\beta^{(1)}_{(1)} \right)}}{2 \pi 
	\left(\mu^{(1)}_{(1)}\right)^{2} \nu_{(1)}^{3}} + \frac{a_{1} a_{2} 
	\cos{\left(\alpha^{(2)}_{(1)} \right)}}{\pi \nu_{(1)}^{2} \nu^{(2)}_{(1)}} 
	- \frac{a_{1} a_{2} \left(d - 1\right) \sin{\left(\alpha^{(2)}_{(1)} 
	\right)} \cos{\left(\alpha^{(2)}_{(1)} \right)}}{\pi \nu_{(1)}^{2} 
	\left(\nu^{(2)}_{(1)}\right)^{2}} + \frac{a_{1} 
	\sin{\left(\beta^{(1)}_{(1)} \right)} \cos{\left(\beta^{(1)}_{(1)} 
	\right)}}{\pi \left(\mu^{(1)}_{(1)}\right)^{2} \nu_{(1)}^{2}} - 
	\frac{a_{2}^{2}}{2 \pi \nu_{(1)}^{2} \nu^{(2)}_{(1)}} + \frac{a_{2}^{2} 
	\left(d - 1\right) \sin{\left(\alpha^{(2)}_{(1)} \right)}}{2 \pi 
	\nu_{(1)}^{2} \left(\nu^{(2)}_{(1)}\right)^{2}} + \frac{\left(d - 1\right) 
	\sin{\left(\alpha^{(2)}_{(1)} \right)}}{2 \pi} + \frac{1}{2} - 
	\frac{\left(d - 1\right) \sin{\left(\beta^{(2)}_{(1)} \right)}}{2 \pi 
	\nu_{(1)}} - \frac{\sin{\left(\beta^{(1)}_{(1)} \right)}}{2 \pi \nu_{(1)}} 
	- \frac{\sin{\left(\beta^{(1)}_{(1)} \right)}}{2 \pi 
	\left(\mu^{(1)}_{(1)}\right)^{2} \nu_{(1)}},
\end{dmath*}
\begin{dmath*}
	T^{\fs}_{1,2} = \frac{a_{1}^{2} d \cos{\left(\alpha^{(2)}_{(1)} 
	\right)}}{2 \pi \nu_{(1)}^{2} \nu^{(2)}_{(1)}} - \frac{a_{1}^{2} d 
	\sin{\left(\alpha^{(2)}_{(1)} \right)} \cos{\left(\alpha^{(2)}_{(1)} 
	\right)}}{2 \pi \nu_{(1)}^{2} \left(\nu^{(2)}_{(1)}\right)^{2}} - 
	\frac{a_{1} a_{2} d}{\pi \nu_{(1)}^{2} \nu^{(2)}_{(1)}} + \frac{a_{1} 
	a_{2} d \sin{\left(\alpha^{(2)}_{(1)} \right)}}{2 \pi \nu_{(1)}^{2} 
	\left(\nu^{(2)}_{(1)}\right)^{2}} + \frac{a_{1} a_{2} d 
	\sin{\left(\beta^{(1)}_{(1)} \right)}}{2 \pi \nu_{(1)}^{3}} - \frac{a_{1} 
	a_{2} d \sin{\left(\beta^{(1)}_{(1)} \right)} 
	\cos^{2}{\left(\beta^{(1)}_{(1)} \right)}}{2 \pi 
	\left(\mu^{(1)}_{(1)}\right)^{2} \nu_{(1)}^{3}} - \frac{a_{1} a_{2} 
	\left(d^{2} - d\right) \sin{\left(\alpha^{(2)}_{(1)} \right)}}{2 \pi 
	\nu_{(1)}^{2}} - \frac{a_{1} a_{2} \left(d^{2} - 2 d\right) 
	\sin{\left(\alpha^{(2)}_{(1)} \right)} \cos{\left(\alpha^{(2)}_{(1)} 
	\right)}}{2 \pi \nu_{(1)}^{2} \left(\nu^{(2)}_{(1)}\right)^{2}} + 
	\frac{a_{1} a_{2} \left(d^{2} - d\right) \sin{\left(\alpha^{(2)}_{(1)} 
	\right)} \cos^{2}{\left(\alpha^{(2)}_{(1)} \right)}}{2 \pi \nu_{(1)}^{2} 
	\left(\nu^{(2)}_{(1)}\right)^{2}} + \frac{a_{1} a_{2} \left(d^{2} - 
	d\right) \sin{\left(\beta^{(2)}_{(1)} \right)}}{2 \pi \nu_{(1)}^{3}} - 
	\frac{a_{1} a_{2} \left(d^{2} - d\right) \sin{\left(\beta^{(2)}_{(1)} 
	\right)} \cos^{2}{\left(\beta^{(2)}_{(1)} \right)}}{2 \pi 
	\left(\mu^{(2)}_{(1)}\right)^{2} \nu_{(1)}^{3}} + \frac{a_{1} d 
	\sin{\left(\beta^{(2)}_{(1)} \right)} \cos{\left(\beta^{(2)}_{(1)} 
	\right)}}{2 \pi \left(\mu^{(2)}_{(1)}\right)^{2} \nu_{(1)}^{2}} + 
	\frac{a_{2}^{2} d \cos{\left(\alpha^{(2)}_{(1)} \right)}}{2 \pi 
	\nu_{(1)}^{2} \nu^{(2)}_{(1)}} + \frac{a_{2}^{2} \left(d^{2} - 2 d\right) 
	\sin{\left(\alpha^{(2)}_{(1)} \right)}}{2 \pi \nu_{(1)}^{2} 
	\left(\nu^{(2)}_{(1)}\right)^{2}} - \frac{a_{2}^{2} \left(d^{2} - d\right) 
	\sin{\left(\alpha^{(2)}_{(1)} \right)} \cos{\left(\alpha^{(2)}_{(1)} 
	\right)}}{2 \pi \nu_{(1)}^{2} \left(\nu^{(2)}_{(1)}\right)^{2}} + 
	\frac{a_{2} d \sin{\left(\beta^{(1)}_{(1)} \right)} 
	\cos{\left(\beta^{(1)}_{(1)} \right)}}{2 \pi 
	\left(\mu^{(1)}_{(1)}\right)^{2} \nu_{(1)}^{2}} + \frac{\alpha^{(2)}_{(1)} 
	d}{2 \pi} - \frac{d}{2},
\end{dmath*}
\begin{dmath*}
	T^{\fs}_{1,3} = - \frac{a_{1}^{2} \left(d - 2\right) 
	\cos{\left(\alpha^{(2)}_{(1)} \right)}}{2 \pi \nu_{(1)}^{2} 
	\nu^{(2)}_{(1)}} - \frac{a_{1}^{2} \left(d - 2\right) 
	\sin{\left(\alpha^{(2)}_{(1)} \right)} \cos{\left(\alpha^{(2)}_{(1)} 
	\right)}}{2 \pi \nu_{(1)}^{2} \left(\nu^{(2)}_{(1)}\right)^{2}} - 
	\frac{a_{1} a_{2} \left(d^{2} - 3 d + 2\right) 
	\sin{\left(\alpha^{(2)}_{(1)} \right)}}{2 \pi \nu_{(1)}^{2}} + \frac{a_{1} 
	a_{2} \left(d - 2\right) \cos{\left(\alpha^{(2)}_{(1)} \right)}}{\pi 
	\nu_{(1)}^{2} \nu^{(2)}_{(1)}} + \frac{a_{1} a_{2} \left(d - 2\right) 
	\sin{\left(\alpha^{(2)}_{(1)} \right)}}{2 \pi \nu_{(1)}^{2} 
	\left(\nu^{(2)}_{(1)}\right)^{2}} - \frac{a_{1} a_{2} \left(d^{2} - 4 d + 
	4\right) \sin{\left(\alpha^{(2)}_{(1)} \right)} 
	\cos{\left(\alpha^{(2)}_{(1)} \right)}}{2 \pi \nu_{(1)}^{2} 
	\left(\nu^{(2)}_{(1)}\right)^{2}} + \frac{a_{1} a_{2} \left(d^{2} - 3 d + 
	2\right) \sin{\left(\alpha^{(2)}_{(1)} \right)} 
	\cos^{2}{\left(\alpha^{(2)}_{(1)} \right)}}{2 \pi \nu_{(1)}^{2} 
	\left(\nu^{(2)}_{(1)}\right)^{2}} + \frac{a_{1} a_{2} \left(d - 2\right) 
	\sin{\left(\beta^{(1)}_{(1)} \right)}}{2 \pi \nu_{(1)}^{3}} + \frac{a_{1} 
	a_{2} \left(d^{2} - 3 d + 2\right) \sin{\left(\beta^{(2)}_{(1)} 
	\right)}}{2 \pi \nu_{(1)}^{3}} - \frac{a_{1} a_{2} \left(d^{2} - 3 d + 
	2\right) \sin{\left(\beta^{(2)}_{(1)} \right)} 
	\cos^{2}{\left(\beta^{(2)}_{(1)} \right)}}{2 \pi 
	\left(\mu^{(2)}_{(1)}\right)^{2} \nu_{(1)}^{3}} - \frac{a_{1} a_{2} 
	\left(d - 2\right) \sin{\left(\beta^{(1)}_{(1)} \right)} 
	\cos^{2}{\left(\beta^{(1)}_{(1)} \right)}}{2 \pi 
	\left(\mu^{(1)}_{(1)}\right)^{2} \nu_{(1)}^{3}} + \frac{a_{1} \left(d - 
	2\right) \sin{\left(\beta^{(2)}_{(1)} \right)} 
	\cos{\left(\beta^{(2)}_{(1)} \right)}}{2 \pi 
	\left(\mu^{(2)}_{(1)}\right)^{2} \nu_{(1)}^{2}} + \frac{a_{2}^{2} \left(d 
	- 2\right) \cos{\left(\alpha^{(2)}_{(1)} \right)}}{2 \pi \nu_{(1)}^{2} 
	\nu^{(2)}_{(1)}} - \frac{a_{2}^{2} \left(d - 2\right)}{\pi \nu_{(1)}^{2} 
	\nu^{(2)}_{(1)}} + \frac{a_{2}^{2} \left(d^{2} - 4 d + 4\right) 
	\sin{\left(\alpha^{(2)}_{(1)} \right)}}{2 \pi \nu_{(1)}^{2} 
	\left(\nu^{(2)}_{(1)}\right)^{2}} - \frac{a_{2}^{2} \left(d^{2} - 3 d + 
	2\right) \sin{\left(\alpha^{(2)}_{(1)} \right)} 
	\cos{\left(\alpha^{(2)}_{(1)} \right)}}{2 \pi \nu_{(1)}^{2} 
	\left(\nu^{(2)}_{(1)}\right)^{2}} + \frac{a_{2} \left(d - 2\right) 
	\sin{\left(\beta^{(1)}_{(1)} \right)} \cos{\left(\beta^{(1)}_{(1)} 
	\right)}}{2 \pi \left(\mu^{(1)}_{(1)}\right)^{2} \nu_{(1)}^{2}} - 
	\frac{\alpha^{(2)}_{(1)} \left(d - 2\right)}{2 \pi} + \frac{d}{2} - 1,
\end{dmath*}
\begin{dmath*}
	T^{\fs}_{1,4} = \frac{a_{1} \left(d - 2\right) 
	\sin{\left(\alpha^{(2)}_{(1)} \right)}}{2 \pi} + a_{1} - \frac{a_{1} 
	\left(d - 1\right) \sin{\left(\beta^{(2)}_{(1)} \right)}}{2 \pi \nu_{(1)}} 
	- \frac{a_{1} \sin{\left(\beta^{(1)}_{(1)} \right)}}{2 \pi \nu_{(1)}} - 
	\frac{a_{2} \alpha^{(2)}_{(1)} \left(d - 2\right)}{2 \pi} + \frac{a_{2} 
	\left(d - 2\right)}{2} + \frac{\beta^{(1)}_{(1)}}{2 \pi} - \frac{1}{2},
\end{dmath*}
\begin{dmath*}
	T^{\fs}_{2,1} = \frac{a_{1}^{2} \cos{\left(\alpha^{(2)}_{(1)} \right)}}{4 
	\pi \nu_{(1)}^{2} \nu^{(2)}_{(1)}} - \frac{a_{1}^{2} 
	\sin{\left(\alpha^{(2)}_{(1)} \right)} \cos{\left(\alpha^{(2)}_{(1)} 
	\right)}}{4 \pi \nu_{(1)}^{2} \left(\nu^{(2)}_{(1)}\right)^{2}} - 
	\frac{a_{1} a_{2} \left(d - 1\right) \sin{\left(\alpha^{(2)}_{(1)} 
	\right)}}{4 \pi \nu_{(1)}^{2}} - \frac{a_{1} a_{2}}{2 \pi \nu_{(1)}^{2} 
	\nu^{(2)}_{(1)}} - \frac{a_{1} a_{2} \left(d - 2\right) 
	\sin{\left(\alpha^{(2)}_{(1)} \right)} \cos{\left(\alpha^{(2)}_{(1)} 
	\right)}}{4 \pi \nu_{(1)}^{2} \left(\nu^{(2)}_{(1)}\right)^{2}} + 
	\frac{a_{1} a_{2} \left(d - 1\right) \sin{\left(\alpha^{(2)}_{(1)} 
	\right)} \cos^{2}{\left(\alpha^{(2)}_{(1)} \right)}}{4 \pi \nu_{(1)}^{2} 
	\left(\nu^{(2)}_{(1)}\right)^{2}} + \frac{a_{1} a_{2} 
	\sin{\left(\alpha^{(2)}_{(1)} \right)}}{4 \pi \nu_{(1)}^{2} 
	\left(\nu^{(2)}_{(1)}\right)^{2}} + \frac{a_{1} a_{2} \left(d - 1\right) 
	\sin{\left(\beta^{(2)}_{(1)} \right)}}{4 \pi \nu_{(1)}^{3}} + \frac{a_{1} 
	a_{2} \sin{\left(\beta^{(1)}_{(1)} \right)}}{4 \pi \nu_{(1)}^{3}} - 
	\frac{a_{1} a_{2} \left(d - 1\right) \sin{\left(\beta^{(2)}_{(1)} \right)} 
	\cos^{2}{\left(\beta^{(2)}_{(1)} \right)}}{4 \pi 
	\left(\mu^{(2)}_{(1)}\right)^{2} \nu_{(1)}^{3}} - \frac{a_{1} a_{2} 
	\sin{\left(\beta^{(1)}_{(1)} \right)} \cos^{2}{\left(\beta^{(1)}_{(1)} 
	\right)}}{4 \pi \left(\mu^{(1)}_{(1)}\right)^{2} \nu_{(1)}^{3}} + 
	\frac{a_{1} \sin{\left(\beta^{(2)}_{(1)} \right)} 
	\cos{\left(\beta^{(2)}_{(1)} \right)}}{4 \pi 
	\left(\mu^{(2)}_{(1)}\right)^{2} \nu_{(1)}^{2}} + \frac{a_{2}^{2} 
	\cos{\left(\alpha^{(2)}_{(1)} \right)}}{4 \pi \nu_{(1)}^{2} 
	\nu^{(2)}_{(1)}} + \frac{a_{2}^{2} \left(d - 2\right) 
	\sin{\left(\alpha^{(2)}_{(1)} \right)}}{4 \pi \nu_{(1)}^{2} 
	\left(\nu^{(2)}_{(1)}\right)^{2}} - \frac{a_{2}^{2} \left(d - 1\right) 
	\sin{\left(\alpha^{(2)}_{(1)} \right)} \cos{\left(\alpha^{(2)}_{(1)} 
	\right)}}{4 \pi \nu_{(1)}^{2} \left(\nu^{(2)}_{(1)}\right)^{2}} + 
	\frac{a_{2} \sin{\left(\beta^{(1)}_{(1)} \right)} 
	\cos{\left(\beta^{(1)}_{(1)} \right)}}{4 \pi 
	\left(\mu^{(1)}_{(1)}\right)^{2} \nu_{(1)}^{2}} + 
	\frac{\alpha^{(2)}_{(1)}}{4 \pi} - \frac{1}{4},
\end{dmath*}
\begin{dmath*}
	T^{\fs}_{2,2} = - \frac{a_{1}^{2}}{2 \pi \nu_{(1)}^{2} \nu^{(2)}_{(1)}} + 
	\frac{a_{1}^{2} \sin{\left(\alpha^{(2)}_{(1)} \right)}}{2 \pi 
	\nu_{(1)}^{2} \left(\nu^{(2)}_{(1)}\right)^{2}} + \frac{a_{1} a_{2} d 
	\cos{\left(\alpha^{(2)}_{(1)} \right)}}{2 \pi \nu_{(1)}^{2} 
	\nu^{(2)}_{(1)}} - \frac{a_{1} a_{2} d \sin{\left(\alpha^{(2)}_{(1)} 
	\right)} \cos{\left(\alpha^{(2)}_{(1)} \right)}}{2 \pi \nu_{(1)}^{2} 
	\left(\nu^{(2)}_{(1)}\right)^{2}} - \frac{a_{1} a_{2} \left(d - 
	2\right)}{2 \pi \nu_{(1)}^{2} \nu^{(2)}_{(1)}} + \frac{a_{1} a_{2} \left(d 
	- 2\right) \sin{\left(\alpha^{(2)}_{(1)} \right)}}{2 \pi \nu_{(1)}^{2} 
	\left(\nu^{(2)}_{(1)}\right)^{2}} + \frac{a_{2}^{2} d 
	\sin{\left(\beta^{(1)}_{(1)} \right)}}{4 \pi \nu_{(1)}^{3}} - 
	\frac{a_{2}^{2} d \sin{\left(\beta^{(1)}_{(1)} \right)} 
	\cos^{2}{\left(\beta^{(1)}_{(1)} \right)}}{4 \pi 
	\left(\mu^{(1)}_{(1)}\right)^{2} \nu_{(1)}^{3}} - \frac{a_{2}^{2} 
	\left(d^{2} - d\right) \sin{\left(\alpha^{(2)}_{(1)} \right)}}{4 \pi 
	\nu_{(1)}^{2}} - \frac{a_{2}^{2}}{2 \pi \nu_{(1)}^{2} \nu^{(2)}_{(1)}} - 
	\frac{a_{2}^{2} \left(d^{2} - 2 d\right) \sin{\left(\alpha^{(2)}_{(1)} 
	\right)} \cos{\left(\alpha^{(2)}_{(1)} \right)}}{2 \pi \nu_{(1)}^{2} 
	\left(\nu^{(2)}_{(1)}\right)^{2}} + \frac{a_{2}^{2} \left(d^{2} - d\right) 
	\sin{\left(\alpha^{(2)}_{(1)} \right)} \cos^{2}{\left(\alpha^{(2)}_{(1)} 
	\right)}}{4 \pi \nu_{(1)}^{2} \left(\nu^{(2)}_{(1)}\right)^{2}} + 
	\frac{a_{2}^{2} \left(d^{2} - 3 d + 2\right) \sin{\left(\alpha^{(2)}_{(1)} 
	\right)}}{4 \pi \nu_{(1)}^{2} \left(\nu^{(2)}_{(1)}\right)^{2}} + 
	\frac{a_{2}^{2} \left(d^{2} - d\right) \sin{\left(\beta^{(2)}_{(1)} 
	\right)}}{4 \pi \nu_{(1)}^{3}} - \frac{a_{2}^{2} \left(d^{2} - d\right) 
	\sin{\left(\beta^{(2)}_{(1)} \right)} \cos^{2}{\left(\beta^{(2)}_{(1)} 
	\right)}}{4 \pi \left(\mu^{(2)}_{(1)}\right)^{2} \nu_{(1)}^{3}} + 
	\frac{a_{2} d \sin{\left(\beta^{(2)}_{(1)} \right)} 
	\cos{\left(\beta^{(2)}_{(1)} \right)}}{2 \pi 
	\left(\mu^{(2)}_{(1)}\right)^{2} \nu_{(1)}^{2}} - \frac{\alpha^{(2)}_{(1)} 
	\left(d - 2\right)}{4 \pi} + \frac{d}{4} + \frac{\left(d - 1\right) 
	\sin{\left(\alpha^{(2)}_{(1)} \right)}}{2 \pi} - \frac{\left(d - 1\right) 
	\sin{\left(\beta^{(2)}_{(1)} \right)}}{2 \pi \nu_{(1)}} - 
	\frac{\sin{\left(\beta^{(1)}_{(1)} \right)}}{2 \pi \nu_{(1)}} - 
	\frac{\sin{\left(\beta^{(2)}_{(1)} \right)}}{2 \pi 
	\left(\mu^{(2)}_{(1)}\right)^{2} \nu_{(1)}},
\end{dmath*}
\begin{dmath*}
	T^{\fs}_{2,3} = - \frac{a_{1} a_{2} \left(d - 2\right) 
	\sin{\left(\alpha^{(2)}_{(1)} \right)} \cos{\left(\alpha^{(2)}_{(1)} 
	\right)}}{2 \pi \nu_{(1)}^{2} \left(\nu^{(2)}_{(1)}\right)^{2}} + 
	\frac{a_{1} a_{2} \left(d - 2\right) \sin{\left(\alpha^{(2)}_{(1)} 
	\right)}}{2 \pi \nu_{(1)}^{2} \left(\nu^{(2)}_{(1)}\right)^{2}} - 
	\frac{a_{2}^{2} \left(d^{2} - 3 d + 2\right) \sin{\left(\alpha^{(2)}_{(1)} 
	\right)}}{4 \pi \nu_{(1)}^{2}} + \frac{a_{2}^{2} \left(d - 2\right) 
	\cos{\left(\alpha^{(2)}_{(1)} \right)}}{2 \pi \nu_{(1)}^{2} 
	\nu^{(2)}_{(1)}} - \frac{a_{2}^{2} \left(d - 2\right)}{2 \pi \nu_{(1)}^{2} 
	\nu^{(2)}_{(1)}} + \frac{a_{2}^{2} \left(d^{2} - 5 d + 6\right) 
	\sin{\left(\alpha^{(2)}_{(1)} \right)}}{4 \pi \nu_{(1)}^{2} 
	\left(\nu^{(2)}_{(1)}\right)^{2}} - \frac{a_{2}^{2} \left(d^{2} - 4 d + 
	4\right) \sin{\left(\alpha^{(2)}_{(1)} \right)} 
	\cos{\left(\alpha^{(2)}_{(1)} \right)}}{2 \pi \nu_{(1)}^{2} 
	\left(\nu^{(2)}_{(1)}\right)^{2}} + \frac{a_{2}^{2} \left(d^{2} - 3 d + 
	2\right) \sin{\left(\alpha^{(2)}_{(1)} \right)} 
	\cos^{2}{\left(\alpha^{(2)}_{(1)} \right)}}{4 \pi \nu_{(1)}^{2} 
	\left(\nu^{(2)}_{(1)}\right)^{2}} + \frac{a_{2}^{2} \left(d - 2\right) 
	\sin{\left(\beta^{(1)}_{(1)} \right)}}{4 \pi \nu_{(1)}^{3}} + 
	\frac{a_{2}^{2} \left(d^{2} - 3 d + 2\right) \sin{\left(\beta^{(2)}_{(1)} 
	\right)}}{4 \pi \nu_{(1)}^{3}} - \frac{a_{2}^{2} \left(d^{2} - 3 d + 
	2\right) \sin{\left(\beta^{(2)}_{(1)} \right)} 
	\cos^{2}{\left(\beta^{(2)}_{(1)} \right)}}{4 \pi 
	\left(\mu^{(2)}_{(1)}\right)^{2} \nu_{(1)}^{3}} - \frac{a_{2}^{2} \left(d 
	- 2\right) \sin{\left(\beta^{(1)}_{(1)} \right)} 
	\cos^{2}{\left(\beta^{(1)}_{(1)} \right)}}{4 \pi 
	\left(\mu^{(1)}_{(1)}\right)^{2} \nu_{(1)}^{3}} + \frac{a_{2} \left(d - 
	2\right) \sin{\left(\beta^{(2)}_{(1)} \right)} 
	\cos{\left(\beta^{(2)}_{(1)} \right)}}{2 \pi 
	\left(\mu^{(2)}_{(1)}\right)^{2} \nu_{(1)}^{2}} + \frac{\alpha^{(2)}_{(1)} 
	\left(d - 2\right)}{4 \pi} - \frac{d}{4} + \frac{1}{2},
\end{dmath*}
\begin{dmath*}
	T^{\fs}_{2,4} = - \frac{a_{2} \alpha^{(2)}_{(1)} \left(d - 2\right)}{4 
	\pi} + \frac{a_{2} d}{4} + \frac{a_{2} \left(d - 2\right) 
	\sin{\left(\alpha^{(2)}_{(1)} \right)}}{4 \pi} - \frac{a_{2} \left(d - 
	1\right) \sin{\left(\beta^{(2)}_{(1)} \right)}}{4 \pi \nu_{(1)}} - 
	\frac{a_{2} \sin{\left(\beta^{(1)}_{(1)} \right)}}{4 \pi \nu_{(1)}} + 
	\frac{\beta^{(2)}_{(1)}}{4 \pi} - \frac{1}{4},
\end{dmath*}
\begin{dmath*}
	T^{\fs}_{3,1} = - \frac{a_{1}^{2} \cos{\left(\alpha^{(2)}_{(1)} 
	\right)}}{4 \pi \nu_{(1)}^{2} \nu^{(2)}_{(1)}} - \frac{a_{1}^{2} 
	\sin{\left(\alpha^{(2)}_{(1)} \right)} \cos{\left(\alpha^{(2)}_{(1)} 
	\right)}}{4 \pi \nu_{(1)}^{2} \left(\nu^{(2)}_{(1)}\right)^{2}} - 
	\frac{a_{1} a_{2} \left(d - 1\right) \sin{\left(\alpha^{(2)}_{(1)} 
	\right)}}{4 \pi \nu_{(1)}^{2}} + \frac{a_{1} a_{2} 
	\cos{\left(\alpha^{(2)}_{(1)} \right)}}{2 \pi \nu_{(1)}^{2} 
	\nu^{(2)}_{(1)}} - \frac{a_{1} a_{2} \left(d - 2\right) 
	\sin{\left(\alpha^{(2)}_{(1)} \right)} \cos{\left(\alpha^{(2)}_{(1)} 
	\right)}}{4 \pi \nu_{(1)}^{2} \left(\nu^{(2)}_{(1)}\right)^{2}} + 
	\frac{a_{1} a_{2} \left(d - 1\right) \sin{\left(\alpha^{(2)}_{(1)} 
	\right)} \cos^{2}{\left(\alpha^{(2)}_{(1)} \right)}}{4 \pi \nu_{(1)}^{2} 
	\left(\nu^{(2)}_{(1)}\right)^{2}} + \frac{a_{1} a_{2} 
	\sin{\left(\alpha^{(2)}_{(1)} \right)}}{4 \pi \nu_{(1)}^{2} 
	\left(\nu^{(2)}_{(1)}\right)^{2}} + \frac{a_{1} a_{2} \left(d - 1\right) 
	\sin{\left(\beta^{(2)}_{(1)} \right)}}{4 \pi \nu_{(1)}^{3}} + \frac{a_{1} 
	a_{2} \sin{\left(\beta^{(1)}_{(1)} \right)}}{4 \pi \nu_{(1)}^{3}} - 
	\frac{a_{1} a_{2} \left(d - 1\right) \sin{\left(\beta^{(2)}_{(1)} \right)} 
	\cos^{2}{\left(\beta^{(2)}_{(1)} \right)}}{4 \pi 
	\left(\mu^{(2)}_{(1)}\right)^{2} \nu_{(1)}^{3}} - \frac{a_{1} a_{2} 
	\sin{\left(\beta^{(1)}_{(1)} \right)} \cos^{2}{\left(\beta^{(1)}_{(1)} 
	\right)}}{4 \pi \left(\mu^{(1)}_{(1)}\right)^{2} \nu_{(1)}^{3}} + 
	\frac{a_{1} \sin{\left(\beta^{(2)}_{(1)} \right)} 
	\cos{\left(\beta^{(2)}_{(1)} \right)}}{4 \pi 
	\left(\mu^{(2)}_{(1)}\right)^{2} \nu_{(1)}^{2}} + \frac{a_{2}^{2} 
	\cos{\left(\alpha^{(2)}_{(1)} \right)}}{4 \pi \nu_{(1)}^{2} 
	\nu^{(2)}_{(1)}} - \frac{a_{2}^{2}}{2 \pi \nu_{(1)}^{2} \nu^{(2)}_{(1)}} + 
	\frac{a_{2}^{2} \left(d - 2\right) \sin{\left(\alpha^{(2)}_{(1)} 
	\right)}}{4 \pi \nu_{(1)}^{2} \left(\nu^{(2)}_{(1)}\right)^{2}} - 
	\frac{a_{2}^{2} \left(d - 1\right) \sin{\left(\alpha^{(2)}_{(1)} \right)} 
	\cos{\left(\alpha^{(2)}_{(1)} \right)}}{4 \pi \nu_{(1)}^{2} 
	\left(\nu^{(2)}_{(1)}\right)^{2}} + \frac{a_{2} 
	\sin{\left(\beta^{(1)}_{(1)} \right)} \cos{\left(\beta^{(1)}_{(1)} 
	\right)}}{4 \pi \left(\mu^{(1)}_{(1)}\right)^{2} \nu_{(1)}^{2}} - 
	\frac{\alpha^{(2)}_{(1)}}{4 \pi} + \frac{1}{4},
\end{dmath*}
\begin{dmath*}
	T^{\fs}_{3,2} = - \frac{a_{1} a_{2} d \sin{\left(\alpha^{(2)}_{(1)} 
	\right)} \cos{\left(\alpha^{(2)}_{(1)} \right)}}{2 \pi \nu_{(1)}^{2} 
	\left(\nu^{(2)}_{(1)}\right)^{2}} + \frac{a_{1} a_{2} d 
	\sin{\left(\alpha^{(2)}_{(1)} \right)}}{2 \pi \nu_{(1)}^{2} 
	\left(\nu^{(2)}_{(1)}\right)^{2}} + \frac{a_{2}^{2} d 
	\cos{\left(\alpha^{(2)}_{(1)} \right)}}{2 \pi \nu_{(1)}^{2} 
	\nu^{(2)}_{(1)}} - \frac{a_{2}^{2} d}{2 \pi \nu_{(1)}^{2} \nu^{(2)}_{(1)}} 
	+ \frac{a_{2}^{2} d \sin{\left(\beta^{(1)}_{(1)} \right)}}{4 \pi 
	\nu_{(1)}^{3}} - \frac{a_{2}^{2} d \sin{\left(\beta^{(1)}_{(1)} \right)} 
	\cos^{2}{\left(\beta^{(1)}_{(1)} \right)}}{4 \pi 
	\left(\mu^{(1)}_{(1)}\right)^{2} \nu_{(1)}^{3}} - \frac{a_{2}^{2} 
	\left(d^{2} - d\right) \sin{\left(\alpha^{(2)}_{(1)} \right)}}{4 \pi 
	\nu_{(1)}^{2}} + \frac{a_{2}^{2} \left(d^{2} - 3 d\right) 
	\sin{\left(\alpha^{(2)}_{(1)} \right)}}{4 \pi \nu_{(1)}^{2} 
	\left(\nu^{(2)}_{(1)}\right)^{2}} - \frac{a_{2}^{2} \left(d^{2} - 2 
	d\right) \sin{\left(\alpha^{(2)}_{(1)} \right)} 
	\cos{\left(\alpha^{(2)}_{(1)} \right)}}{2 \pi \nu_{(1)}^{2} 
	\left(\nu^{(2)}_{(1)}\right)^{2}} + \frac{a_{2}^{2} \left(d^{2} - d\right) 
	\sin{\left(\alpha^{(2)}_{(1)} \right)} \cos^{2}{\left(\alpha^{(2)}_{(1)} 
	\right)}}{4 \pi \nu_{(1)}^{2} \left(\nu^{(2)}_{(1)}\right)^{2}} + 
	\frac{a_{2}^{2} \left(d^{2} - d\right) \sin{\left(\beta^{(2)}_{(1)} 
	\right)}}{4 \pi \nu_{(1)}^{3}} - \frac{a_{2}^{2} \left(d^{2} - d\right) 
	\sin{\left(\beta^{(2)}_{(1)} \right)} \cos^{2}{\left(\beta^{(2)}_{(1)} 
	\right)}}{4 \pi \left(\mu^{(2)}_{(1)}\right)^{2} \nu_{(1)}^{3}} + 
	\frac{a_{2} d \sin{\left(\beta^{(2)}_{(1)} \right)} 
	\cos{\left(\beta^{(2)}_{(1)} \right)}}{2 \pi 
	\left(\mu^{(2)}_{(1)}\right)^{2} \nu_{(1)}^{2}} + \frac{\alpha^{(2)}_{(1)} 
	d}{4 \pi} - \frac{d}{4},
\end{dmath*}
\begin{dmath*}
	T^{\fs}_{3,3} = \frac{a_{1}^{2}}{2 \pi \nu_{(1)}^{2} \nu^{(2)}_{(1)}} + 
	\frac{a_{1}^{2} \sin{\left(\alpha^{(2)}_{(1)} \right)}}{2 \pi 
	\nu_{(1)}^{2} \left(\nu^{(2)}_{(1)}\right)^{2}} + \frac{a_{1} a_{2} 
	\left(d - 4\right)}{2 \pi \nu_{(1)}^{2} \nu^{(2)}_{(1)}} - \frac{a_{1} 
	a_{2} \left(d - 2\right) \cos{\left(\alpha^{(2)}_{(1)} \right)}}{2 \pi 
	\nu_{(1)}^{2} \nu^{(2)}_{(1)}} + \frac{a_{1} a_{2} \left(d - 4\right) 
	\sin{\left(\alpha^{(2)}_{(1)} \right)}}{2 \pi \nu_{(1)}^{2} 
	\left(\nu^{(2)}_{(1)}\right)^{2}} - \frac{a_{1} a_{2} \left(d - 2\right) 
	\sin{\left(\alpha^{(2)}_{(1)} \right)} \cos{\left(\alpha^{(2)}_{(1)} 
	\right)}}{2 \pi \nu_{(1)}^{2} \left(\nu^{(2)}_{(1)}\right)^{2}} - 
	\frac{a_{2}^{2} \left(d^{2} - 3 d + 2\right) \sin{\left(\alpha^{(2)}_{(1)} 
	\right)}}{4 \pi \nu_{(1)}^{2}} - \frac{a_{2}^{2} \left(d - 
	\frac{5}{2}\right)}{\pi \nu_{(1)}^{2} \nu^{(2)}_{(1)}} + \frac{a_{2}^{2} 
	\left(d - 2\right) \cos{\left(\alpha^{(2)}_{(1)} \right)}}{\pi 
	\nu_{(1)}^{2} \nu^{(2)}_{(1)}} + \frac{a_{2}^{2} \left(d^{2} - 5 d + 
	8\right) \sin{\left(\alpha^{(2)}_{(1)} \right)}}{4 \pi \nu_{(1)}^{2} 
	\left(\nu^{(2)}_{(1)}\right)^{2}} - \frac{a_{2}^{2} \left(d^{2} - 4 d + 
	4\right) \sin{\left(\alpha^{(2)}_{(1)} \right)} 
	\cos{\left(\alpha^{(2)}_{(1)} \right)}}{2 \pi \nu_{(1)}^{2} 
	\left(\nu^{(2)}_{(1)}\right)^{2}} + \frac{a_{2}^{2} \left(d^{2} - 3 d + 
	2\right) \sin{\left(\alpha^{(2)}_{(1)} \right)} 
	\cos^{2}{\left(\alpha^{(2)}_{(1)} \right)}}{4 \pi \nu_{(1)}^{2} 
	\left(\nu^{(2)}_{(1)}\right)^{2}} + \frac{a_{2}^{2} \left(d - 2\right) 
	\sin{\left(\beta^{(1)}_{(1)} \right)}}{4 \pi \nu_{(1)}^{3}} + 
	\frac{a_{2}^{2} \left(d^{2} - 3 d + 2\right) \sin{\left(\beta^{(2)}_{(1)} 
	\right)}}{4 \pi \nu_{(1)}^{3}} - \frac{a_{2}^{2} \left(d^{2} - 3 d + 
	2\right) \sin{\left(\beta^{(2)}_{(1)} \right)} 
	\cos^{2}{\left(\beta^{(2)}_{(1)} \right)}}{4 \pi 
	\left(\mu^{(2)}_{(1)}\right)^{2} \nu_{(1)}^{3}} - \frac{a_{2}^{2} \left(d 
	- 2\right) \sin{\left(\beta^{(1)}_{(1)} \right)} 
	\cos^{2}{\left(\beta^{(1)}_{(1)} \right)}}{4 \pi 
	\left(\mu^{(1)}_{(1)}\right)^{2} \nu_{(1)}^{3}} + \frac{a_{2} \left(d - 
	2\right) \sin{\left(\beta^{(2)}_{(1)} \right)} 
	\cos{\left(\beta^{(2)}_{(1)} \right)}}{2 \pi 
	\left(\mu^{(2)}_{(1)}\right)^{2} \nu_{(1)}^{2}} - \frac{\alpha^{(2)}_{(1)} 
	\left(d - 4\right)}{4 \pi} + \frac{d}{4} + \frac{\left(d - 1\right) 
	\sin{\left(\alpha^{(2)}_{(1)} \right)}}{2 \pi} - \frac{1}{2} - 
	\frac{\left(d - 1\right) \sin{\left(\beta^{(2)}_{(1)} \right)}}{2 \pi 
	\nu_{(1)}} - \frac{\sin{\left(\beta^{(1)}_{(1)} \right)}}{2 \pi \nu_{(1)}} 
	- \frac{\sin{\left(\beta^{(2)}_{(1)} \right)}}{2 \pi 
	\left(\mu^{(2)}_{(1)}\right)^{2} \nu_{(1)}},
\end{dmath*}
\begin{dmath*}
	T^{\fs}_{3,4} = - \frac{a_{1} \alpha^{(2)}_{(1)}}{2 \pi} + \frac{a_{1}}{2} 
	- \frac{a_{2} \alpha^{(2)}_{(1)} \left(d - 4\right)}{4 \pi} + \frac{a_{2} 
	\left(d - 2\right) \sin{\left(\alpha^{(2)}_{(1)} \right)}}{4 \pi} + 
	\frac{a_{2} \left(d - 2\right)}{4} - \frac{a_{2} \left(d - 1\right) 
	\sin{\left(\beta^{(2)}_{(1)} \right)}}{4 \pi \nu_{(1)}} - \frac{a_{2} 
	\sin{\left(\beta^{(1)}_{(1)} \right)}}{4 \pi \nu_{(1)}} + 
	\frac{\beta^{(2)}_{(1)}}{4 \pi} - \frac{1}{4},
\end{dmath*}
\begin{dmath*}
	T^{\fs}_{4,1} = \frac{a_{1} \left(d - 2\right) 
	\sin{\left(\alpha^{(2)}_{(1)} \right)}}{2 \pi} + a_{1} - \frac{a_{1} 
	\left(d - 1\right) \sin{\left(\beta^{(2)}_{(1)} \right)}}{2 \pi \nu_{(1)}} 
	- \frac{a_{1} \sin{\left(\beta^{(1)}_{(1)} \right)}}{2 \pi \nu_{(1)}} - 
	\frac{a_{2} \alpha^{(2)}_{(1)} \left(d - 2\right)}{2 \pi} + \frac{a_{2} 
	\left(d - 2\right)}{2} + \frac{\beta^{(1)}_{(1)}}{2 \pi} - \frac{1}{2},
\end{dmath*}
\begin{dmath*}
	T^{\fs}_{4,2} = - \frac{a_{2} \alpha^{(2)}_{(1)} \left(d^{2} - 2 
	d\right)}{2 \pi} + \frac{a_{2} d^{2}}{2} - \frac{a_{2} d 
	\sin{\left(\beta^{(1)}_{(1)} \right)}}{2 \pi \nu_{(1)}} + \frac{a_{2} 
	\left(d^{2} - 2 d\right) \sin{\left(\alpha^{(2)}_{(1)} \right)}}{2 \pi} - 
	\frac{a_{2} \left(d^{2} - d\right) \sin{\left(\beta^{(2)}_{(1)} 
	\right)}}{2 \pi \nu_{(1)}} + \frac{\beta^{(2)}_{(1)} d}{2 \pi} - 
	\frac{d}{2},
\end{dmath*}
\begin{dmath*}
	T^{\fs}_{4,3} = - \frac{a_{1} \alpha^{(2)}_{(1)} \left(d - 2\right)}{\pi} 
	+ a_{1} \left(d - 2\right) - \frac{a_{2} \alpha^{(2)}_{(1)} \left(d^{2} - 
	6 d + 8\right)}{2 \pi} + \frac{a_{2} \left(d^{2} - 4 d + 4\right) 
	\sin{\left(\alpha^{(2)}_{(1)} \right)}}{2 \pi} + \frac{a_{2} \left(d^{2} - 
	4 d + 4\right)}{2} - \frac{a_{2} \left(d - 2\right) 
	\sin{\left(\beta^{(1)}_{(1)} \right)}}{2 \pi \nu_{(1)}} - \frac{a_{2} 
	\left(d^{2} - 3 d + 2\right) \sin{\left(\beta^{(2)}_{(1)} \right)}}{2 \pi 
	\nu_{(1)}} + \frac{\beta^{(2)}_{(1)} \left(d - 2\right)}{2 \pi} - 
	\frac{d}{2} + 1,
\end{dmath*}
\begin{dmath*}
	T^{\fs}_{4,4} = \frac{\alpha^{(2)}_{(1)} \nu_{(1)}^{2} 
	\cos{\left(\alpha^{(2)}_{(1)} \right)}}{2 \pi} - \frac{\nu_{(1)}^{2} 
	\sin{\left(\alpha^{(2)}_{(1)} \right)}}{2 \pi} - \frac{\nu_{(1)}^{2} 
	\cos{\left(\alpha^{(2)}_{(1)} \right)}}{2} + \frac{\nu_{(1)}^{2}}{2}.
\end{dmath*}

\paragraph{$\ft$-rep.} Let the associated $3\times3$ transition matrix be 
denoted by $T^\ft$. Then,
\begin{dmath*}
	T^{\ft}_{1,1} = - \frac{a_{1}^{2} \left(d - 1\right) 
	\sin{\left(\alpha^{(2)}_{(1)} \right)}}{2 \pi \nu_{(1)}^{2}} + 
	\frac{a_{1}^{2} \left(d - 1\right)}{2 \pi \nu_{(1)}^{2} \nu^{(2)}_{(1)}} + 
	\frac{a_{1}^{2} \left(d - 1\right) \sin{\left(\alpha^{(2)}_{(1)} \right)} 
	\cos^{2}{\left(\alpha^{(2)}_{(1)} \right)}}{2 \pi \nu_{(1)}^{2} 
	\left(\nu^{(2)}_{(1)}\right)^{2}} + \frac{a_{1}^{2} \left(d - 1\right) 
	\sin{\left(\beta^{(2)}_{(1)} \right)}}{2 \pi \nu_{(1)}^{3}} + 
	\frac{a_{1}^{2} \sin{\left(\beta^{(1)}_{(1)} \right)}}{2 \pi 
	\nu_{(1)}^{3}} - \frac{a_{1}^{2} \left(d - 1\right) 
	\sin{\left(\beta^{(2)}_{(1)} \right)} \cos^{2}{\left(\beta^{(2)}_{(1)} 
	\right)}}{2 \pi \left(\mu^{(2)}_{(1)}\right)^{2} \nu_{(1)}^{3}} - 
	\frac{a_{1}^{2} \sin{\left(\beta^{(1)}_{(1)} \right)} 
	\cos^{2}{\left(\beta^{(1)}_{(1)} \right)}}{2 \pi 
	\left(\mu^{(1)}_{(1)}\right)^{2} \nu_{(1)}^{3}} - \frac{a_{1} a_{2} 
	\left(d - 1\right) \cos{\left(\alpha^{(2)}_{(1)} \right)}}{\pi 
	\nu_{(1)}^{2} \nu^{(2)}_{(1)}} - \frac{a_{1} a_{2} \left(d - 1\right) 
	\sin{\left(\alpha^{(2)}_{(1)} \right)} \cos{\left(\alpha^{(2)}_{(1)} 
	\right)}}{\pi \nu_{(1)}^{2} \left(\nu^{(2)}_{(1)}\right)^{2}} + 
	\frac{a_{1} \sin{\left(\beta^{(1)}_{(1)} \right)} 
	\cos{\left(\beta^{(1)}_{(1)} \right)}}{\pi 
	\left(\mu^{(1)}_{(1)}\right)^{2} \nu_{(1)}^{2}} + \frac{a_{2}^{2} \left(d 
	- 1\right)}{2 \pi \nu_{(1)}^{2} \nu^{(2)}_{(1)}} + \frac{a_{2}^{2} \left(d 
	- 1\right) \sin{\left(\alpha^{(2)}_{(1)} \right)}}{2 \pi \nu_{(1)}^{2} 
	\left(\nu^{(2)}_{(1)}\right)^{2}} + \frac{\left(d - 1\right) 
	\sin{\left(\alpha^{(2)}_{(1)} \right)}}{2 \pi} + \frac{1}{2} - 
	\frac{\left(d - 1\right) \sin{\left(\beta^{(2)}_{(1)} \right)}}{2 \pi 
	\nu_{(1)}} - \frac{\sin{\left(\beta^{(1)}_{(1)} \right)}}{2 \pi \nu_{(1)}} 
	- \frac{\sin{\left(\beta^{(1)}_{(1)} \right)}}{2 \pi 
	\left(\mu^{(1)}_{(1)}\right)^{2} \nu_{(1)}},
\end{dmath*}
\begin{dmath*}
	T^{\ft}_{1,2} = - \frac{a_{1}^{2} \left(d - 1\right) 
	\cos{\left(\alpha^{(2)}_{(1)} \right)}}{2 \pi \nu_{(1)}^{2} 
	\nu^{(2)}_{(1)}} - \frac{a_{1}^{2} \left(d - 1\right) 
	\sin{\left(\alpha^{(2)}_{(1)} \right)} \cos{\left(\alpha^{(2)}_{(1)} 
	\right)}}{2 \pi \nu_{(1)}^{2} \left(\nu^{(2)}_{(1)}\right)^{2}} - 
	\frac{a_{1} a_{2} \left(d^{2} - 2 d + 1\right) 
	\sin{\left(\alpha^{(2)}_{(1)} \right)}}{2 \pi \nu_{(1)}^{2}} + \frac{a_{1} 
	a_{2} \left(d^{2} - d\right)}{2 \pi \nu_{(1)}^{2} \nu^{(2)}_{(1)}} - 
	\frac{a_{1} a_{2} \left(d^{2} - 3 d + 2\right) 
	\cos{\left(\alpha^{(2)}_{(1)} \right)}}{2 \pi \nu_{(1)}^{2} 
	\nu^{(2)}_{(1)}} + \frac{a_{1} a_{2} \left(d - 1\right) 
	\sin{\left(\alpha^{(2)}_{(1)} \right)}}{2 \pi \nu_{(1)}^{2} 
	\left(\nu^{(2)}_{(1)}\right)^{2}} - \frac{a_{1} a_{2} \left(d^{2} - 3 d + 
	2\right) \sin{\left(\alpha^{(2)}_{(1)} \right)} 
	\cos{\left(\alpha^{(2)}_{(1)} \right)}}{2 \pi \nu_{(1)}^{2} 
	\left(\nu^{(2)}_{(1)}\right)^{2}} + \frac{a_{1} a_{2} \left(d^{2} - 2 d + 
	1\right) \sin{\left(\alpha^{(2)}_{(1)} \right)} 
	\cos^{2}{\left(\alpha^{(2)}_{(1)} \right)}}{2 \pi \nu_{(1)}^{2} 
	\left(\nu^{(2)}_{(1)}\right)^{2}} + \frac{a_{1} a_{2} \left(d - 1\right) 
	\sin{\left(\beta^{(1)}_{(1)} \right)}}{2 \pi \nu_{(1)}^{3}} + \frac{a_{1} 
	a_{2} \left(d^{2} - 2 d + 1\right) \sin{\left(\beta^{(2)}_{(1)} 
	\right)}}{2 \pi \nu_{(1)}^{3}} - \frac{a_{1} a_{2} \left(d^{2} - 2 d + 
	1\right) \sin{\left(\beta^{(2)}_{(1)} \right)} 
	\cos^{2}{\left(\beta^{(2)}_{(1)} \right)}}{2 \pi 
	\left(\mu^{(2)}_{(1)}\right)^{2} \nu_{(1)}^{3}} - \frac{a_{1} a_{2} 
	\left(d - 1\right) \sin{\left(\beta^{(1)}_{(1)} \right)} 
	\cos^{2}{\left(\beta^{(1)}_{(1)} \right)}}{2 \pi 
	\left(\mu^{(1)}_{(1)}\right)^{2} \nu_{(1)}^{3}} + \frac{a_{1} \left(d - 
	1\right) \sin{\left(\beta^{(2)}_{(1)} \right)} 
	\cos{\left(\beta^{(2)}_{(1)} \right)}}{2 \pi 
	\left(\mu^{(2)}_{(1)}\right)^{2} \nu_{(1)}^{2}} + \frac{a_{2}^{2} 
	\left(d^{2} - 3 d + 2\right)}{2 \pi \nu_{(1)}^{2} \nu^{(2)}_{(1)}} - 
	\frac{a_{2}^{2} \left(d^{2} - 2 d + 1\right) \cos{\left(\alpha^{(2)}_{(1)} 
	\right)}}{2 \pi \nu_{(1)}^{2} \nu^{(2)}_{(1)}} + \frac{a_{2}^{2} 
	\left(d^{2} - 3 d + 2\right) \sin{\left(\alpha^{(2)}_{(1)} \right)}}{2 \pi 
	\nu_{(1)}^{2} \left(\nu^{(2)}_{(1)}\right)^{2}} - \frac{a_{2}^{2} 
	\left(d^{2} - 2 d + 1\right) \sin{\left(\alpha^{(2)}_{(1)} \right)} 
	\cos{\left(\alpha^{(2)}_{(1)} \right)}}{2 \pi \nu_{(1)}^{2} 
	\left(\nu^{(2)}_{(1)}\right)^{2}} + \frac{a_{2} \left(d - 1\right) 
	\sin{\left(\beta^{(1)}_{(1)} \right)} \cos{\left(\beta^{(1)}_{(1)} 
	\right)}}{2 \pi \left(\mu^{(1)}_{(1)}\right)^{2} \nu_{(1)}^{2}} - 
	\frac{\alpha^{(2)}_{(1)} \left(d - 1\right)}{2 \pi} + \frac{d}{2} - 
	\frac{1}{2},
\end{dmath*}
\begin{dmath*}
	T^{\ft}_{1,3} = \frac{a_{1} \left(d - 1\right) 
	\sin{\left(\alpha^{(2)}_{(1)} \right)}}{\pi} + a_{1} - \frac{a_{1} \left(d 
	- 1\right) \sin{\left(\beta^{(2)}_{(1)} \right)}}{2 \pi \nu_{(1)}} - 
	\frac{a_{1} \sin{\left(\beta^{(1)}_{(1)} \right)}}{2 \pi \nu_{(1)}} - 
	\frac{a_{2} \alpha^{(2)}_{(1)} \left(d - 1\right)}{\pi} + a_{2} \left(d - 
	1\right) + \frac{\beta^{(1)}_{(1)}}{2 \pi} - \frac{1}{2},
\end{dmath*}
\begin{dmath*}
	T^{\ft}_{2,1} = - \frac{a_{1}^{2} \cos{\left(\alpha^{(2)}_{(1)} 
	\right)}}{2 \pi \nu_{(1)}^{2} \nu^{(2)}_{(1)}} - \frac{a_{1}^{2} 
	\sin{\left(\alpha^{(2)}_{(1)} \right)} \cos{\left(\alpha^{(2)}_{(1)} 
	\right)}}{2 \pi \nu_{(1)}^{2} \left(\nu^{(2)}_{(1)}\right)^{2}} + 
	\frac{a_{1} a_{2} d}{2 \pi \nu_{(1)}^{2} \nu^{(2)}_{(1)}} - \frac{a_{1} 
	a_{2} \left(d - 1\right) \sin{\left(\alpha^{(2)}_{(1)} \right)}}{2 \pi 
	\nu_{(1)}^{2}} - \frac{a_{1} a_{2} \left(d - 2\right) 
	\cos{\left(\alpha^{(2)}_{(1)} \right)}}{2 \pi \nu_{(1)}^{2} 
	\nu^{(2)}_{(1)}} - \frac{a_{1} a_{2} \left(d - 2\right) 
	\sin{\left(\alpha^{(2)}_{(1)} \right)} \cos{\left(\alpha^{(2)}_{(1)} 
	\right)}}{2 \pi \nu_{(1)}^{2} \left(\nu^{(2)}_{(1)}\right)^{2}} + 
	\frac{a_{1} a_{2} \left(d - 1\right) \sin{\left(\alpha^{(2)}_{(1)} 
	\right)} \cos^{2}{\left(\alpha^{(2)}_{(1)} \right)}}{2 \pi \nu_{(1)}^{2} 
	\left(\nu^{(2)}_{(1)}\right)^{2}} + \frac{a_{1} a_{2} 
	\sin{\left(\alpha^{(2)}_{(1)} \right)}}{2 \pi \nu_{(1)}^{2} 
	\left(\nu^{(2)}_{(1)}\right)^{2}} + \frac{a_{1} a_{2} \left(d - 1\right) 
	\sin{\left(\beta^{(2)}_{(1)} \right)}}{2 \pi \nu_{(1)}^{3}} + \frac{a_{1} 
	a_{2} \sin{\left(\beta^{(1)}_{(1)} \right)}}{2 \pi \nu_{(1)}^{3}} - 
	\frac{a_{1} a_{2} \left(d - 1\right) \sin{\left(\beta^{(2)}_{(1)} \right)} 
	\cos^{2}{\left(\beta^{(2)}_{(1)} \right)}}{2 \pi 
	\left(\mu^{(2)}_{(1)}\right)^{2} \nu_{(1)}^{3}} - \frac{a_{1} a_{2} 
	\sin{\left(\beta^{(1)}_{(1)} \right)} \cos^{2}{\left(\beta^{(1)}_{(1)} 
	\right)}}{2 \pi \left(\mu^{(1)}_{(1)}\right)^{2} \nu_{(1)}^{3}} + 
	\frac{a_{1} \sin{\left(\beta^{(2)}_{(1)} \right)} 
	\cos{\left(\beta^{(2)}_{(1)} \right)}}{2 \pi 
	\left(\mu^{(2)}_{(1)}\right)^{2} \nu_{(1)}^{2}} + \frac{a_{2}^{2} \left(d 
	- 2\right)}{2 \pi \nu_{(1)}^{2} \nu^{(2)}_{(1)}} - \frac{a_{2}^{2} \left(d 
	- 1\right) \cos{\left(\alpha^{(2)}_{(1)} \right)}}{2 \pi \nu_{(1)}^{2} 
	\nu^{(2)}_{(1)}} + \frac{a_{2}^{2} \left(d - 2\right) 
	\sin{\left(\alpha^{(2)}_{(1)} \right)}}{2 \pi \nu_{(1)}^{2} 
	\left(\nu^{(2)}_{(1)}\right)^{2}} - \frac{a_{2}^{2} \left(d - 1\right) 
	\sin{\left(\alpha^{(2)}_{(1)} \right)} \cos{\left(\alpha^{(2)}_{(1)} 
	\right)}}{2 \pi \nu_{(1)}^{2} \left(\nu^{(2)}_{(1)}\right)^{2}} + 
	\frac{a_{2} \sin{\left(\beta^{(1)}_{(1)} \right)} 
	\cos{\left(\beta^{(1)}_{(1)} \right)}}{2 \pi 
	\left(\mu^{(1)}_{(1)}\right)^{2} \nu_{(1)}^{2}} - 
	\frac{\alpha^{(2)}_{(1)}}{2 \pi} + \frac{1}{2},
\end{dmath*}
\begin{dmath*}
	T^{\ft}_{2,2} = \frac{a_{1}^{2}}{2 \pi \nu_{(1)}^{2} \nu^{(2)}_{(1)}} + 
	\frac{a_{1}^{2} \sin{\left(\alpha^{(2)}_{(1)} \right)}}{2 \pi 
	\nu_{(1)}^{2} \left(\nu^{(2)}_{(1)}\right)^{2}} + \frac{a_{1} a_{2} 
	\left(d - 2\right)}{\pi \nu_{(1)}^{2} \nu^{(2)}_{(1)}} - \frac{a_{1} a_{2} 
	\left(d - 1\right) \cos{\left(\alpha^{(2)}_{(1)} \right)}}{\pi 
	\nu_{(1)}^{2} \nu^{(2)}_{(1)}} + \frac{a_{1} a_{2} \left(d - 2\right) 
	\sin{\left(\alpha^{(2)}_{(1)} \right)}}{\pi \nu_{(1)}^{2} 
	\left(\nu^{(2)}_{(1)}\right)^{2}} - \frac{a_{1} a_{2} \left(d - 1\right) 
	\sin{\left(\alpha^{(2)}_{(1)} \right)} \cos{\left(\alpha^{(2)}_{(1)} 
	\right)}}{\pi \nu_{(1)}^{2} \left(\nu^{(2)}_{(1)}\right)^{2}} - 
	\frac{a_{2}^{2} \left(d^{2} - 2 d + 1\right) \sin{\left(\alpha^{(2)}_{(1)} 
	\right)}}{2 \pi \nu_{(1)}^{2}} - \frac{a_{2}^{2} \left(d^{2} - 3 d + 
	2\right) \cos{\left(\alpha^{(2)}_{(1)} \right)}}{\pi \nu_{(1)}^{2} 
	\nu^{(2)}_{(1)}} + \frac{a_{2}^{2} \left(d^{2} - 3 d + 
	\frac{5}{2}\right)}{\pi \nu_{(1)}^{2} \nu^{(2)}_{(1)}} + \frac{a_{2}^{2} 
	\left(d^{2} - 4 d + 4\right) \sin{\left(\alpha^{(2)}_{(1)} \right)}}{2 \pi 
	\nu_{(1)}^{2} \left(\nu^{(2)}_{(1)}\right)^{2}} - \frac{a_{2}^{2} 
	\left(d^{2} - 3 d + 2\right) \sin{\left(\alpha^{(2)}_{(1)} \right)} 
	\cos{\left(\alpha^{(2)}_{(1)} \right)}}{\pi \nu_{(1)}^{2} 
	\left(\nu^{(2)}_{(1)}\right)^{2}} + \frac{a_{2}^{2} \left(d^{2} - 2 d + 
	1\right) \sin{\left(\alpha^{(2)}_{(1)} \right)} 
	\cos^{2}{\left(\alpha^{(2)}_{(1)} \right)}}{2 \pi \nu_{(1)}^{2} 
	\left(\nu^{(2)}_{(1)}\right)^{2}} + \frac{a_{2}^{2} \left(d - 1\right) 
	\sin{\left(\beta^{(1)}_{(1)} \right)}}{2 \pi \nu_{(1)}^{3}} + 
	\frac{a_{2}^{2} \left(d^{2} - 2 d + 1\right) \sin{\left(\beta^{(2)}_{(1)} 
	\right)}}{2 \pi \nu_{(1)}^{3}} - \frac{a_{2}^{2} \left(d^{2} - 2 d + 
	1\right) \sin{\left(\beta^{(2)}_{(1)} \right)} 
	\cos^{2}{\left(\beta^{(2)}_{(1)} \right)}}{2 \pi 
	\left(\mu^{(2)}_{(1)}\right)^{2} \nu_{(1)}^{3}} - \frac{a_{2}^{2} \left(d 
	- 1\right) \sin{\left(\beta^{(1)}_{(1)} \right)} 
	\cos^{2}{\left(\beta^{(1)}_{(1)} \right)}}{2 \pi 
	\left(\mu^{(1)}_{(1)}\right)^{2} \nu_{(1)}^{3}} + \frac{a_{2} \left(d - 
	1\right) \sin{\left(\beta^{(2)}_{(1)} \right)} 
	\cos{\left(\beta^{(2)}_{(1)} \right)}}{\pi 
	\left(\mu^{(2)}_{(1)}\right)^{2} \nu_{(1)}^{2}} - \frac{\alpha^{(2)}_{(1)} 
	\left(d - 2\right)}{2 \pi} + \frac{d}{2} + \frac{\left(d - 1\right) 
	\sin{\left(\alpha^{(2)}_{(1)} \right)}}{2 \pi} - \frac{1}{2} - 
	\frac{\left(d - 1\right) \sin{\left(\beta^{(2)}_{(1)} \right)}}{2 \pi 
	\nu_{(1)}} - \frac{\sin{\left(\beta^{(1)}_{(1)} \right)}}{2 \pi \nu_{(1)}} 
	- \frac{\sin{\left(\beta^{(2)}_{(1)} \right)}}{2 \pi 
	\left(\mu^{(2)}_{(1)}\right)^{2} \nu_{(1)}},
\end{dmath*}
\begin{dmath*}
	T^{\ft}_{2,3} = - \frac{a_{1} \alpha^{(2)}_{(1)}}{\pi} + a_{1} - 
	\frac{a_{2} \alpha^{(2)}_{(1)} \left(d - 2\right)}{\pi} + \frac{a_{2} 
	\left(d - 1\right) \sin{\left(\alpha^{(2)}_{(1)} \right)}}{\pi} + a_{2} 
	\left(d - 1\right) - \frac{a_{2} \left(d - 1\right) 
	\sin{\left(\beta^{(2)}_{(1)} \right)}}{2 \pi \nu_{(1)}} - \frac{a_{2} 
	\sin{\left(\beta^{(1)}_{(1)} \right)}}{2 \pi \nu_{(1)}} + 
	\frac{\beta^{(2)}_{(1)}}{2 \pi} - \frac{1}{2},
\end{dmath*}
\begin{dmath*}
	T^{\ft}_{3,1} = \frac{a_{1} \left(d - 1\right) 
	\sin{\left(\alpha^{(2)}_{(1)} \right)}}{\pi} + a_{1} - \frac{a_{1} \left(d 
	- 1\right) \sin{\left(\beta^{(2)}_{(1)} \right)}}{2 \pi \nu_{(1)}} - 
	\frac{a_{1} \sin{\left(\beta^{(1)}_{(1)} \right)}}{2 \pi \nu_{(1)}} - 
	\frac{a_{2} \alpha^{(2)}_{(1)} \left(d - 1\right)}{\pi} + a_{2} \left(d - 
	1\right) + \frac{\beta^{(1)}_{(1)}}{2 \pi} - \frac{1}{2},
\end{dmath*}
\begin{dmath*}
	T^{\ft}_{3,2} = - \frac{a_{1} \alpha^{(2)}_{(1)} \left(d - 1\right)}{\pi} 
	+ a_{1} \left(d - 1\right) - \frac{a_{2} \alpha^{(2)}_{(1)} \left(d^{2} - 
	3 d + 2\right)}{\pi} + \frac{a_{2} \left(d^{2} - 2 d + 1\right) 
	\sin{\left(\alpha^{(2)}_{(1)} \right)}}{\pi} + a_{2} \left(d^{2} - 2 d + 
	1\right) - \frac{a_{2} \left(d - 1\right) \sin{\left(\beta^{(1)}_{(1)} 
	\right)}}{2 \pi \nu_{(1)}} - \frac{a_{2} \left(d^{2} - 2 d + 1\right) 
	\sin{\left(\beta^{(2)}_{(1)} \right)}}{2 \pi \nu_{(1)}} + 
	\frac{\beta^{(2)}_{(1)} \left(d - 1\right)}{2 \pi} - \frac{d}{2} + 
	\frac{1}{2},
\end{dmath*}
\begin{dmath*}
	T^{\ft}_{3,3} = - \frac{\alpha^{(2)}_{(1)} \nu_{(1)}^{2} \left(d - 
	1\right) \cos{\left(\alpha^{(2)}_{(1)} \right)}}{2 \pi} + 
	\frac{\nu_{(1)}^{2} \left(d - 1\right) \sin{\left(\alpha^{(2)}_{(1)} 
	\right)}}{2 \pi} + \frac{\nu_{(1)}^{2} \left(d - 1\right) 
	\cos{\left(\alpha^{(2)}_{(1)} \right)}}{2} + \frac{\nu_{(1)}^{2}}{2}.
\end{dmath*}

\section{Power series and Hessian spectrum by representation} 
\label{sec:power_series_m5o2}

We present the (fractional) power series of the minima described in 
\pref{thm:power_series} to $O(d^{-5/2})$-order, along with the respective 
eigenvalues arranged by their representation.

\subsection{\pref{thm:power_series}, case \ref{typeII}}
\begin{dmath*}
{a_1 = } 1 + \frac{8}{\pi d^{2}} +  O\prn*{d^{\frac{-5}{2}}},~\\
{a_2 = } - \frac{4}{\pi d^{2}} +  O\prn*{d^{\frac{-5}{2}}},~\\
{a_3 = } \frac{2}{d} + \frac{- \frac{8}{\pi} - 2}{d^{2}} +  
O\prn*{d^{\frac{-5}{2}}},~\\
{a_4 = } \frac{4}{\pi d} + \frac{32}{\pi^{3} d^{\frac{3}{2}}} + \frac{8 
\left(- 
7 \pi^{3} - 8 \pi^{2} + 64\right)}{\pi^{5} d^{2}} +  
O\prn*{d^{\frac{-5}{2}}},~\\\\
{a_5 = } -1 + \frac{\frac{8}{\pi^{2}} + 2 + \frac{8}{\pi}}{d} + - \frac{64 
\left(12 - \pi\right)}{3 \pi^{4} d^{\frac{3}{2}} \left(-2 + \pi\right)} + - 
\frac{2 \left(- 128 \pi^{3} - 40 \pi^{4} - 224 \pi^{2} - 512 \pi + 2560 + 
\pi^{7} + 10 \pi^{6} + 52 \pi^{5}\right)}{\pi^{6} d^{2} \left(-2 + \pi\right)} 
+  O\prn*{d^{\frac{-5}{2}}}.
\end{dmath*}

\subsection{Eigenvalues}

\begin{tabular}{l|l|l|l|l|l|l|l|l}
	$\fx$-Representation& $\frac{(d-2)(d-3)}{2}$ & $ \frac{-2 + \pi}{4 \pi} $  
	\\\hline  
	$\fy$-Representation& $\frac{(d-1)(d-4)}{2}$ & $ \frac{2 + \pi}{4 \pi} $ 
	\\\hline  
	Standard Representation &$d-2$ & 
	$ 0 $ & 
	$ \frac{-2 + \pi}{4 \pi} $ & 
	$ \frac{-2 + \pi}{2 \pi} $ & 
	$ \frac{1}{4} $ & 
	$ \frac{2 + \pi}{4 \pi} $ & 
	$ \frac{d}{4} + \frac{1}{2} $ & 
	\\ \hline  
	Trivial Representation& $1$ & 
	$ 0 $ & 
	$ 0 $ & 
	$ \frac{-2 + \pi}{2 \pi} $ & 
	$ \frac{1}{4} $ & 
	$ \frac{d}{4} + \frac{-4 + \pi + \pi^{2}}{- 8 \pi + 2 \pi^{2}} $ & 
	$ \frac{d}{4} + \frac{1}{2} $ & 
	$ \frac{d}{\pi} + \frac{- 10 \pi + 8 + \pi^{2}}{2 \pi \left(-4 + 
	\pi\right)} $ 
\end{tabular}

\subsection{\pref{thm:power_series}, case \ref{typeI}}
\begin{dmath*}
{a_1 = } -1 + \frac{2}{d} + - \frac{4 \left(- 16 \pi^{2} + \left(-2 + 
\pi\right) \left(- 6 \pi^{3} + 16 \pi + 8 \pi^{2} + \pi^{4}\right) + 32 
\pi\right)}{\pi^{3} d^{2} \left(-2 + \pi\right)^{2}} +  
O\prn*{d^{\frac{-5}{2}}},\\
{a_2 = } \frac{2}{d} + \frac{-2 + \frac{8}{\pi}}{d^{2}} +  
O\prn*{d^{\frac{-5}{2}}},\\
{a_3 = } - \frac{4 \left(- 32 \pi + \left(-2 + \pi\right) \left(- 10 \pi^{2} - 
8 \pi + 3 \pi^{3}\right) + 16 \pi^{2}\right)}{\pi^{3} d^{2} \left(-2 + 
\pi\right)^{2}} +  O\prn*{d^{\frac{-5}{2}}},\\
{a_4 = } \frac{2 - \frac{4}{\pi}}{d} + \frac{32 \left(1 - \pi\right)}{\pi^{3} 
d^{\frac{3}{2}}} + \frac{- \frac{136}{\pi^{2}} - \frac{128}{\pi^{3}} - 2 - 
\frac{512}{\pi^{5}} + \frac{768}{\pi^{4}} + \frac{52}{\pi}}{d^{2}} +  
O\prn*{d^{\frac{-5}{2}}},\\
{a_5 = } 1 + \frac{8 \left(-1 + \pi\right)}{\pi^{2} d} + - \frac{2 \left(- 90 
\pi^{3} - 792 \pi + \pi \sqrt{- 160 \pi^{3} - 12 \pi^{5} - 192 \pi + 64 + 
\pi^{6} + 240 \pi^{2} + 60 \pi^{4}} + 384 + 11 \pi^{4} + 468 \pi^{2}\right)}{3 
\pi^{4} d^{\frac{3}{2}} \left(-2 + \pi\right)} + \frac{h_{4}}{d^{2}} +  
O\prn*{d^{\frac{-5}{2}}}.
\end{dmath*}

\subsection{Eigenvalues}

\begin{tabular}{l|l|l|l|l|l|l|l|l|l}
	$\fx$-Representation& $\frac{(d-2)(d-3)}{2}$ & $ \frac{-2 + \pi}{4 \pi} $  
	\\\hline  
	$\fy$-Representation& $\frac{(d-1)(d-4)}{2}$ & $ \frac{2 + \pi}{4 \pi} $ 
	\\\hline  
	Standard Representation &$d-2$ & 
	$ 0 $ & 
	$ \frac{-2 + \pi}{4 \pi} $ & 
	$ \frac{-2 + \pi}{2 \pi} $ & 
	$ \frac{1}{4} $ & 
	$ \frac{2 + \pi}{4 \pi} $ & 
	$ \frac{d}{4} + \frac{1}{2} $ 
	\\ \hline  
	Trivial Representation& $1$ & 
	$ 0 $ & 
	$ 0 $ & 
	$ \frac{-2 + \pi}{2 \pi} $ & 
	$ \frac{1}{4} $ & 
	$ \frac{d}{4} + \frac{-4 + \pi + \pi^{2}}{- 8 \pi + 2 \pi^{2}} $ & 
	$ \frac{d}{4} + \frac{1}{2} $ & 
	$ \frac{d}{\pi} + \frac{- 10 \pi + 8 + \pi^{2}}{2 \pi \left(-4 + 
		\pi\right)} $ 
\end{tabular}

\subsection{\pref{thm:power_series}, case \ref{typeA}}

\begin{dmath*}
{a_1 =} -1 + \frac{2}{d} + - \frac{4 \left(- 16 \pi^{2} + \left(-2 + 
\pi\right) 
\left(- 6 \pi^{3} + 16 \pi + 8 \pi^{2} + \pi^{4}\right) + 32 
\pi\right)}{\pi^{3} d^{2} \left(-2 + \pi\right)^{2}} +  
O\prn*{d^{\frac{-5}{2}}},~\\
{a_2 =} \frac{2}{d} + \frac{-2 + \frac{8}{\pi}}{d^{2}} +  
O\prn*{d^{\frac{-5}{2}}}.
\end{dmath*}

\subsection{Eigenvalues}

\begin{tabular}{l|l|l|l|l|l|l|l|l}
	$\fx$-Representation& $\frac{(d-1)(d-2)}{2}$ &
	$ \frac{-2 + \pi}{4 \pi} $ & 
	\\\hline  
	$\fy$-Representation& $\frac{d(d-3)}{2}$ &
	$ \frac{2 + \pi}{4 \pi} $ & 
	\\\hline  
	Standard Representation &$d-1$ & 
	$ 0 $ & 
	$ \frac{-2 + \pi}{2 \pi} $ & 
	$ \frac{1}{4} $ & 
	$ \frac{d}{4} + \frac{1}{2} $ 
	\\ \hline  
	Trivial Representation& $1$ & 
	$0$ &
	$ \frac{d}{4} + \frac{-4 + \pi + \pi^{2}}{2 \pi \left(-4 + \pi\right)} $ & 
	$ \frac{d}{\pi} + \frac{- 10 \pi + 8 + \pi^{2}}{2 \pi \left(-4 + 
		\pi\right)} $ 
	
\end{tabular}

\subsection{\pref{thm:power_series}, case \ref{typeM_II}}

\begin{dmath*}
{a_1 = } 1 + \frac{16}{\pi d^{2}} +  O\prn*{d^{\frac{-5}{2}}},\\
{a_2 = } - \frac{8}{\pi d^{2}} +  O\prn*{d^{\frac{-5}{2}}},\\
{a_3 = } \frac{2}{d} + \frac{2 \left(- 8 \pi^{2} - \pi^{3} - 16 + 4 
\pi\right)}{\pi^{2} d^{2} \left(2 + \pi\right)} +  O\prn*{d^{\frac{-5}{2}}},\\
{a_4 = } \frac{4}{\pi d} + \frac{32}{\pi^{3} d^{\frac{3}{2}}} + \frac{4 \left(- 
24 \pi^{4} - 160 \pi^{2} - \pi^{5} + 256 + 192 \pi + 28 \pi^{3}\right)}{\pi^{5} 
d^{2} \left(2 + \pi\right)} +  O\prn*{d^{\frac{-5}{2}}},\\
{a_5 = } -1 + \frac{\frac{8}{\pi^{2}} + 2 + \frac{8}{\pi}}{d} + - \frac{64 
\left(12 - \pi\right)}{3 \pi^{4} d^{\frac{3}{2}} \left(-2 + \pi\right)} + 
\frac{2 \left(- 112 \pi^{5} - 3008 \pi^{2} - 240 \pi^{4} + 5120 + 2560 \pi + 
\pi^{8} + 8 \sqrt{2} \pi^{6} + 4 \sqrt{2} \pi^{7} + 672 \pi^{3} + 12 \pi^{7} + 
104 \pi^{6}\right)}{\pi^{6} d^{2} \left(4 - \pi^{2}\right)} +  
O\prn*{d^{\frac{-5}{2}}},\\
{a_6 = } \frac{2 \left(- 12 \pi + 16 + \pi^{3} + 4 \pi^{2}\right)}{\pi^{2} d 
\left(2 + \pi\right)} + \frac{16 \left(- 24 \pi^{2} + 64 + 24 \pi + \pi^{4} + 4 
\pi^{3}\right)}{\pi^{4} d^{\frac{3}{2}} \left(4 + \pi^{2} + 4 \pi\right)} + 
\frac{2 \left({- 184 \pi^{7} - 26 \pi^{8} - 3968 \pi^{3} - 8192 \pi^{2} - 16 
\sqrt{2} \pi^{7} - 4 \sqrt{2} \pi^{8} - 112 \pi^{5} - \pi^{9} \atop - 16 
\sqrt{2} \pi^{6} + 20480 + 24576 \pi + 864 \pi^{4} + 112 
\pi^{6}}\right)}{\pi^{6} d^{2} \left(8 + \pi^{3} + 12 \pi + 6 \pi^{2}\right)} 
+  O\prn*{d^{\frac{-5}{2}}}.
\end{dmath*}

\subsection{Eigenvalues}

\begin{tabular}{l|l|l|l|l|l|l|l|l|l|l|l}
	$\fx$-Representation &$\frac{(d-3)(d-4)}{2}$ &
	$ \frac{-2 + \pi}{4 \pi} $ & 
	\\\hline  
	$\fy$-Representation &$\frac{(d-2)(d-5)}{2}$&
	$ \frac{2 + \pi}{4 \pi} $ & 
	\\\hline  
	Standard Representation $\fs_{d-2}$ & $d-3$ &
	$ 0 $ & 
	$ \frac{-2 + \pi}{4 \pi} $ & 
	$ \frac{-2 + \pi}{2 \pi} $ & 
	$ \frac{1}{4} $ & 
	$ \frac{2 + \pi}{4 \pi} $ & 
	$ \frac{d}{4} + \frac{1}{2} $ 
	
	\\ \hline  
	
	Standard Representation $\fs_2$ & $1$ &
	$ 0 $ & 
	$ \frac{-2 + \pi}{4 \pi} $ & 
	$ \frac{-2 + \pi}{2 \pi} $ & 
	$ \frac{1}{4} $ & 
	$ \frac{d}{4} + \frac{1}{2} $ &

	\\ \hline  
	Trivial Representation &$1$ & 
	$ 0 $ & 
	$ 0 $ & 
	$ \frac{-2 + \pi}{2 \pi} $ & 
	$ \frac{1}{4} $ & 
	$ \frac{2 + \pi}{4 \pi} $ & 
	$ \frac{d}{4} + \frac{-4 + \pi + \pi^{2}}{- 8 \pi + 2 \pi^{2}} $ \\&& 
	$ \frac{d}{4} + \frac{1}{2} $ & 
	&	$ \frac{d}{\pi} + \frac{- 10 \pi + 8 + \pi^{2}}{2 \pi \left(-4 + 
		\pi\right)} $ & 
	
	\\ \hline  	
	Tensor Representation $\fs_{d-2}\otimes \fs_2$ &$d-3$ & 
	$ \frac{-2 + \pi}{4 \pi} $ & 
	$ \frac{2 + \pi}{4 \pi} $ &

\end{tabular}

\subsection{Identity}
$a_1 = 1, a_2=0$.
\subsection{Eigenvalues}

\begin{tabular}{l|l|l|l|l|l|l|l|l}
	$\fx$-Representation& $\frac{(d-1)(d-2)}{2}$ &
	$ \frac{-2 + \pi}{4 \pi} $ & 
	\\\hline  
	$\fy$-Representation& $\frac{d(d-3)}{2}$ &
	$ \frac{2 + \pi}{4 \pi} $ & 
	\\\hline  
	Standard Representation &$d-1$ & 
	$ 0 $ & 
	$ \frac{-2 + \pi}{2 \pi} $ & 
	$ \frac{1}{4} $ & 
	$ \frac{d}{4} + \frac{1}{2} $ & 
	\\ \hline  
	Trivial Representation& $1$ & 
	$ 0 $ & 
	$ \frac{d}{4} + \frac{-4 + \pi + \pi^{2}}{2 \pi \left(-4 + \pi\right)} $ & 
	$ \frac{d}{\pi} + \frac{- 10 \pi + 8 + \pi^{2}}{2 \pi \left(-4 + 
		\pi\right)} $ &

\end{tabular}

\subsection{\pref{thm:power_series}, case \ref{typeM_I}}

\begin{dmath*}
{a_1 = }-1 + \frac{2}{d} + \frac{-4 + \frac{24}{\pi}}{d^{2}} +  
O\prn*{d^{\frac{-5}{2}}},\\
{a_2 =} \frac{2}{d} + \frac{-2 + \frac{12}{\pi}}{d^{2}} +  
O\prn*{d^{\frac{-5}{2}}},\\
{a_3 =} \frac{8 \left(- 4 \pi - \pi^{2} + 8\right)}{\pi^{2} d^{2} \left(2 + 
\pi\right)} +  O\prn*{d^{\frac{-5}{2}}},\\
{a_4 =} \frac{2 - \frac{4}{\pi}}{d} + \frac{32 \left(1 - \pi\right)}{\pi^{3} 
d^{\frac{3}{2}}} + \frac{2 \left(- 344 \pi^{3} - \pi^{6} - 512 + 4 \pi^{4} + 
384 \pi + 512 \pi^{2} + 26 \pi^{5}\right)}{\pi^{5} d^{2} \left(2 + \pi\right)} 
+  O\prn*{d^{\frac{-5}{2}}},\\
{a_5 =} 1 + \frac{8 \left(-1 + \pi\right)}{\pi^{2} d} + - \frac{8 \left(- 24 
\pi^{3} - 200 \pi + 96 + 3 \pi^{4} + 120 \pi^{2}\right)}{3 \pi^{4} 
d^{\frac{3}{2}} \left(-2 + \pi\right)} + \frac{2 \left(- 224 \pi^{6} - 6816 
\pi^{3} - 496 \pi^{5} - 1088 \pi^{2} - \pi^{8} - 5120 + 9728 \pi + 52 \pi^{7} 
+ 4112 \pi^{4}\right)}{\pi^{6} d^{2} \left(4 - \pi^{2}\right)} +  
O\prn*{d^{\frac{-5}{2}}},\\
{a_6 =} \frac{4 \left(- \pi^{2} - 8 + 6 \pi\right)}{\pi^{2} d \left(2 + 
\pi\right)} + \frac{16 \left(- 40 \pi^{2} - 40 \pi - \pi^{4} + 64 + 20 
\pi^{3}\right)}{\pi^{4} d^{\frac{3}{2}} \left(4 + \pi^{2} + 4 \pi\right)} + 
\frac{4 \left(- 4816 \pi^{4} - 116 \pi^{7} - 4544 \pi^{3} - 10240 + 4096 \pi + 
5 \pi^{8} + 14848 \pi^{2} + 176 \pi^{6} + 1632 \pi^{5}\right)}{\pi^{6} d^{2} 
\left(8 + \pi^{3} + 12 \pi + 6 \pi^{2}\right)} +  O\prn*{d^{\frac{-5}{2}}},\\
\end{dmath*}

\subsection{Eigenvalues}

\begin{tabular}{l|l|l|l|l|l|l|l|l|l|l|l}
	$\fx$-Representation &$\frac{(d-3)(d-4)}{2}$ &
	$ \frac{-2 + \pi}{4 \pi} $ & 
	\\\hline  
	$\fy$-Representation &$\frac{(d-2)(d-5)}{2}$&
	$ \frac{2 + \pi}{4 \pi} $ & 
	\\\hline  
	Standard Representation $\fs_{d-2}$ & $d-3$ &
	$ 0 $ & 
	$ \frac{-2 + \pi}{4 \pi} $ & 
	$ \frac{-2 + \pi}{2 \pi} $ & 
	$ \frac{1}{4} $ & 
	$ \frac{2 + \pi}{4 \pi} $ & 
	$ \frac{d}{4} + \frac{1}{2} $ & 
	
	\\ \hline  
	
	Standard Representation $\fs_2$ & $1$ &
	$ 0 $ & 
	$ \frac{-2 + \pi}{4 \pi} $ & 
	$ \frac{-2 + \pi}{2 \pi} $ & 
	$ \frac{1}{4} $ & 
	$ \frac{d}{4} + \frac{1}{2} $ &

	\\ \hline  
	Trivial Representation &$1$ & 
	$ 0 $ & 
	$ 0 $ & 
	$ \frac{-2 + \pi}{2 \pi} $ & 
	$ \frac{1}{4} $ & 
	$ \frac{2 + \pi}{4 \pi} $ & 
	$ \frac{d}{4} + \frac{-4 + \pi + \pi^{2}}{- 8 \pi + 2 \pi^{2}} $ \\&& 
	$ \frac{d}{4} + \frac{1}{2} $ & 
	&	$ \frac{d}{\pi} + \frac{- 10 \pi + 8 + \pi^{2}}{2 \pi \left(-4 + 
		\pi\right)} $ & 
	
	\\ \hline  	
	Tensor Representation $\fs_{d-2}\otimes \fs_2$ &$d-3$ & 
	$ \frac{-2 + \pi}{4 \pi} $ & 
	$ \frac{2 + \pi}{4 \pi} $ & 
	
\end{tabular}

\subsection{\pref{thm:power_series}, case \ref{typeN_II}}

\begin{dmath*}
{a_1 =} 1 + \frac{24}{\pi d^{2}} +  O\prn*{d^{\frac{-5}{2}}},\\
{a_2 =} - \frac{12}{\pi d^{2}} +  O\prn*{d^{\frac{-5}{2}}},\\
{a_3 =} \frac{2}{d} + \frac{2 \left(- 10 \pi^{2} - 32 - \pi^{3} + 16 
\pi\right)}{\pi^{2} d^{2} \left(2 + \pi\right)} +  O\prn*{d^{\frac{-5}{2}}},\\
a_4 = \frac{4}{\pi d} + \frac{32}{\pi^{3} d^{\frac{3}{2}}} + \frac{8 \left(- 
17 \pi^{4} - 144 \pi^{2} - \pi^{5} + 128 + 128 \pi + 50 
\pi^{3}\right)}{\pi^{5} d^{2} \left(2 + \pi\right)} +  
O\prn*{d^{\frac{-5}{2}}},\\
{a_5 =} -1 + \frac{\frac{8}{\pi^{2}} + 2 + \frac{8}{\pi}}{d} + - \frac{64 
\left(12 - \pi\right)}{3 \pi^{4} d^{\frac{3}{2}} \left(-2 + \pi\right)} + 
\frac{2 \left(- 288 \pi^{5} - 5056 \pi^{2} - 272 \pi^{4} + 5120 + \pi^{8} + 
3584 \pi + 8 \sqrt{3} \pi^{6} + 4 \sqrt{3} \pi^{7} + 16 \pi^{7} + 1824 \pi^{3} 
+ 144 \pi^{6}\right)}{\pi^{6} d^{2} \left(4 - \pi^{2}\right)} +  
O\prn*{d^{\frac{-5}{2}}},\\
{a_6 =} \frac{2 \left(- 12 \pi + 16 + \pi^{3} + 4 \pi^{2}\right)}{\pi^{2} d 
\left(2 + \pi\right)} + \frac{16 \left(- 24 \pi^{2} + 64 + 24 \pi + \pi^{4} + 
4 \pi^{3}\right)}{\pi^{4} d^{\frac{3}{2}} \left(4 + \pi^{2} + 4 \pi\right)} + 
\frac{2 \left({- 1920 \pi^{5} - 164 \pi^{7} - 28 \pi^{8} - 4992 \pi^{3} - 13824 
\pi^{2} - 16 \sqrt{3} \pi^{7} - 4 \sqrt{3} \pi^{8} - \pi^{9} \atop - 16 
\sqrt{3} 
\pi^{6} + 20480 + 28672 \pi + 312 \pi^{6} + 4960 \pi^{4}} \right) 
}{\pi^{6} d^{2} 
\left(8 + \pi^{3} + 12 \pi + 6 \pi^{2}\right)} +  
O\prn*{d^{\frac{-5}{2}}}.
\end{dmath*}

\subsection{Eigenvalues}

\begin{tabular}{l|l|l|l|l|l|l|l|l|l|l|l}
	$\fx_{d-3}$-Representation &$\frac{(d-5)(d-4)}{2}$ &
	$ \frac{-2 + \pi}{4 \pi}$ & 
	\\\hline  
	$\fx_3$-Representation &$1$ &
	$ \frac{-2 + \pi}{4 \pi}$ & 
	\\\hline  
	$\fy$-Representation &$\frac{(d-3)(d-6)}{2}$&
	$ \frac{2 + \pi}{4 \pi} $ 
	\\\hline  
	Standard Representation $\fs_{d-3}$ & $d-4$ &
	$ 0 $ & 
	$ \frac{-2 + \pi}{4 \pi} $ & 
	$ \frac{-2 + \pi}{2 \pi} $ & 
	$ \frac{1}{4} $ & 
	$ \frac{2 + \pi}{4 \pi} $ & 
	$ \frac{d}{4} + \frac{1}{2} $ & 
	
	\\ \hline  
	
	Standard Representation $\fs_3$ & $2$ &
	$ 0 $ & 
	$ \frac{-2 + \pi}{4 \pi} $ & 
	$ \frac{-2 + \pi}{2 \pi} $ & 
	$ \frac{1}{4} $ & 
	$ \frac{d}{4} + \frac{1}{2} $ &

	\\ \hline  
	Trivial Representation &$1$ & 
	$ 0 $ & 
	$ 0 $ & 
	$ \frac{-2 + \pi}{2 \pi} $ & 
	$ \frac{1}{4} $ & 
	$ \frac{2 + \pi}{4 \pi} $ & 
	$ \frac{d}{4} + \frac{-4 + \pi + \pi^{2}}{- 8 \pi + 2 \pi^{2}} $ \\&& 
	$ \frac{d}{4} + \frac{1}{2} $ & 
	&	$ \frac{d}{\pi} + \frac{- 10 \pi + 8 + \pi^{2}}{2 \pi \left(-4 + 
		\pi\right)} $ & 
	
	\\ \hline  	
	Tensor Representation $\fs_{d-3}\otimes \fs_3$ &$2d-8$ & 
	$ \frac{-2 + \pi}{4 \pi} $ & 
	$ \frac{2 + \pi}{4 \pi} $ &

\end{tabular}